%% file: lrpaper.tex
\documentclass[journal]{IEEEtran}

\ifCLASSINFOpdf

\else

\fi

\usepackage{graphicx}
\graphicspath{ {./images/} }

\usepackage[T1]{fontenc}
\usepackage[hidelinks]{hyperref}

\usepackage{amsmath}

\usepackage{amssymb}

\usepackage{algorithm}

\usepackage{algpseudocode}

\usepackage{array}

\usepackage{multirow}

\usepackage{setspace}

\usepackage{subcaption}

\usepackage{booktabs}

\usepackage{caption}

\usepackage{cite}

\newtheorem{theorem}{Theorem}

\begin{document}

\renewcommand{\thetable}{\arabic{table}}
\captionsetup[table]{labelformat=simple, labelsep=colon} 
\renewcommand{\tablename}{Table} 

\title{ExpTest: Loss-Curve Hypothesis Testing for Autonomous Learning-Rate Selection in Deep Neural Networks}

\author{Zan~Chaudhry
        and~Naoko~Mizuno
\thanks{Z. Chaudhry and N. Mizuno were with the Laboratory of Structural Cell Biology, National Institutes of Health, Bethesda,
MD, 20892 USA; Z. Chaudhry is now at the Ragon Institute, Cambridge, MA, 02139.}
}

\maketitle

\begin{abstract}
Hyperparameter tuning remains a significant challenge in the training of deep neural networks (DNNs), requiring manual search or time-intensive grid searches that increase resource costs and limit the accessibility of machine learning. The global initial learning rate is among the most consequential of these hyperparameters. Adaptive and scheduling-based methods manage the learning rate during training but still require manual selection of an initial global value; learning-rate-free alternatives remove this selection at the cost of performance or stability on non-convex problems. We present ExpTest, an autonomous learning-rate controller that treats the training loss curve as an online signal and performs sequential statistical tests on theoretically motivated windows to detect convergent behavior and trigger learning-rate reductions. The framework combines a covariance-based initial learning-rate estimate, curvature-motivated window sizing, and two-phase test-driven decay, drawing on the approximately exponential decay behavior predicted under linearized network dynamics. We provide a mathematical motivation for ExpTest and evaluate it on regression, classification, forecasting, and natural-language tasks across fully-connected, convolutional, transformer-based, and pretrained architectures. Across these tasks, ExpTest achieves competitive performance relative to hand-tuned SGD-based baselines and recent learning-rate-free methods, without manual initial learning-rate selection or predefined scheduling.
\end{abstract}

\begin{IEEEkeywords}
Learning rate tuning, learning rate scheduling, stochastic gradient descent (SGD), linearized neural networks, loss curve, learning curve, classification, regression, neural networks.
\end{IEEEkeywords}

\section{Introduction}
\IEEEPARstart{M}{achine} learning (ML) and artificial intelligence (AI) are experiencing tremendous growth, particularly the development of deep neural networks (DNNs), which have become ubiquitous tools, achieving state-of-the-art performance in applications from computer vision to natural language processing, in fields from consumer devices to scientific research.

Despite these successes, DNN training remains an open problem. Hyperparameter tuning has earned comparisons to alchemy and requires trial and error, expertise, and luck \cite{doi:10.1126/science.360.6388.478}. Offline methods exist, predominantly grid/random searches; however, these are time and resource intensive. Hyperparameter tuning remains a major barrier that prevents end-users of DNNs from fine-tuning models, ultimately delaying the diffusion of AI/ML innovations \cite{yu2020hyperparameteroptimizationreviewalgorithms}.

DNNs are typically trained with gradient descent-based optimizations, iteratively minimizing an objective function. In its most basic form, the gradient descent rule is given by:
\begin{equation*}
    \theta_{t+1} = \theta_{t} - \eta \nabla \mathcal{L}_{\theta} (t)
\end{equation*}
The model parameters are adjusted by taking the derivative of the objective function ($\mathcal{L}$) with respect to the parameters ($\theta$) at each time point ($t$), and moving towards minima by stepping in the negative gradient direction, as controlled by the step size parameter ($\eta$). This step size is commonly referred to as the learning rate.

The learning rate is often considered the most important hyperparameter, arising in all forms of gradient-descent-based optimization \cite{bengio2012practicalrecommendationsgradientbasedtraining}. Learning rate adjustment techniques have become popular in DNN training, including momentum, annealing, cycling, and adaptive algorithms such as Adam \cite{POLYAK19641, JACOBS1988295, smith2017cyclicallearningratestraining, kingma2017adammethodstochasticoptimization}. However, these methods still require the selection of an initial global learning rate. Currently, heuristic methods are used to choose the initial global learning rate by interpreting the loss curves acquired during DNN training \cite{smith2018disciplinedapproachneuralnetwork}. The aim is to identify the highest learning rate that produces a converging loss curve to minimize training time. Empirical work supports power-law or exponential decay-like loss curve behavior to be indicative of convergence \cite{9944190}.

Recently, linearized networks have seen increasing use as tools for studying the evolution of DNNs under training \cite{misiakiewicz2023lectureslinearizedneuralnetworks, NEURIPS2018_5a4be1fa, NEURIPS2021_4b5deb9a, NEURIPS2019_0d1a9651}. Additionally, signal processing approaches to learning rate tuning have been explored, treating the loss curve data as a time series signal \cite{9930853}. Here, we utilize insights from linearized networks to design a set of lightweight algorithms for autonomous learning-rate control that do not require a choice of initial global learning rate. The general framework of ExpTest involves initial estimation of a starting learning rate based on linear models, potential revision of this starting value using real-time curvature information, and sequential statistical tests on the loss curve to detect convergence conditions over windows motivated by the approximately exponential decay behavior predicted under linearized network dynamics. ExpTest introduces two interpretable hyperparameters (similar to momentum values or annealing decay rates), and training results are robust to their choice within the ranges tested. We provide a mathematical motivation for the framework and validate it experimentally across several datasets and architectures. We emphasize what ExpTest is and is not. ExpTest is not a new per-parameter adaptive optimizer, and it is not intended to replace extensively hand-tuned training recipes where maximum benchmark performance is the sole objective. It is a controller-style wrapper that can be paired with SGD or SGD with momentum, targeting the practical setting in which manual initial learning-rate selection is burdensome or unavailable and autonomous adaptation during training is valuable.

\section{Related Work}

\subsection{Learning Rate Scheduling}
A variety of learning rate scheduling strategies have been proposed to improve DNN training. Fixed schedules, such as step decay and exponential decay, reduce the learning rate at predetermined intervals and remain widely used in practice \cite{bengio2012practicalrecommendationsgradientbasedtraining}. Cyclical learning rate policies periodically vary the learning rate between bounds \cite{smith2017cyclicallearningratestraining}, and warm restart methods such as SGDR reset the learning rate according to a cosine annealing schedule \cite{loshchilov2017sgdr}. Heuristic approaches, such as the learning rate range test, estimate a suitable range by sweeping the learning rate during a short pre-training phase \cite{smith2018disciplinedapproachneuralnetwork}. While effective when well-tuned, all of these methods require the practitioner to specify one or more learning rate bounds or an initial learning rate, and their performance is sensitive to these choices.

\subsection{Adaptive Gradient Methods}
Per-parameter adaptive methods construct element-wise learning rates from gradient history. AdaGrad accumulates squared gradients to scale updates, RMSprop uses an exponential moving average of squared gradients, and Adam combines first- and second-moment estimates \cite{kingma2017adammethodstochasticoptimization, RMSProp}. AdamW introduces decoupled weight decay for improved regularization \cite{loshchilov2019decoupledweightdecayregularization}. These methods reduce sensitivity to the learning rate relative to plain SGD, but still require an initial global learning rate, and empirical evidence suggests they can generalize worse than SGD with a well-chosen schedule \cite{10.5555/3294996.3295170}. Adadelta is a notable exception that requires no initial learning rate \cite{zeiler2012adadeltaadaptivelearningrate}; however, it often converges more slowly and achieves worse final performance than other adaptive methods, limiting its practical adoption.

\subsection{Learning Rate-Free and Automatic Step Size Methods}
Recent work has sought to eliminate learning rate selection entirely. D-Adaptation estimates the distance to the optimum online and sets the step size accordingly, requiring no learning rate hyperparameter \cite{defazio2023dadaptation}. Prodigy improves upon D-Adaptation with faster adaptation dynamics \cite{mishchenko2024prodigy}. DoG (Distance over Gradients) derives a parameter-free step size from the ratio of the distance traveled to the accumulated gradient norms \cite{ivgi2023dog}. Hypergradient descent methods differentiate through the learning rate update itself to adapt the step size during training \cite{baydin2018hypergradient}. Stochastic Polyak step sizes extend the classical Polyak step to the stochastic setting by leveraging knowledge of the optimal loss value \cite{loizou2021sps}. These approaches address the same fundamental problem as ExpTest; however, they are grounded in optimization-theoretic constructions (distance estimation, hypergradients, or known optimal values). ExpTest takes a complementary approach: it derives its learning rate schedule from the statistical structure of the loss curve, using linearized neural network theory to motivate exponential decay detection and hypothesis testing to drive learning rate adjustments.

\subsection{Linearized Neural Networks and Neural Tangent Kernel}
The neural tangent kernel (NTK) framework shows that sufficiently wide networks evolve as linear models under gradient descent \cite{NEURIPS2018_5a4be1fa, NEURIPS2019_0d1a9651}. This linearization has been used to study generalization \cite{NEURIPS2021_4b5deb9a} and training dynamics \cite{misiakiewicz2023lectureslinearizedneuralnetworks}. ExpTest draws on this body of work to motivate the use of exponential-like loss decay as a convergence signature, providing the basis for the window size calculation and initial convergence tests.

\subsection{Loss Curve Analysis}
The shape of learning curves has been studied extensively as a diagnostic tool for model selection and training progress \cite{9944190}. Signal processing perspectives have also been applied: Wang and Zhang treat the loss curve as a feedback signal and design a PID controller-based learning rate scheduler \cite{9930853}. ExpTest similarly treats the loss curve as a real-time signal, but employs statistical hypothesis testing ($F$-tests and $t$-tests) rather than control-theoretic feedback, and grounds the expected signal shape in NTK theory rather than empirical observation alone.

\section{Mathematical Motivation}

\subsection{The Linear Case}
We aim to motivate exponential-like loss decay as a useful diagnostic signature under convergent training, beginning from cases where this behavior can be derived explicitly. Additionally, we aim to define an upper bound on the learning rate with which to begin learning rate searching. We start with the linear case, re-establishing and extending some classic results \cite{10.5555/249049}. Consider a single linear layer $\mathbf{T}\in \mathbb{R}^{m \times n}$ with input vector $\Vec{x}\in \mathbb{R}^{n}$ and its corresponding true output vector $\Vec{y}\in \mathbb{R}^{m}$. We define the predicted output vector for this input as:
\begin{align} \label{eq:Tx}
\mathbf{T}\Vec{x}=\hat{y}
\end{align}
We define the mean-squared-error (MSE) loss function:
\begin{align}
\mathcal{L}=\frac{1}{2m}\sum_{i=1}^{m} \Big(\hat{y}_i-{y}_i\Big)^2=\frac{1}{2m}\Big(\hat{y}-\Vec{y}\Big)^{\text{T}}\Big(\hat{y}-\Vec{y}\Big)
\end{align}
We then generalize these definitions to matrix-valued inputs/outputs. We define each input vector $\Vec{x}_{j} \in X$ as a member of the set of input vectors constituting the training data $X \subseteq \mathbb{R}^{n}$ with size: $|X|=s$. We assemble all of these vectors as the columns of a matrix, $\mathbf{X} \in \mathbb{R}^{n \times s}$. Then, we assemble all of the true output vectors similarly into a matrix, $\mathbf{Y} \in \mathbb{R}^{m \times s}$. Thus, we redefine our objective, with:
\begin{align}
\mathbf{T} \mathbf{X} = \mathbf{\hat{Y}}
\end{align}
Where we aim to find the $\mathbf{T}$ that minimizes a new loss function defined using the trace operator, or $\text{tr}$:
\begin{align}
\mathcal{L}=\frac{1}{2ms}\text{tr}\Big(\Big(\mathbf{\hat{Y}}-\mathbf{Y}\Big)^{\text{T}}\Big(\mathbf{\hat{Y}}-\mathbf{Y}\Big)\Big)
\end{align}
We consider the tuning of $\mathbf{T}$ by gradient descent:
\begin{align} \label{eq:gd_update}
\mathbf{T}[k+1]=\mathbf{T}[k] - \eta \frac{\partial\mathcal{L}}{\partial \mathbf{T}} \Biggr|_{\mathbf{T}=\mathbf{T}[k]}
\end{align}
Here, $\mathbf{T}[k]$ is defined as a discrete, matrix-valued function in terms of iteration number, $k$, and $\eta$ is the step size taken in the direction opposite the gradient (the learning rate). From here onward, we present main results, relegating more extensive derivations to the Supplementary Material (SM). In SM Section~\ref{sec:sm_linear}, we revisit the convergence properties of the update rule in Eq.~\ref{eq:gd_update}, showing that $\mathbf{T}[k]$ in explicit form is given by:
\begin{align}
\mathbf{T}[k] = (\mathbf{T}[0]-\mathbf{T_\infty})\mathbf{A}^k + \mathbf{T_\infty}
\end{align}
For constant matrices $\mathbf{A} = \mathbf{I_n} - \frac{\eta}{ms} \mathbf{X}\mathbf{X}^\text{T}$ and $\mathbf{T_\infty} = \mathbf{Y}\mathbf{X}^\text{T}(\mathbf{X}\mathbf{X}^\text{T})^{-1}$. Clearly, the eigenvalues of $\mathbf{A}$ determine the convergence of $\mathbf{T}$, and these eigenvalues are dependent on the learning rate. This gives rise to the classic result \cite{10.5555/249049}:
\begin{align} \label{eq:lincase_bound}
\text{ }\frac{2ms}{\lambda_{\text{max}}(s-1)} > \eta
\end{align}
which upper bounds the learning rate based on the maximum eigenvalue of the sample covariance matrix, $\mathbf{\hat{\Sigma}_{xx}} = \frac{1}{s-1} \mathbf{X}\mathbf{X}^\text{T}$. Given this upper bound, the eigenvalues of $\mathbf{A}$ all have magnitude less than 1 and act as decaying exponential terms when raised to the power of $k$. Thus, the elements of $\mathbf{T}[k]$ are given by linear combinations of decaying exponentials plus a constant term (from the contribution of $\mathbf{T_\infty}$); the explicit matrix form is given in SM Section~\ref{sec:sm_linear}. Taking a Taylor approximation and adopting continuous-time dynamics (``gradient flow''), we recover a differential equation:
\begin{align} \label{eq:gradient_flow}
\frac{d \mathcal{L}}{dt}= -||\nabla\mathcal{L}|| ^2
\end{align}
It follows (from the form of $\mathbf{T}(t)$, the continuous-time counterpart to $\mathbf{T}[k]$) that:
\begin{align}
||\nabla \mathcal{L}||^2 = \sum_{j=1}^{mn}\Bigg(\sum_{i=1}^n C_{ij} e^{-\alpha_it}\Bigg)^2
\end{align}
Returning to the differential equation:
\begin{align}
d\mathcal{L} &= -dt \sum_{j=1}^{mn}\Bigg(\sum_{i=1}^n C_{ij} e^{-\alpha_it}\Bigg)^2 \nonumber \\
\mathcal{L} &= C_{\text{int}} + \sum_{j=1}^{mn}\sum_{i=1}^n \sum_{h=1}^n \frac{C_{hj}C_{ij}}{\alpha_{hj} + \alpha_{ij}} e^{-(\alpha_i + \alpha_h) t}
\end{align}
Once again, for more explicit derivations, please refer to SM Section~\ref{sec:sm_linear}. Because the loss in general is a sum of many exponentials with disparate decay rates, we adopt a single exponential as a \textit{motivating surrogate} for the dominant decay signature that the statistical tests of Section~\ref{sec:alg_dev} will attempt to detect, rather than as a claim that a given training trajectory is literally described by one exponential. In practice, we solve numerically for a single exponential decay via least-squares regression. We also provide two analytical justifications in SM Section~\ref{sec:sm_linear}: a Taylor polynomial matching argument that is exact when the decay rates are similar, and a dominant-term argument that applies when some decay rates are much smaller than others. In both cases:
$$\mathcal{L} \approx C_{\text{int}} + C e^{-ct}$$
Furthermore, any approximation of this sum as a single exponential is significantly improved by the monotonicity of $\mathcal{L}$ in the gradient flow approximation (Equation~\ref{eq:gradient_flow}). We extend to stochastic gradient descent (SGD) in SM Section~\ref{sec:sm_sgd}.

\subsection{Nonlinearities and Classification Loss Functions} \label{sec:nonlinear}

A significant body of previous work has demonstrated that the early-time training dynamics of multilayer networks with nonlinearities can be well-approximated by linear models \cite{misiakiewicz2023lectureslinearizedneuralnetworks, NEURIPS2018_5a4be1fa, NEURIPS2021_4b5deb9a, NEURIPS2019_0d1a9651}. Thus, we continue in this vein to justify extending the findings from linear theory to neural networks with nonlinearities. We begin by considering the arbitrary depth neural network, $f(\theta, \Vec{x})$, where $\theta$ refers to the vectorized parameters of the network (we omit the vector arrow for $f$ and $\theta$  for clarity). Each layer of the network is defined as:

\begin{align} \label{eq:layer}
\Vec{y}_{l} = \sigma \Big( \mathbf{W}_l  \Vec{y}_{l-1} + \Vec{b}_{l}\Big)
\end{align}
Here, the input $\Vec{y}_{l-1}$ is the output of the previous layer (where $l$ refers to layer number), $\textbf{W}_l$ and $\Vec{b}_l$ are the weight matrix and bias term, respectively, of the current layer, and $\sigma$ is a coordinate-wise nonlinear function (an ``activation'' function). We assume $\sigma$ to be Lipschitz-continuous. Adopting continuous-time dynamics as before (see SM Section~\ref{sec:sm_nonlinear} for the full derivation), we have:

\begin{align}
\frac{df}{dt} = -\nabla_{\theta}f \text{ } \nabla_{\theta}f ^\text{ T }\nabla_{f}\mathcal{L}
\end{align}

If we adopt a linear approximation of $f$ around $\theta = \theta(0)$, then we arrive at the neural tangent kernel (NTK) description, which is an exact solution for infinite-width networks \cite{NEURIPS2018_5a4be1fa}. We substitute, $\mathbf{K} =\nabla_{\theta_0}f \text{ } \nabla_{\theta_0}f ^\text{ T}$ (notice that this is a positive semi-definite matrix):

\begin{align}
\frac{df}{dt} = -\mathbf{K}\nabla_{f}\mathcal{L}
\end{align}

In general, this is a nonlinear equation without a simple solution. However, in the cases of MSE-loss and cross-entropy loss (up to a second order approximation), which cover the majority of regression and classification tasks, respectively, the solution can easily be found. For both MSE-loss and CE-loss (up to second order; see SM Section~\ref{sec:sm_nonlinear} for full derivations), we have for some constant vector, $\Vec{c}$, and some positive semi-definite constant matrix, $\mathbf{C}$:

\begin{align} \label{eq:two}
\frac{df}{dt} = -\mathbf{KC}(f - \Vec{c})
\end{align}

The solution is:

\begin{align}
f(t) = e^{-t\mathbf{KC}}\Big(f_0 - \Vec{c}\Big) + \Vec{c}
\end{align}

This is familiar as each element of $f$ is a sum of exponential decays plus a constant term, as in the linear case. Thus, for MSE-loss and CE-loss (up to second order), the early-time dynamics of $\mathcal{L}(t)$ admit the same single-exponential surrogate:

$$\mathcal{L}(t) = C + \sum A e^{-Bt} \approx c + ae^{-bt}$$

\subsection{Revising the Learning Rate Bound} \label{sec:revising_bound}
Previously, we showed the upper bound learning rate for a linear network trained with MSE loss (Eq.~\ref{eq:lincase_bound}). Now we demonstrate the potential impact of nonlinearities on this upper bound with a simple, illustrative example. Instead of applying the transformation matrix, $\mathbf{T}$, of Eq.~\ref{eq:Tx} to the data matrix, $\mathbf{X}$, we apply it to $\mathbf{Z} = \sigma(\mathbf{WX})$ for some additional transformation matrix, $\mathbf{W}$, and some coordinate-wise nonlinearity, $\sigma$. Now we know that the convergence properties of gradient descent on $\mathbf{T}$ are proportional to the reciprocal of the maximum eigenvalue of $\mathbf{XX}^\text{T}$ in the original case. But now we instead have the matrix, $\mathbf{ZZ}^\text{T}$. If the nonlinearity behaves in such a way as to reduce the maximum eigenvalue of $\mathbf{ZZ}^\text{T}$, then the maximum learning rate for training $\mathbf{T}$ by gradient descent will increase. Thus, nonlinearities can warp the loss surface in such a way as to reduce the curvature and increase the maximum learning rate of convergence \cite{KAWAGUCHI2019167}.

So then, how can we estimate a reasonable upper bound learning rate for a nonlinear network? We utilize a well-known fact for a Lipschitz continuous gradient. Lipschitz continuity implies that for some $L<\infty$:
\begin{equation} \label{eq:lipschitz}
||\nabla \mathcal{L}(\mathbf{x}) - \nabla \mathcal{L}(\mathbf{y})|| \leq L||\mathbf{x} - \mathbf{y}||
\end{equation}
It can easily be shown that for gradient descent steps to have monotonic decrease, we must have $\eta <2/L$ (we omit the proof and direct the reader to \cite{bottou2018}). 

\section{Algorithm Development} \label{sec:alg_dev}
\subsection{ExpTest Framework}
The exponential loss curve behavior established thus far is already used as part of heuristic manual learning rate tuning methods \cite{9944190}. Here, we propose a basic framework to automate this initial process and subsequent learning rate tuning. In its simplest form, ExpTest entails 1) estimating an upper bound learning rate to begin training; 2) training for an initial window with window size based around the expected exponential decay behavior; 3) performing a statistical test to detect convergence behavior and adjusting the learning rate if convergence is not detected; and 4) repeating for future windows. We utilize idealized models of the loss to generate heuristic (but automate-able) tests to achieve these steps. While based on idealized/heuristic methods, we rigorously investigate the convergence properties mathematically (Section~\ref{sec:convergence}) and empirically (Section~\ref{sec:experiments}) in the following sections.

\subsubsection{Initial Learning Rate Estimation} \label{sec:lr_upper}
For the first part of the algorithm, empirical work has shown that as model complexity (layer number, layer width, etc.) increases, the learning rate of convergence generally decreases, likely due to increasing complexity of the loss surface \cite{10.5555/3618408.3618999}. Theoretical analyses of the convergence properties of multilayer linear networks have similarly demonstrated inverse power relationships between learning rate and network depth \cite{arora2019convergenceanalysisgradientdescent}. Thus, the learning rate that guarantees convergence of a given multilayer nonlinear network is likely less than the learning rate that guarantees convergence for a single layer linear network on the same problem. So we can set a starting upper bound using single-layer linear models, as described for MSE-loss in Equation~\ref{eq:lincase_bound} (derived more fully in SM Section~\ref{sec:sm_linear}) and for CE-loss in SM Section~\ref{sec:sm_ce_bound}.

However, we highlight two caveats. First, as described in Section~\ref{sec:revising_bound}, nonlinearities may relax this upper bound. To address this, we propose a simple estimator of the Lipschitz constant to set an initial learning rate $\eta_0$. During a short warm-up window, we calculate the local curvature estimate $\hat{L}_k$ at each step $k$:
\begin{equation}
\hat{L}_k = \frac{||\nabla \mathcal{L}(\theta_k) - \nabla \mathcal{L}(\theta_{k-1})||} {||\theta_{k} - \theta_{k-1}||}
\end{equation}
At the end of the window, we set the initial learning rate using the \textit{minimum} observed curvature: $\eta_{0} = 2/\min(\hat{L})$.

We deliberately choose $\min(\hat{L})$ rather than the maximum to establish an \textit{optimistic} learning rate limit. Since the true global Lipschitz constant $L^*$ satisfies $L^* \ge \min(\hat{L})$, this choice often results in $\eta_0 > 2/L^*$. While theoretically unstable for fixed-step SGD, this aggressive initialization allows extensive exploration of the loss surface, for example through the ``catapult mechanism'' \cite{lewkowycz2020largelearningratephase}, allowing the model to escape sharp local minima early in training. Crucially, the subsequent stability of the algorithm is provided not by this initial choice, but by the ExpTest mechanism itself, which monotonically decays the learning rate until the stability condition $\eta_t < 2/L^*$ is satisfied (Theorem~\ref{thm:convergence}, Condition 1). This Lipschitz-based estimate is used only once, at initialization, to select an optimistic starting learning rate. All subsequent learning rate updates in ExpTest are driven exclusively by statistical convergence tests.

Second, the linear model estimate is related to the maximum eigenvalue of the covariance matrix of the input data, but in many tasks, the full covariance structure is disregarded in favor of local features (i.e. the main reasoning behind convolutional neural networks). Therefore, if the kernel receptive field is significantly smaller than the input size, it may be advantageous to instead compute the covariance over patches of the inputs corresponding to the receptive field size.

\subsubsection{Window Size Calculation}
Now we explicitly define the window size for performing the regression steps of this algorithm (and over which to measure $\hat{L}$). We propose a method based around the NTK description. We begin by considering the natural distinguishing feature of exponential decay relative to linear decay: curvature. Curvature is defined by:

\begin{align}
\kappa(t) = \frac{|f''(t)|}{(1+f'(t)^2)^\frac{3}{2}}
\end{align}
Now consider once more the NTK description:

\begin{equation} \label{repeated}\tag{\ref{eq:two}}
f(t) = e^{-t\mathbf{KC}}\Big(f_0 - \Vec{c}\Big) + \Vec{c}
\end{equation}

There is some decay rate within the eigenvalues of the matrix $\mathbf{KC}$ that maximizes the time-point at which the point of maximum curvature exists (i.e. an exponential possessing this decay rate achieves maximum curvature at the latest point in time and thus provides a reasonable upper bound for the time-point at which the loss curve achieves maximum curvature). We can call it $\lambda_{t, \text{max}}$. Furthermore, when dealing with the discretized case, we consider a step-size in time of $\eta$. Thus, we are interested in the function (for some constant $C_\text{exp}$):

\begin{align} \label{eq:decay_fn}
f_{\lambda_{t,\text{max}}}(t) = C_\text{exp}e^{-\eta \lambda_{t, \text{max}}t}
\end{align}

Rather than eigendecompose $\mathbf{KC}$, we can solve for the $t_\text{max}$ that maximizes $\kappa$ as a function of $\lambda_{t, \text{max}}$, and then find the maximum of that function (i.e. the maximum possible time-point of maximum curvature). Then we can select a window size such that we have an equal number of points before and after this maximum time-point. We show in SM Section~\ref{sec:sm_window} that the maximum possible time-point of maximum curvature for the function in Equation~\ref{eq:decay_fn} occurs at:

\begin{align}
t_\text{max} = \frac{\sqrt{2} C_\text{exp}}{e}
\end{align}
We can set $C_\text{exp}$ as the loss value at time zero, $\mathcal{L}_0$, giving the window size for point collection, $w$, as:

\begin{align} \label{eq:window_size}
w = \Biggr\lfloor \frac{2 \sqrt{2}\mathcal{L}_0}{\eta e} + \frac{1}{2} \Biggr\rfloor
\end{align}
We adopt nearest-integer rounding conventions to avoid a fractional window size. 

Thus far, this analysis has been independent of batch size, which will naturally impact the window of convergence. We propose a simple correction factor for differing batch sizes. In mini-batch SGD, we approximate the full gradient at each step by the mean gradient over some batch of size $B$. Thus, there will be some amount of noise in the gradient prediction, deviating from the optimal path. We can view the mini-batch gradient steps as oscillating around some true path, which we can approximately recover by averaging the direction over several steps (as in the exponential moving averages of momentum) \cite{POLYAK19641}. To determine the degree of ``misstepping," we can measure the path length over some window (the sum of the magnitudes of the gradient step vectors), and we can divide this by the displacement over this window (the magnitude of the sum of the gradient step vectors). Explicitly, we define the window correction, $c_w$ over some window of gradient descent steps, $w$, as:

\begin{align} \label{eq:window_correction}
c_w = \frac{\sum_{i=1}^w \Big| \Big| \Big(\nabla_\theta \mathcal{L} \Big)_i \Big| \Big|}{\Big| \Big| \sum_{i=1}^w \Big( \nabla_\theta \mathcal{L} \Big)_i \Big| \Big|}
\end{align}

Note that for full-batch gradient descent, the gradient direction is exact at every step, so the path length equals the displacement and $c_w = 1$, recovering the uncorrected window size as expected. For mini-batch SGD, $c_w > 1$, and for a given predicted window size from Equation~\ref{eq:window_size}, $w$, we can approximate the corrected window size as $c_w w$, to account for misstepping in the gradient direction at smaller batch sizes, which increases convergence time.

\subsubsection{Convergence Hypothesis Testing}
 We present two variations on convergence hypothesis testing. In the first version, we can directly check for exponential decay behavior by fitting two least-squares regressions to the loss curve, an exponential decay and a linear function, and we can compare the fits with an $F$-test on the sum of squared errors (``the exponential test''). This provides an initial stringent check on the convergence behavior which we perform until exponential decay behavior is detected. In the second version, we fit just a linear function to the loss curve and perform a one-tailed $t$-test to detect whether the slope of the loss is significantly less than zero (a general decreasing trend, ``the linear test''). The standard variant of ExpTest combines these two: performing initial exponential testing, followed by linear testing on subsequent windows. For tasks with high gradient noise (for example, very large tasks with many classes), the exponential test may be too stringent. In these cases, we only perform linear testing. We note that the $F$- and $t$-test statistics serve here as operational decision rules on the loss-curve window rather than as strict inferential tests: within-window loss observations are serially correlated, so the nominal $\alpha$ acts as a threshold on the test statistic rather than a conventional Type-I error rate. We retain the $F$- and $t$-scales because they yield interpretable thresholds and remain stable across window lengths. The convergence result for ExpTest (Theorem~\ref{thm:convergence}) relies on the assumption that these tests trigger a learning rate decay with non-vanishing probability across windows; we formalize this as the $p_{\min}$ condition in Section~\ref{sec:convergence}.

 \subsubsection{Additional Variations} \label{sec:additional_variations}
We can further vary certain phases of the framework (and we explore this experimentally in the following section), for example, by choosing whether or not to re-initialize the model after Lipschitz upper bounding, or by choosing whether or not to re-initialize if the exponential test fails. These are minor design choices that can be particularly relevant for tasks such as fine-tuning, where the objective is to prevent significant deviation from the current model state.
 
\subsubsection{Summary}
Algorithm~\ref{alg:exptest} summarizes the standard version of ExpTest. The key design rationale is as follows. The covariance-based and Lipschitz-based estimates together provide an optimistic but bounded starting learning rate, enabling aggressive early exploration. The curvature-derived window size is designed so that each evaluation window is long enough to distinguish exponential decay from noise, while the correction factor $c_w$ adapts this window to the gradient noise level of mini-batch SGD. The two-phase testing strategy (exponential then linear) provides an initially stringent convergence check that relaxes once decay behavior is established, avoiding premature learning rate reduction. Unlike classical Armijo or Wolfe line-search procedures, which impose sufficient decrease conditions at every iteration, ExpTest tolerates non-monotonic loss trajectories and enforces stability only once convergence behavior is detected statistically over finite training windows \cite{nocedal2006numerical}.

The ExpTest framework introduces two hyperparameters: the significance level ($\alpha$), which controls the strictness of the convergence tests, and the decay factor ($\beta$), which governs the logarithmic learning-rate search. We use $\alpha=0.05$ and $\beta=0.33$ as default values, and show in Section~\ref{sec:cifar_robust_hp} that ExpTest performs consistently across reasonable variations in both parameters. We also extend to SGD with momentum \cite{POLYAK19641}.

\begin{algorithm}
\footnotesize
\setstretch{1.22}
\caption{ExpTest}
\label{alg:exptest}
\begin{algorithmic}[1]

\State \textbf{Input:} model $f(\theta)$, data stream $(\Vec{x}_t,\Vec{y}_t)$, confidence $\alpha$, decay factor $\beta$
\State \textbf{Compute} $\eta_{\max}$ from input covariance (method varies by architecture; see Section~\ref{sec:lr_upper})
\State \textbf{Initialize:} $\eta \leftarrow \eta_{\max}$; compute $\mathcal{L}_0$; set
\[
w \leftarrow \left\lfloor \frac{2\sqrt{2}\mathcal{L}_0}{\eta e} + \frac{1}{2}\right\rfloor
\]
\State $t_{\text{start}}\leftarrow 0$; $\texttt{exp\_phase}\leftarrow\textbf{false}$; $\texttt{did\_upper}\leftarrow\textbf{false}$

\While{training not terminated}

    \State \textbf{Reset window buffers} (losses; and, if SGD: $S_{\text{mag}},S_{\text{vec}}$; and if $\neg\texttt{did\_upper}$: Lipschitz stats)

    \State \textbf{Train through a nominal window (for accumulation)}
    \Statex \hspace{1em}Run SGD for $w$ steps at LR $\eta$, recording losses
    \Statex \hspace{1em}If SGD: accumulate $S_{\text{mag}}=\sum\|g_t\|$ and $S_{\text{vec}}=\sum g_t$
    \Statex \hspace{1em}If $\neg\texttt{did\_upper}$: accumulate curvature estimates $\hat L$ over these $w$ steps

    \State \textbf{Compute corrected test horizon (every window, SGD only)}
    \Statex \hspace{1em}$c_w \leftarrow \frac{S_{\text{mag}}}{\|S_{\text{vec}}\|}$,\quad $w_{\text{full}}\leftarrow \left\lfloor c_w\,w+\frac{1}{2}\right\rfloor$
    \Statex \hspace{1em}(If GD: $c_w\!=\!1$ and $w_{\text{full}}\!=\!w$)

    \State \textbf{(Optional) finish collecting losses to the corrected horizon}
    \Statex \hspace{1em}Continue training for $w_{\text{full}}-w$ more steps at LR $\eta$, recording losses

    \State \textbf{One-time Lipschitz upper-bounding (first window only)}
    \If{$\neg\texttt{did\_upper}$}
        \If{$2/\min(\hat L) > \eta$}
            \State $\eta \leftarrow 2/\min(\hat L)$; reinitialize model and optimizer at LR $\eta$
        \EndIf
        \State Recompute window size $w$ (using updated $\eta$ if changed)
        \State $\texttt{did\_upper}\leftarrow\textbf{true}$
        \State \textbf{continue} \Comment{restart window collection under final initial LR}
    \EndIf

    \State \textbf{Convergence testing on the full window losses}
    \If{$\neg\texttt{exp\_phase}$}
        \State $p \leftarrow \textsc{ExponentialTest}(\text{losses }[t_{\text{start}}:t])$
        \If{$p < \alpha$}
            \State $\texttt{exp\_phase}\leftarrow\textbf{true}$
        \Else
            \State $\eta \leftarrow \beta\eta$; update optimizer; recompute $w$
        \EndIf
    \Else
        \State $p \leftarrow \textsc{LinearTest}(\text{losses }[t_{\text{start}}:t])$
        \If{$p > \alpha$}
            \State $\eta \leftarrow \beta\eta$; update optimizer; recompute $w$
        \EndIf
    \EndIf

    \State $t_{\text{start}}\leftarrow t$

\EndWhile

\end{algorithmic}
\end{algorithm}

\subsection{ExpTest Convergence} \label{sec:convergence}

\begin{theorem} \label{thm:convergence}
Let $\beta \in (0,1)$ be the multiplicative decay factor and suppose there exists $\Gamma \in \mathbb{N}$ and $p_{\min} > 0$ such that the probability of a convergence test triggering a learning rate decay satisfies $p_\kappa \ge p_{\min}$ for all windows $\kappa \ge \Gamma$. Then the ExpTest learning rate schedule $\{\eta_k\}$ satisfies, almost surely:
\begin{enumerate}
    \item $\eta_k < 2/L$ for all sufficiently large $k$,
    \item $\sum^{\infty}_{k=1}\eta_{k} = \infty$,
    \item $\sum^{\infty}_{k=1}\eta_{k}^2 < \infty$.
\end{enumerate}
\end{theorem}

\noindent \textit{Proof sketch.} The full proof is given in SM Section~\ref{sec:sm_convergence_proof}. Condition~1 follows from the monotonic decay of $\eta$: since ExpTest only decreases the learning rate, it almost surely falls below $2/L$ in finitely many windows. For Condition~2, the window size formula $w \propto 1/\eta$ ensures that the per-window contribution $w_\kappa \eta_\kappa$ is bounded below by a positive constant, so the step-wise sum diverges. For Condition~3, the same window size relation gives $w_\kappa \eta_\kappa^2 \le C\eta_\kappa + \frac{1}{2}\eta_\kappa^2$, reducing the problem to showing that the window-wise sum $\sum \eta_\kappa$ converges. This follows because the stochastic recursion $\eta_\kappa = \beta^{H_\kappa} \eta_{\kappa-1}$ yields geometric decay in expectation under the $p_{\min}$ assumption, and Tonelli’s theorem converts finite expected sum to almost sure finiteness. \hfill $\square$

The $p_{\min}$ condition has a simple form: it requires only that the convergence tests trigger a decay with non-vanishing probability, which is empirically supported by the observation that loss curves flatten over the course of training (i.e., there exist locally quadratic minima). This is also corroborated by standard SGD convergence results (SM Section~\ref{sec:sm_sgd_convergence}), which demonstrate a decreasing expected average gradient magnitude with increasing steps.

We emphasize what Theorem~\ref{thm:convergence} does and does not establish. The three conclusions are the classical step-size conditions of Robbins and Monro \cite{robbins1951stochastic}: the learning rate eventually falls below the stability threshold $2/L$, the step-size sum diverges (so SGD retains indefinite exploratory capacity), and the squared step-size sum converges (so variance contributions remain finite). These are conditions on the schedule $\{\eta_k\}$ alone. They do not by themselves constitute a convergence proof for the loss, and any downstream convergence claim for a specific objective inherits the standard regularity assumptions of the underlying stochastic-approximation result being invoked (Lipschitz-smoothness, lower-boundedness, bounded-variance gradient estimates; see SM Section~\ref{sec:sm_sgd_convergence}). The content of the theorem is that ExpTest's autonomous, test-driven schedule recovers these classical conditions without a hand-designed annealing sequence: a practitioner who would otherwise need to specify an initial learning rate and decay schedule to satisfy Robbins--Monro can instead run ExpTest and obtain a schedule that satisfies them almost surely under the explicit $p_{\min}$ condition.

\section{Experimental Results} \label{sec:experiments}
We evaluate variations of ExpTest on six different tasks (three image classification tasks, one regression task, one time series forecasting task, and one natural language processing task) with six different nonlinear network architectures (logistic regression, MLP, VGG-16, Informer, BERT, and ResNet-50). We compare ExpTest with SGD, SGD with momentum (0.9), Adam (default parameters), RMSprop (default parameters), and Adadelta (default parameters) \cite{POLYAK19641, kingma2017adammethodstochasticoptimization, RMSProp, zeiler2012adadeltaadaptivelearningrate}. For BERT, we compare to AdamW (Adam with decoupled weight-decay) \cite{loshchilov2019decoupledweightdecayregularization}. We include Adadelta as a direct comparison, as it similarly does not require a choice of initial global learning rate, though it is less widely adopted than the other methods due to slower convergence rates and worse final performance. We additionally compare ExpTest to three recent learning-rate-free methods--D-Adaptation, Prodigy, and DoG--on the CIFAR-10 task \cite{defazio2023dadaptation, mishchenko2024prodigy, ivgi2023dog}. Additionally, we examine ExpTest's behavior across choices of $\alpha$, $\beta$, and mini-batch size, and we combine ExpTest with momentum to assess its use as a learning rate search method in conjunction with other methods. We also perform ablation studies to investigate each facet of ExpTest. Code for experiments/figures and data generated by experiments are available at: \url{https://github.com/ZanChaudhry/ExpTest}. All code is in \texttt{Python}, using the \texttt{NumPy}, \texttt{SciPy}, \texttt{scikit-learn}, \texttt{PyTorch}, \texttt{Matplotlib}, and \texttt{HuggingFace} libraries \cite{10.5555/1593511, harris2020array, 2020SciPy-NMeth, scikit-learn, 10.5555/3454287.3455008, Hunter:2007, wolf2020huggingfacestransformersstateoftheartnatural}.

\subsection{Handwritten Digit Classification}
We begin with the MNIST handwritten digit classification task \cite{6296535}. We use this as a toy task for validating the simplest form of the ExpTest framework: 1) linear model LR upper-bounding; 2) window calculation/correction; 3) initial exponential testing for convergence detection; and 4) subsequent linear plateau detection. We train logistic regression models with a default mini-batch size of 32 and a fixed epoch limit of 5. For this simple task, we omit Lipschitz recalibration.

We calculate $\eta_{\text{max}}=0.0494$ on the MNIST dataset, as described in Algorithm~\ref{alg:exptest} for classification tasks, calculating the covariance matrix over the entire training dataset. We consider five learning rates over a logarithmic range: [$0.001\eta_{\text{max}}$, $0.01\eta_{\text{max}}$, $0.1\eta_{\text{max}}$, $\eta_{\text{max}}$, and $10\eta_{\text{max}}$]. We train with SGD, SGD with momentum, Adam, and RMSprop at each of these initial learning rates and compare to ExpTest ($\alpha = 0.05$, $\beta = 0.33$) and Adadelta without choosing a global initial learning rate. We perform five trials with different random initializations for each experiment. The training loss curves are shown in SM Section~\ref{sec:sm_mnist_lr} (moving averaged over 1 epoch, with loss curves above 0.4 excluded for clarity), and the test accuracies in Table~\ref{tab:mnist_accuracies}.

ExpTest displays the fastest initial convergence compared to all optimizers at all learning rates. Additionally, ExpTest achieves the minimum training loss compared to Adadelta and all optimizers with learning rate selection, except for Adam and RMSprop at $\eta = 0.01 \eta_{\text{max}}$. ExpTest achieves the maximum test accuracy compared to all optimizers, except the two previously mentioned, though it still produces comparable performance. All optimizers with learning rate selection (Adam, SGD, SGD w/ Mom., and RMSprop) display considerable variability in loss curve behavior and test accuracy over the learning rate range.

\begin{table*}[htbp]
    \centering
    \caption{Test Accuracies on MNIST Logistic Regression ($\mu \pm \sigma$, $n=5$)}
    \label{tab:mnist_accuracies}
    \begin{tabular}{@{}c|ccccc@{}} 
        \toprule
        Optimizer & $0.001\eta_{\text{max}}$ & $0.01\eta_{\text{max}}$ & $0.1\eta_{\text{max}}$ & $\eta_{\text{max}}$ & $10\eta_{\text{max}}$ \\ \midrule
        ExpTest & $\mathbf{92.30 \pm 0.07}$ & $92.30 \pm 0.07$ & $\mathbf{92.30 \pm 0.07}$ & $\mathbf{92.30 \pm 0.07}$ & $\mathbf{92.30 \pm 0.07}$ \\
        Adam & $91.45 \pm 0.07$ & $\mathbf{92.35 \pm 0.06}$ & $90.20 \pm 0.50$ & $88.92 \pm 0.76$ & $88.30 \pm 1.55$ \\
        SGD & $82.74 \pm 0.60$ & $89.86 \pm 0.16$ & $91.96 \pm 0.05$ & $91.65 \pm 0.29$ & $87.96 \pm 2.65$ \\
        SGD w/ Mom. & $89.86 \pm 0.09$ & $91.93 \pm 0.08$ & $91.94 \pm 0.28$ & $88.48 \pm 1.06$ & $87.62 \pm 1.25$ \\
        RMSprop & $91.53 \pm 0.04$ & $92.34 \pm 0.18$ & $89.77 \pm 1.69$ & $88.90 \pm 1.87$ & $88.71 \pm 1.01$ \\
        Adadelta & $91.46 \pm 0.35$ & $91.46 \pm 0.35$ & $91.46 \pm 0.35$ & $91.46 \pm 0.35$ & $91.46 \pm 0.35$ \\ \bottomrule
    \end{tabular}
\end{table*}

\subsection{Housing Price Regression} \label{sec:housing}
We test the same simple ExpTest framework as MNIST on a regression task: the California Housing Dataset \cite{KELLEYPACE1997291}. We train a fully connected network with three layers: $\mathbf{W_1}\in \mathbb{R}^{8 \times 32}$ ; $\mathbf{W_2}\in \mathbb{R}^{32 \times 32}$ ; and $\mathbf{W_3}\in \mathbb{R}^{32 \times 1}$ and ReLU activations. We chose this architecture based on experiments in SM Section~\ref{sec:sm_housing_model}. We split the dataset into 60\% training data, 20\% validation data, and 20\% test data. We train with full-batch gradient descent for a fixed epoch limit of 10000, saving the model with the best validation loss and evaluating it on the test set. We consider five learning rates over a logarithmic range: [$0.0001 \eta_{\text{max}}$, $0.001 \eta_{\text{max}}$, $0.01 \eta_{\text{max}}$, $0.1 \eta_{\text{max}}$, and $\eta_{\text{max}}$], calculating $\eta_{\text{max}}=0.987$ as in Algorithm~\ref{alg:exptest} for MSE loss. We use the standard values of ExpTest ($\alpha = 0.05$ and $\beta = 0.33$) and exclude Lipschitz recalibration once more due to network shallowness. Training loss curves are shown in SM Section~\ref{sec:sm_housing_train} ($y$-axis limited at 0.5 for clarity), and test MSE-losses are reported in Table~\ref{tab:cali_losses}. Validation loss curves are discussed in SM Section~\ref{sec:sm_housing_val}.

ExpTest displays faster initial convergence than Adadelta, but Adadelta achieves a lower minimum training loss. ExpTest achieves the best test performance at both high and low learning rate extremes. Interestingly, ExpTest displays the second slowest convergence for the middle three learning rate choices. Yet, ExpTest still possesses the second best test performance in these conditions, comparable to the optimal optimizer. As seen in previous work, the per-parameter adaptive optimizers tend to produce fast convergence but worse generalization performance than plain SGD-based methods \cite{10.5555/3294996.3295170}.

\begin{table*}[htbp]
    \centering
    \caption{Test MSE Losses on California Housing Fully Connected Network Regression ($\mu \pm \sigma$, $n=5$, *$n=4$)}
    \label{tab:cali_losses}
    \begin{tabular}{@{}c|ccccc@{}} 
        \toprule
        Optimizer & $0.0001\eta_{\text{max}}$ & $0.001\eta_{\text{max}}$ & $0.01\eta_{\text{max}}$ & $0.1\eta_{\text{max}}$ & $\eta_{\text{max}}$ \\ \midrule
        ExpTest & $\mathbf{0.2846 \pm 0.0070}$ & $0.2846 \pm 0.0070$ & $0.2846 \pm 0.0070$ & $0.2846 \pm 0.0070$ & $\mathbf{0.2846 \pm 0.0070}$ \\
        Adam & $0.2907 \pm 0.0048$ & $0.2881 \pm 0.0075$ & $0.3094 \pm 0.0190$ & $0.2990 \pm 0.0208$ & $1.3106 \pm 0.0000$ \\
        SGD & $0.7676 \pm 0.0409$ & $0.4488 \pm 0.0108$ & $0.3176 \pm 0.0094$ & $0.2858 \pm 0.0053$ & $2.0047 \pm 1.2023$* \\
        SGD w/ Mom. & $0.4492 \pm 0.0108$ & $0.3177 \pm 0.0091$ & $\mathbf{0.2686 \pm 0.0063}$ & $\mathbf{0.2712 \pm 0.0077}$ & $1.3106 \pm 0.0000$ \\
        RMSprop & $0.2893 \pm 0.0033$ & $\mathbf{0.2816 \pm 0.0039}$ & $0.3566 \pm 0.0386$ & $0.6217 \pm 0.3453$ & $1.5514 \pm 0.0126$ \\
        Adadelta & $0.3360 \pm 0.0122$ & $0.3360 \pm 0.0122$ & $0.3360 \pm 0.0122$ & $0.3360 \pm 0.0122$ & $0.3360 \pm 0.0122$ \\ \bottomrule
    \end{tabular}
\end{table*}

\subsection{Image Classification on CIFAR-10}
We test ExpTest combined with momentum on an additional classification task, the CIFAR-10 dataset, with a modified version of the VGG-16 deep convolutional neural network \cite{krizhevsky2009learning, SimonyanZ14a}. We also use this task for ablation studies, showing the impact of each component of ExpTest on performance.

\subsubsection{Simple Comparison} \label{sec:cifar_simple}
\begin{figure}[htbp]
    \centering
    \begin{subfigure}[t]{0.324\textwidth}
        \centering
        \includegraphics[width=\linewidth]{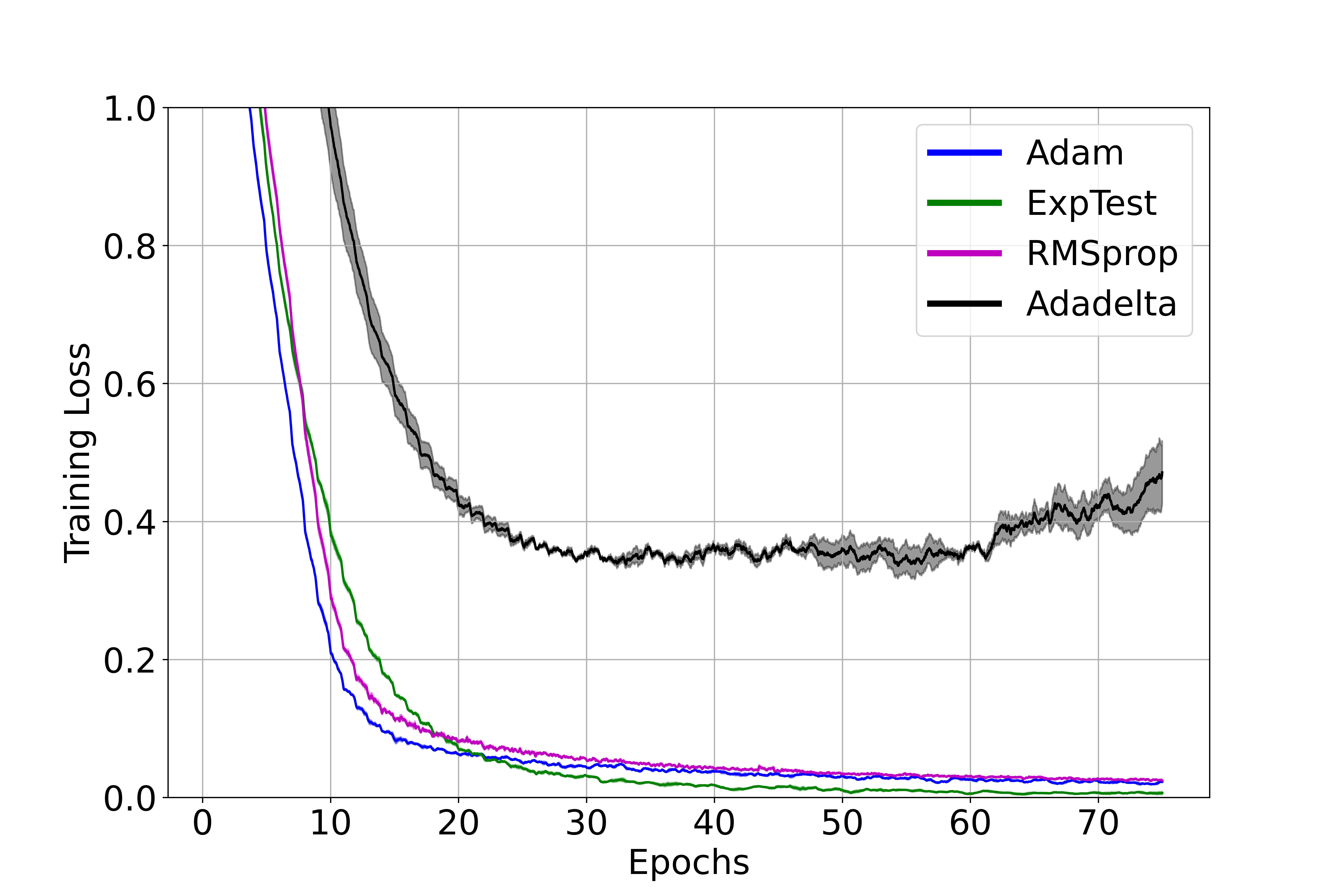}
        \caption{$\eta = 0.01 \eta_{\text{max}}$}
        \label{fig:cifar_sub1}
    \end{subfigure}
    \begin{subfigure}[t]{0.324\textwidth}
        \centering
        \includegraphics[width=\linewidth]{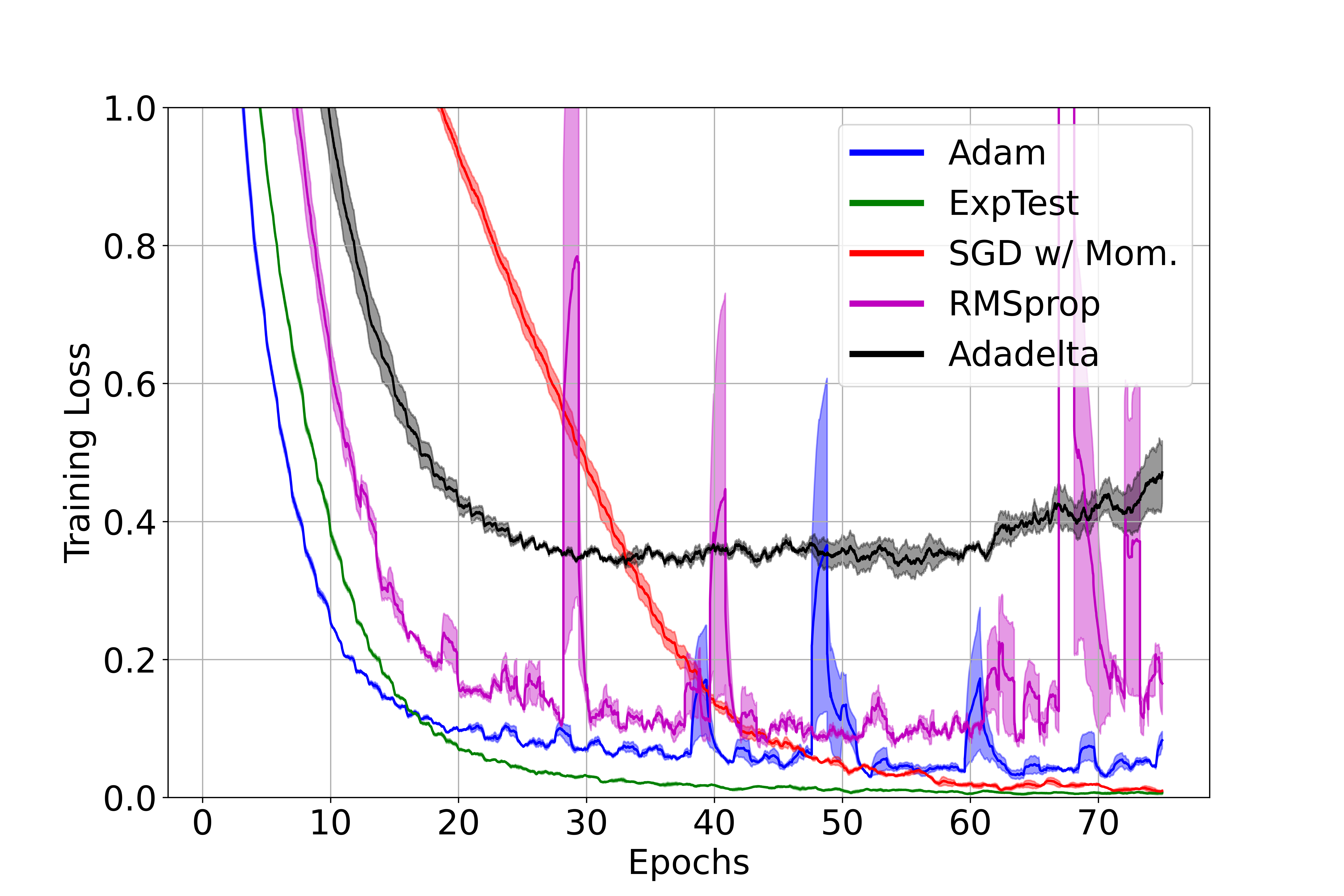}
        \caption{$\eta = 0.1 \eta_{\text{max}}$}
        \label{fig:cifar_sub2}
    \end{subfigure}
    \begin{subfigure}[t]{0.324\textwidth}
        \centering
        \includegraphics[width=\linewidth]{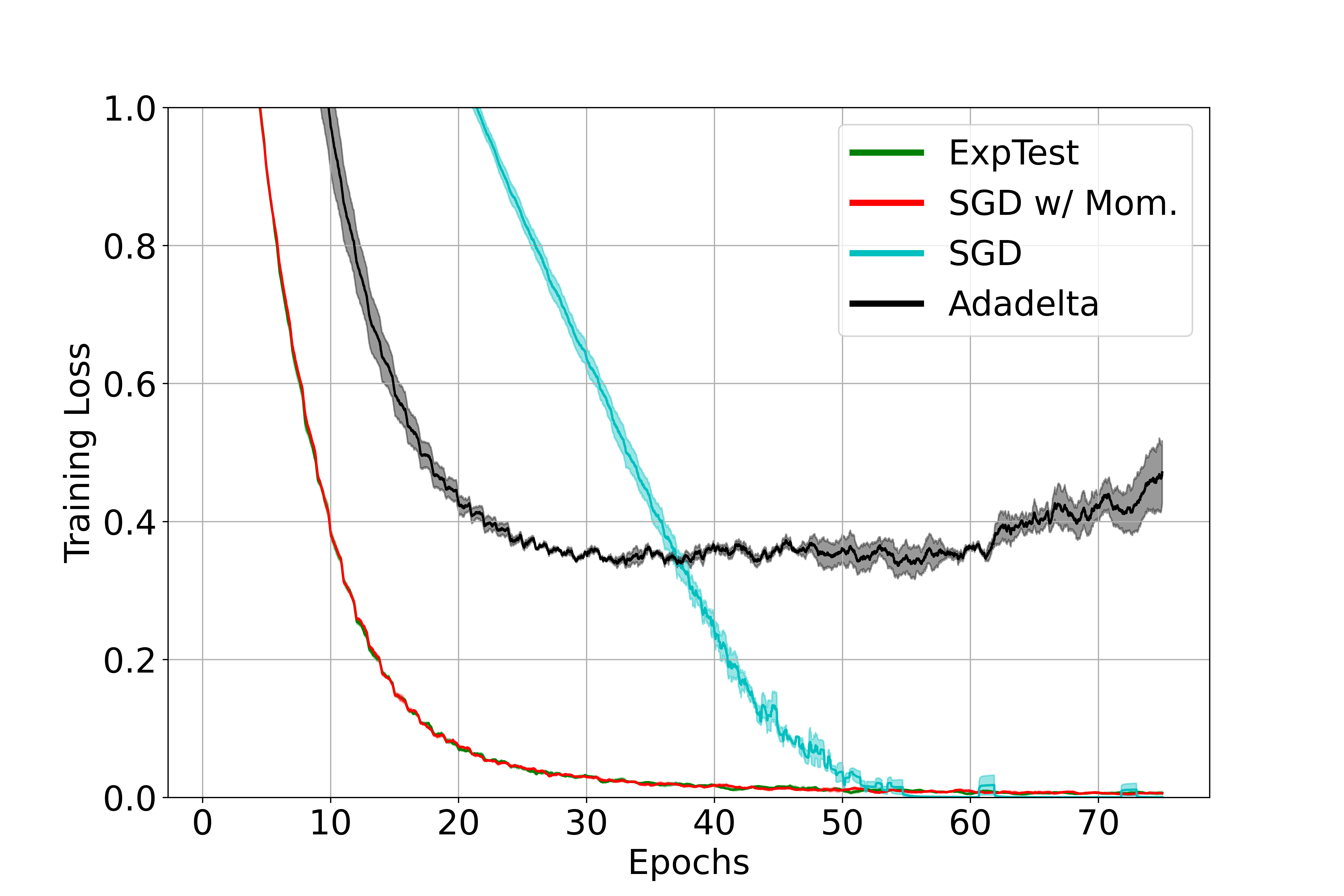}
        \caption{$\eta = \eta_{\text{max}}=0.00544$}
        \label{fig:cifar_sub3}
    \end{subfigure}
    \begin{subfigure}[t]{0.324\textwidth}
        \centering
        \includegraphics[width=\linewidth]{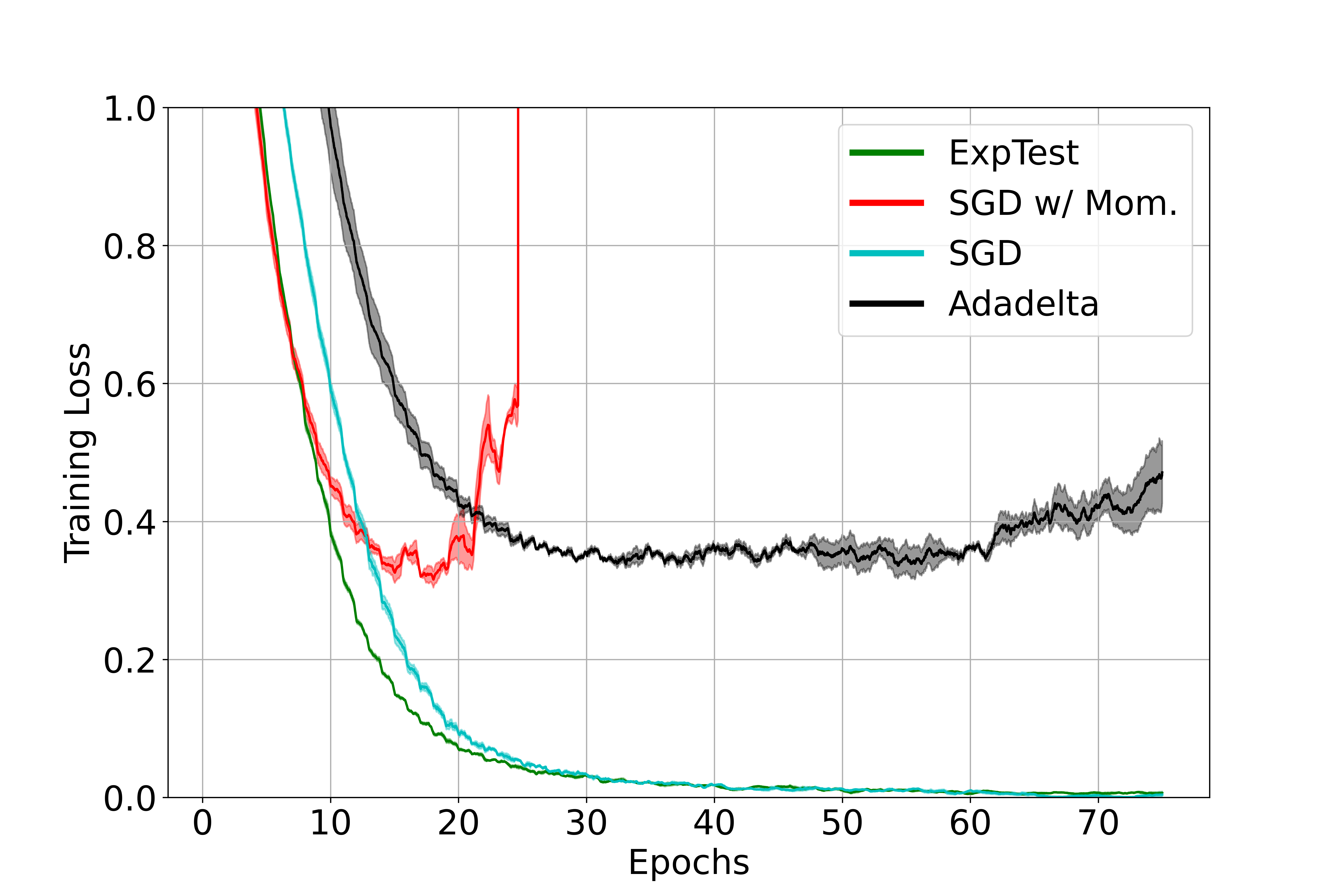}
        \caption{$\eta = 10 \eta_{\text{max}}$}
        \label{fig:cifar_sub4}
    \end{subfigure}
    \begin{subfigure}[t]{0.324\textwidth}
        \centering
        \includegraphics[width=\linewidth]{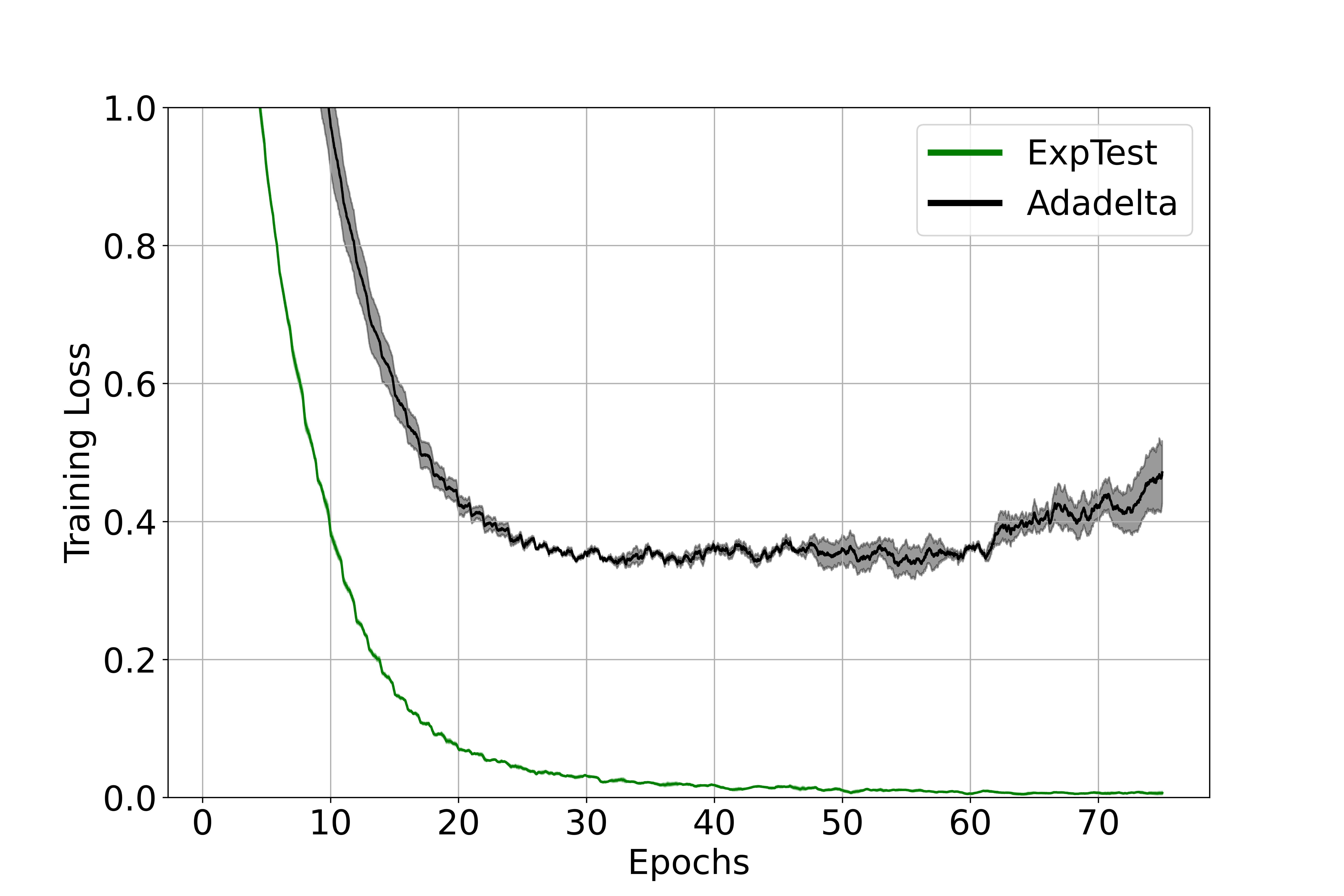}
        \caption{$\eta = 100 \eta_{\text{max}}$}
        \label{fig:cifar_sub5}
    \end{subfigure}
    \caption{Training loss curves for modified VGG-16 on CIFAR-10, moving averaged over window of 1 epoch. Center line shows mean of 3 trials with standard error shown by shading. $y$-axis is limited at 1.0 for clarity.}
    \label{fig:cifar_lr}
\end{figure}

We use a modified VGG-16 with a reduced classifier block to accommodate the $32 \times 32$ input size of CIFAR-10 (see code for architectural details). We train with mini-batch size of 128 for a fixed epoch limit of 75, evaluating on the test set at each epoch and reporting the best test accuracy. We consider five learning rates over a logarithmic range: [$0.01\eta_{\text{max}}$, $0.1\eta_{\text{max}}$, $\eta_{\text{max}}=0.00544$, $10 \eta_{\text{max}}$, and $100\eta_{\text{max}}$]. We use the standard values of ExpTest ($\alpha = 0.05$ and $\beta = 0.33$), and this time we incorporate a momentum of 0.9. Training loss curves are displayed in the five panels of Figure~\ref{fig:cifar_lr}, and test accuracies are reported in Table~\ref{tab:cifar_accuracies}. The plots in Figure~\ref{fig:cifar_lr} are y-axis limited at 1.0 to isolate extreme values and focus on convergence behavior.

ExpTest with momentum produces the highest test accuracy of all optimizers at all learning rate settings, though Adam at $0.1 \eta_{\text{max}}$ gives essentially the same performance (Table~\ref{tab:cifar_accuracies}). ExpTest with momentum achieves the minimum training loss in all conditions. Adam and RMSprop with well-chosen learning rates display faster initial convergence but plateau at a higher training loss. ExpTest with momentum and SGD with momentum produce essentially the same training curve for a well-chosen learning rate, though the learning rate decay of ExpTest produces a marginal improvement in test accuracy. Interestingly, though Adadelta displays high training loss, it produces highly competitive test accuracy. Furthermore, both SGD and SGD with momentum display convergent behavior at learning rates above the upper bound ($10\eta_{\text{max}}$), though SGD with momentum only converges initially before diverging. As described in Section~\ref{sec:revising_bound} (and in past work), there can be implicit regularization in DNNs, which we see can play a role in relaxing the upper bound learning rate by improving the conditioning of the minimization \cite{neyshabur2017implicitregularizationdeeplearning, martin2018implicitselfregularizationdeepneural, barrett2021implicit}. This motivated us to investigate Lipschitz upper-bounding, which produced much more dramatic performance improvements, in the following section.

\begin{table*}[htbp]
    \centering
    \caption{Test Accuracies on CIFAR-10 with Modified VGG-16 ($\mu \pm \sigma$, $n=3$)}
    \label{tab:cifar_accuracies}
    \begin{tabular}{@{}c|ccccc@{}} 
        \toprule
        Optimizer & $0.01\eta_{\text{max}}$ & $0.1\eta_{\text{max}}$ & $\eta_{\text{max}}$ & $10\eta_{\text{max}}$ & $100\eta_{\text{max}}$ \\ \midrule
        ExpTest w/ Mom. & $\mathbf{81.52 \pm 0.23}$ & $\mathbf{81.52 \pm 0.23}$ & $\mathbf{81.52 \pm 0.23}$ & $\mathbf{81.52 \pm 0.23}$ & $\mathbf{81.52 \pm 0.23}$ \\
        Adam & $73.45 \pm 0.11$ & $81.51\pm 0.35$ & $10.00 \pm 0.00$ & $10.00 \pm 0.00$ & $10.00 \pm 0.00$ \\
        SGD & $13.58 \pm 0.68$ & $38.49 \pm 3.24$ & $75.16 \pm 0.22$ & $79.83 \pm 0.48$ & $40.00 \pm 24.1$ \\
        SGD w/ Mom. & $41.72 \pm 1.38$ & $75.19 \pm 0.54$ & $81.42 \pm 0.10$ & $79.15 \pm 0.37$ & $10.00 \pm 0.00$ \\
        RMSprop & $71.14 \pm 0.70$ & $79.68 \pm 0.35$ & $20.40 \pm 4.13$ & $10.00 \pm 0.00$ & $10.00 \pm 0.00$ \\
        Adadelta & $79.76 \pm 0.31$ & $79.76 \pm 0.31$ & $79.76 \pm 0.31$ & $79.76 \pm 0.31$ & $79.76 \pm 0.31$ \\ \bottomrule
    \end{tabular}
\end{table*}

\subsubsection{Revised ExpTest Framework and Ablation Studies}
We implement Lipschitz upper-bounding and experiment with the various steps of the ExpTest framework in this section, using the same training settings as before on CIFAR-10 with modified VGG-16 and using SGD with momentum 0.9. We present results for the following variations:
\begin{itemize}
    \item Simple ExpTest (seen in previous sections)
    \item Standard ExpTest (including Lipschitz upper-bounding)
    \item ExpTest with model reset (*)
    \item ExpTest without exponential test 
    \item ExpTest without exp. test or Lipschitz upper-bound
    \item ExpTest without linear test
    \item ExpTest without linear test or Lipschitz upper-bound
\end{itemize}
In condition *, the model is reset (weights reinitialized) each time the exponential test fails until the test succeeds. This is the form used for the language model fine-tuning task in Section~\ref{sec:sst2}, as a safety mechanism to prevent considerable deviation from initialization. We present the test accuracies for each of these variations in Table~\ref{tab:cifar_ablation}.

\begin{table}[htbp]
    \centering
    \caption{Test Accuracies on CIFAR-10 with Modified VGG-16 for ExpTest Variants ($\mu \pm \sigma$, $n=3$)}
    \label{tab:cifar_ablation}
    \begin{tabular}{@{}c|c@{}} 
        \toprule
        Optimizer & Test Accuracy \\ \midrule
        Simple ExpTest & $81.52 \pm 0.23$ \\
        Standard ExpTest  & $\mathbf{83.44 \pm 0.40}$ \\
        Standard ExpTest w/ Reset & $\mathbf{83.44 \pm 0.40}$ \\
        Ablate Exp. Test & $83.36 \pm 0.44$ \\
        Ablate Exp. Test \& Lipschitz & $81.45 \pm 0.38$ \\
        Ablate Lin. Test & $81.35 \pm 0.48$ \\
        Ablate Lin. Test \& Lipschitz & $81.45 \pm 0.38$ \\
        Ablate All & $81.42 \pm 0.10$  \\
        \bottomrule
    \end{tabular}
\end{table}

The standard variant of ExpTest (with Lipschitz upper-bounding) produced consistent performance gains, producing a roughly 2\% increase in test accuracy over the simple variant (making it the best-performing method relative to those listed in Table~\ref{tab:cifar_ablation} by a reasonable margin). For this task, the addition of the model reset does not change performance, suggesting that the exponential test passes for each trial (and the model is thus never reset by the test). The ablation studies demonstrate the utility of each step of the framework. Interestingly, with the exponential test ablated, the test accuracy does drop very slightly (by 0.08\%), even though the previous experiment suggested that the exponential test passes anyway. This is likely due to minor randomness introduced by removing the test computations. Clearly for this task, the linear test and Lipschitz upper-bounding are essential to the performance gains introduced by ExpTest, as with either of them removed, performance drops down to roughly the level of plain SGD with momentum. Notably, the ``Ablate All'' condition (81.42\%) recovers the performance of plain SGD with momentum at its optimal learning rate (81.42\%, Table~\ref{tab:cifar_accuracies}), confirming that the ablation baseline matches the expected behavior of the underlying optimizer without any ExpTest components.

\subsubsection{Robustness to Hyperparameter Selection} \label{sec:cifar_robust_hp}
We train at three different values of $\alpha$: [0.1, 0.05, 0.01] and three different values of $\beta$: [0.05, 0.1, 0.33] with the standard ExpTest framework and the previously described training setup on CIFAR-10 with modified VGG-16 (Section~\ref{sec:cifar_simple}). The variability in performance across hyper-parameter choices for ExpTest is significantly less than for optimizers with learning rate selection across learning rate choices (Table~\ref{tab:cifar_accuracies}). We also show results over hyperparameter ranges on MNIST in SM Section~\ref{sec:sm_mnist_hp}.

\begin{table}[htbp]
    \centering
    \caption{Test Accuracies on CIFAR-10 with Modified VGG-16 Over Hyperparameters $\alpha$ and $\beta$ ($\mu \pm \sigma$, $n=3$)}
    \label{tab:cifar_hp_accuracies}
    \begin{tabular}{@{}c|ccc@{}} 
        \toprule
         & $\beta = 0.05$ & $\beta = 0.1$ & $\beta = 0.33$ \\ \midrule
         $\alpha = 0.1$ &$82.82 \pm 0.30$ &$83.01 \pm 0.31$ &$\mathbf{83.44 \pm 0.40}$ \\
         $\alpha = 0.05$ &$82.82 \pm 0.30$ &$83.01 \pm 0.31$ &$\mathbf{83.44 \pm 0.40}$ \\
         $\alpha = 0.01$ &$82.82 \pm 0.30$ &$83.01 \pm 0.31$ &$83.43 \pm 0.40$ 
         \\ \bottomrule
    \end{tabular}
\end{table}

\subsubsection{Robustness to Batch Size}
\begin{figure}[htbp]
    \centering
    \begin{subfigure}[t]{0.324\textwidth}
        \centering
        \includegraphics[width=\linewidth]{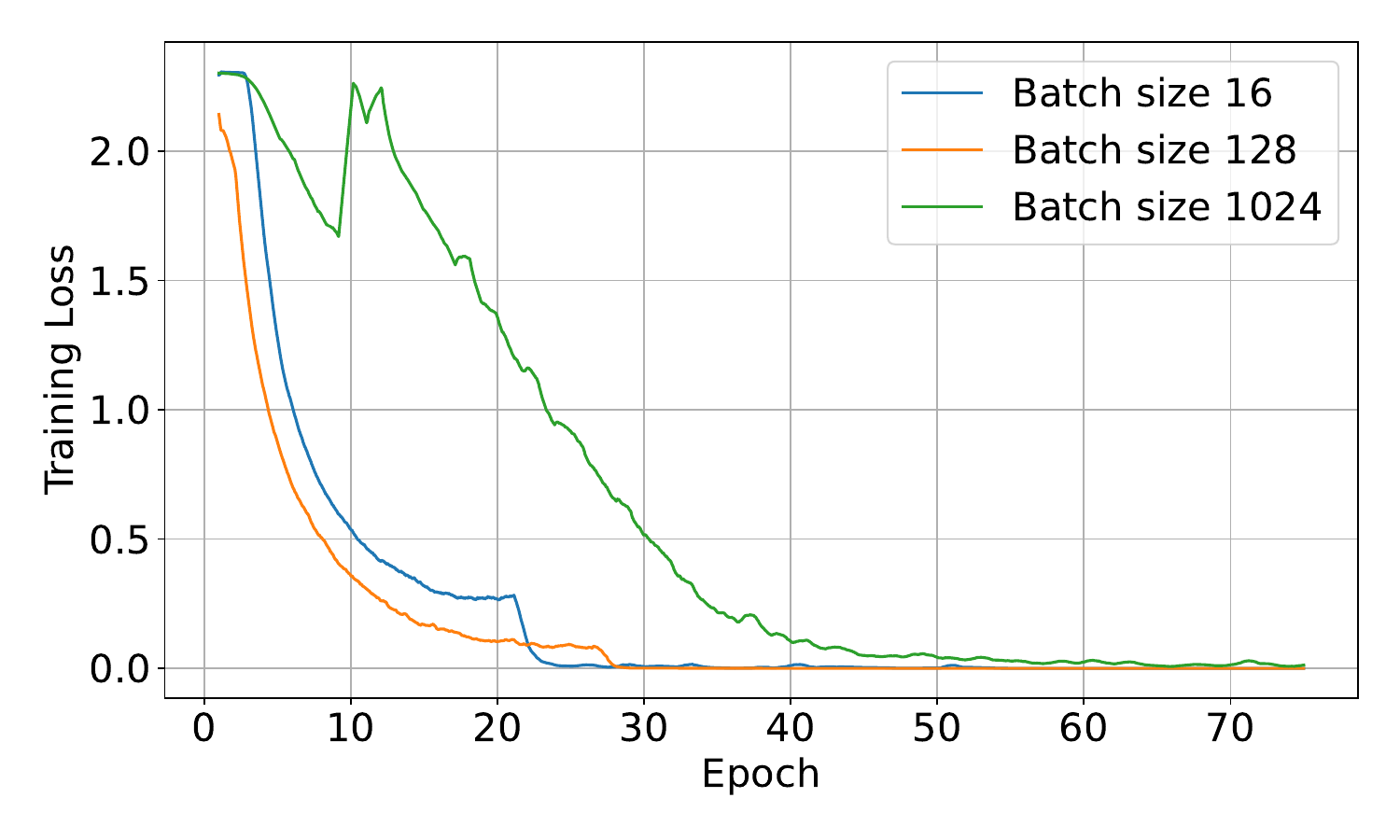}
        \caption{75 Epochs}
        \label{fig:cifar_bs_sub1}
    \end{subfigure}
    \begin{subfigure}[t]{0.324\textwidth}
        \centering
        \includegraphics[width=\linewidth]{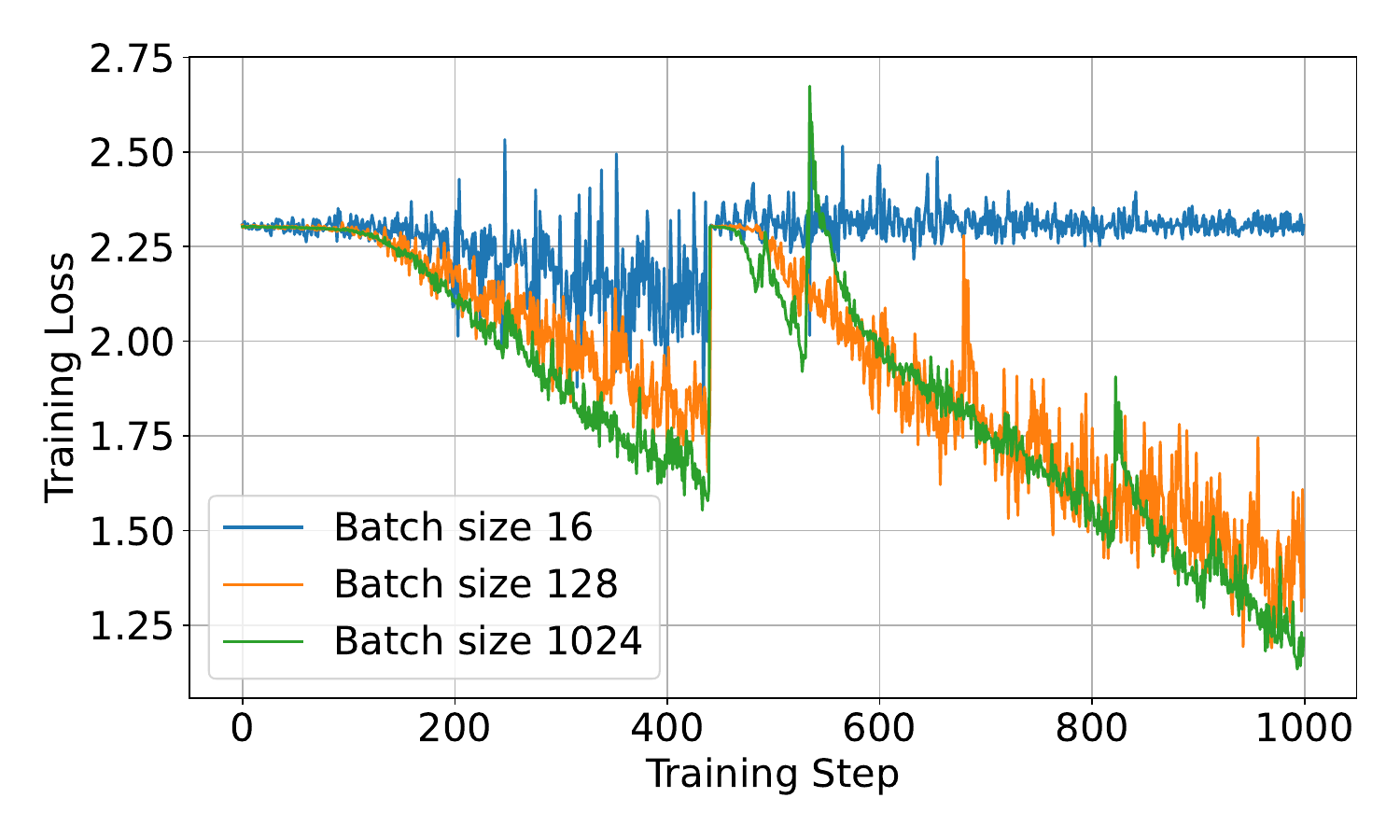}
        \caption{1000 Steps}
        \label{fig:cifar_bs_sub2}
    \end{subfigure}
    \caption{Single-run training loss curves for modified VGG-16 on CIFAR-10 at batch sizes 16, 128, and 1024; a) training over epochs, moving averaged over window of 1 epoch; b) training over first 1000 steps}
    \label{fig:bs_multi}
\end{figure}

We train at three different batch sizes: [16, 128, 1024], spanning three orders of magnitude, with the standard ExpTest framework and the previously described training setup on CIFAR-10 with modified VGG-16 (Section~\ref{sec:cifar_simple}) to examine the behavior of the ExpTest mechanism across batch sizes (though for batch size 1024, we train for 300 epochs instead of 75). The training loss curves are shown in Figure~\ref{fig:bs_multi}. When plotted over epochs (Figure~\ref{fig:bs_multi}a), all three batch sizes converge to similar loss values, though smaller batch sizes display noisier trajectories. When plotted over training steps (Figure~\ref{fig:bs_multi}b), larger batch sizes converge more rapidly per step, reflecting the reduced gradient noise from larger mini-batches. These results, together with the batch size experiments on MNIST (SM Section~\ref{sec:sm_mnist_bs}), indicate that ExpTest's window correction mechanism (Equation~\ref{eq:window_correction}) adapts to differing levels of gradient noise across batch sizes.

\subsubsection{Comparison to Modern Hyperparameter-Less Methods}
\begin{figure}[htbp]
    \centering
    \includegraphics[width=0.95\linewidth]{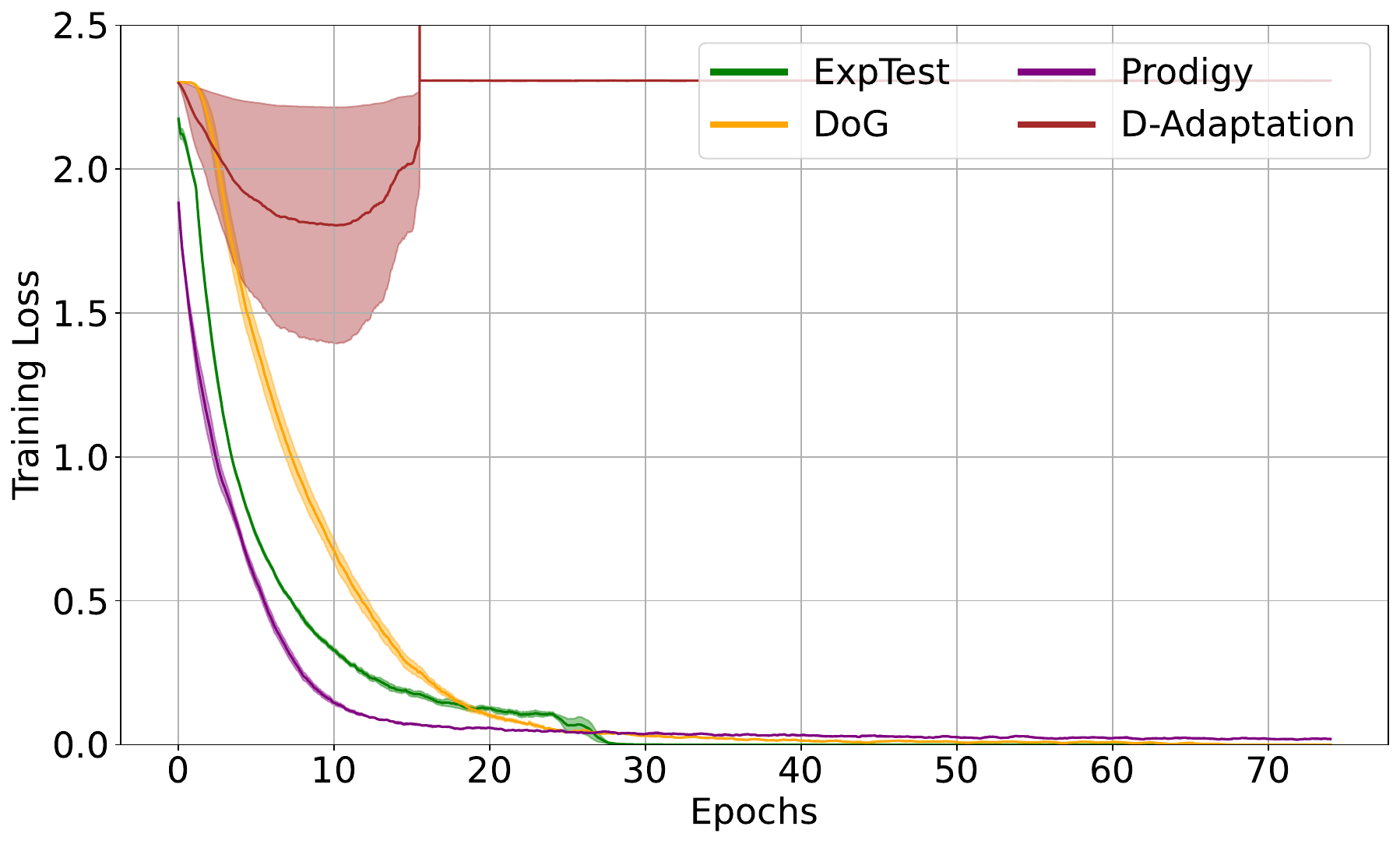}
    \caption{Training loss curves for modified VGG-16 on CIFAR-10 for Standard ExpTest and three learning-rate-free methods (D-Adaptation, Prodigy, DoG), moving averaged over window of 1 epoch. Center line shows mean of 3 trials with standard error shown by shading.}
    \label{fig:cifar_lrfree}
\end{figure}

We additionally compare the standard ExpTest framework to three recent learning-rate-free methods: D-Adaptation \cite{defazio2023dadaptation}, Prodigy \cite{mishchenko2024prodigy}, and DoG \cite{ivgi2023dog}. For D-Adaptation, we use the SGD variant (DAdaptSGD) with momentum 0.9, matching the ExpTest setup. Prodigy is only implemented as an Adam-like method in the reference implementation and is evaluated in that form. DoG is used with the polynomial-decay weight averager for evaluation, per the authors' recommended protocol. All three methods are used with their default parameters (no manual tuning), and all other experimental settings are held identical to the previous CIFAR-10 experiments. Test accuracies are reported in Table~\ref{tab:cifar_lrfree} and training loss curves are shown in Figure~\ref{fig:cifar_lrfree}.

Standard ExpTest achieves the highest test accuracy (83.44\%), followed by DoG (81.41\%) and Prodigy (77.89\%). D-Adaptation exhibits highly unstable behavior on this task: one of three trials reached 72.47\% test accuracy before diverging, while the remaining two trials collapsed to random-chance accuracy (10.00\%) within the first few epochs, yielding a mean of 30.82\% with a standard deviation of 29.45\%. This divergence is consistent with a known failure mode of D-Adaptation on non-convex problems, which partly motivated the subsequent development of Prodigy \cite{mishchenko2024prodigy}. Notably, of these parameter-free methods only ExpTest matches or exceeds the best hand-tuned SGD-based result (81.42\% for SGD with momentum at its optimal learning rate; Table~\ref{tab:cifar_accuracies}), indicating that ExpTest does not sacrifice performance to achieve hyperparameter-free operation in this comparison.

\begin{table}[htbp]
    \centering
    \caption{Test Accuracies on CIFAR-10 with Modified VGG-16 for Learning-Rate-Free Methods ($\mu \pm \sigma$, $n=3$)}
    \label{tab:cifar_lrfree}
    \begin{tabular}{@{}c|c@{}}
        \toprule
        Method & Test Accuracy (\%) \\ \midrule
        Standard ExpTest & $\mathbf{83.44 \pm 0.40}$ \\
        DoG & $81.41 \pm 0.35$ \\
        Prodigy & $77.89 \pm 0.62$ \\
        D-Adaptation & $30.82 \pm 29.45$ \\
        \bottomrule
    \end{tabular}
\end{table}

\subsection{Time Series Forecasting}
We evaluate ExpTest on a time series forecasting task using the Informer architecture on the ETTh1 (Electricity Transformer Temperature) dataset \cite{zhou2021informer}. We train the Informer model (multivariate setting, $d_{\text{model}}=512$, $n_{\text{heads}}=8$, $e_{\text{layers}}=2$, $d_{\text{layers}}=1$, $d_{\text{ff}}=2048$, dropout$=0.05$) with SGD and momentum 0.9, a mini-batch size of 32, gradient clipping (max norm 1.0), and a fixed epoch limit of 8. We use the standard values of ExpTest ($\alpha = 0.05$, $\beta = 0.33$) with Lipschitz upper-bounding. We compare across five prediction horizons (24, 48, 168, 336, 720) against the results reported in the original Informer paper \cite{zhou2021informer}. We perform three trials with different random seeds for each horizon and report both the mean $\pm$ standard deviation and the single best trial result. We note that the Informer paper reports results averaged over 5 runs with different train/validation splits; our results use 3 trials with different random seeds on a fixed split, so the comparison is approximate. Results are presented in Table~\ref{tab:ett_results}.

ExpTest achieves lower MSE and MAE than the published Informer results at four of the five prediction horizons (24, 48, 168, and 720), with the improvement most pronounced at shorter horizons. At the 336 horizon, the Informer paper reports superior performance. Notably, ExpTest achieves these results without any manual learning rate selection or scheduling, using only SGD with momentum as the base optimizer.

\begin{table*}[htbp]
    \centering
    \caption{Test MSE and MAE on ETTh1 Multivariate Forecasting with Informer. Informer results are the mean of 5 runs as reported in \cite{zhou2021informer}. ExpTest results: $\mu \pm \sigma$, $n=3$.}
    \label{tab:ett_results}
    \begin{tabular}{@{}cc|ccccc@{}}
        \toprule
        Metric & Method & 24 & 48 & 168 & 336 & 720 \\ \midrule
        \multirow{3}{*}{MSE} & Informer & $0.577$ & $0.685$ & $0.931$ & $\mathbf{1.128}$ & $1.215$ \\
        & ExpTest ($\mu \pm \sigma$) & $\mathbf{0.513 \pm 0.037}$ & $\mathbf{0.590 \pm 0.018}$ & $\mathbf{0.866 \pm 0.019}$ & $1.347 \pm 0.036$ & $\mathbf{1.198 \pm 0.056}$ \\
        & ExpTest (best) & $0.476$ & $0.579$ & $0.853$ & $1.308$ & $1.134$ \\ \midrule
        \multirow{3}{*}{MAE} & Informer & $0.549$ & $0.625$ & $0.752$ & $\mathbf{0.873}$ & $0.896$ \\
        & ExpTest ($\mu \pm \sigma$) & $\mathbf{0.506 \pm 0.016}$ & $\mathbf{0.575 \pm 0.008}$ & $\mathbf{0.741 \pm 0.006}$ & $0.945 \pm 0.022$ & $\mathbf{0.882 \pm 0.028}$ \\
        & ExpTest (best) & $0.490$ & $0.570$ & $0.734$ & $0.922$ & $0.850$ \\
        \bottomrule
    \end{tabular}
\end{table*}

\subsection{Image Classification on ImageNet}
We evaluate ExpTest on ImageNet (ILSVRC) classification with a ResNet-50 trained from scratch \cite{he2016deep}. We train with SGD and momentum 0.9, weight decay $10^{-4}$, cross-entropy loss with label smoothing 0.1, a mini-batch size of 1024, and a fixed epoch limit of 90. We use the standard values of ExpTest ($\alpha = 0.05$, $\beta = 0.33$) with Lipschitz upper-bounding. Standard data augmentation is applied (random resized crop to $224 \times 224$, random horizontal flip).

ImageNet presents unique challenges for the ExpTest framework due to the scale and structure of the task. First, the input images are high-dimensional ($224 \times 224 \times 3 = 150{,}528$ dimensions), yet the convolutional filters in ResNet-50 operate on small local receptive fields. Computing the covariance matrix over full images would yield eigenvalues reflecting global image statistics irrelevant to the local features extracted by convolutional layers, and would be computationally prohibitive. We therefore compute the covariance matrix over $16 \times 16$ patches of the input images, providing a learning rate bound that better reflects the data statistics encountered by the network's early layers, as discussed in Section~\ref{sec:lr_upper}. Second, the large number of classes (1000) introduces substantial gradient direction variability across mini-batches, as each batch covers only a small fraction of the class space. The standard window size formula (Equation~\ref{eq:window_size}) can produce windows spanning multiple epochs under these conditions, during which the accumulated gradient statistics become unreliable. We therefore cap the window size at a maximum of one epoch, ensuring the convergence tests remain responsive to changing training dynamics.

We compare against the original ResNet paper's single-crop result for ResNet-50 \cite{he2016deep}. Results are reported for a single training run and presented in Table~\ref{tab:imagenet_results}.

ExpTest achieves 75.73\% top-1 accuracy without any manual learning rate selection or predefined schedule. The original ResNet-50 result (79.26\%) uses a hand-tuned initial learning rate with a step-decay schedule. We note that our setup differs from the original paper in batch size (1024 vs.\ 256) and includes label smoothing, which was not used in the original work. Despite these differences, ExpTest closes the majority of the gap to the published result through fully autonomous learning rate management.

\begin{table}[htbp]
    \centering
    \caption{Top-1 Accuracy on ImageNet with ResNet-50 (single run)}
    \label{tab:imagenet_results}
    \begin{tabular}{@{}c|c@{}}
        \toprule
        Method & Top-1 Accuracy (\%) \\ \midrule
        ResNet-50 (He et al.) & $\mathbf{79.26}$ \\
        ExpTest & $75.73$ \\
        \bottomrule
    \end{tabular}
\end{table}

\subsection{Sentiment Classification} \label{sec:sst2}
We evaluate ExpTest on a natural language processing task: fine-tuning BERT-base-uncased on the SST-2 sentiment classification benchmark from GLUE \cite{devlin2019bert}. We train with SGD and momentum 0.9, a mini-batch size of 64, maximum sequence length of 128, and a fixed epoch limit of 5. For the initial learning rate bound, we compute the covariance matrix over the CLS token embeddings extracted from the pretrained BERT model across the training set, rather than over the raw input tokens, as the embeddings represent the continuous-valued input space to the classification head. We use the standard values of ExpTest ($\alpha = 0.05$, $\beta = 0.33$) with Lipschitz upper-bounding and model re-initialization on learning rate reduction during the search phase (as described in Section~\ref{sec:additional_variations}). We compare to AdamW across four learning rates spanning three orders of magnitude ($5 \times 10^{-4}$, $5 \times 10^{-5}$, $5 \times 10^{-6}$, $5 \times 10^{-7}$), where $5 \times 10^{-5}$ falls within the range recommended by Devlin et al.\ for BERT fine-tuning \cite{devlin2019bert, loshchilov2019decoupledweightdecayregularization}. Since the SST-2 test labels are not publicly available, we report performance on the GLUE validation split. We perform three trials with different random seeds. Results are presented in Table~\ref{tab:sst2_results}.

ExpTest matches AdamW at its optimal learning rate ($5 \times 10^{-5}$), with both methods achieving 92.39\% mean accuracy and 92.89\% best single-trial accuracy. However, AdamW exhibits extreme sensitivity to learning rate selection: performance ranges from 50.92\% at $5 \times 10^{-4}$ (no better than chance) to 92.39\% at $5 \times 10^{-5}$. ExpTest achieves this competitive performance without any manual learning rate selection, indicating its utility for fine-tuning tasks where the optimal learning rate is not known a priori. The use of model re-initialization during the search phase is particularly relevant for fine-tuning, as it prevents significant deviation from the pretrained weights during the initial learning rate search.

\begin{table}[htbp]
    \centering
    \caption{Validation Accuracy on SST-2 with BERT-base ($\mu \pm \sigma$, $n=3$)}
    \label{tab:sst2_results}
    \begin{tabular}{@{}c|cc@{}}
        \toprule
        Method & $\mu \pm \sigma$ (\%) & Best (\%) \\ \midrule
        ExpTest & $\mathbf{92.39 \pm 0.42}$ & $\mathbf{92.89}$ \\
        AdamW ($5 \times 10^{-4}$) & $50.92 \pm 0.00$ & $50.92$ \\
        AdamW ($5 \times 10^{-5}$) & $\mathbf{92.39 \pm 0.36}$ & $\mathbf{92.89}$ \\
        AdamW ($5 \times 10^{-6}$) & $92.24 \pm 0.30$ & $92.66$ \\
        AdamW ($5 \times 10^{-7}$) & $90.56 \pm 0.19$ & $90.83$ \\
        \bottomrule
    \end{tabular}
\end{table}

\subsubsection{Comparison to Modern Hyperparameter-Less Methods}
\begin{figure}[htbp]
    \centering
    \includegraphics[width=0.95\linewidth]{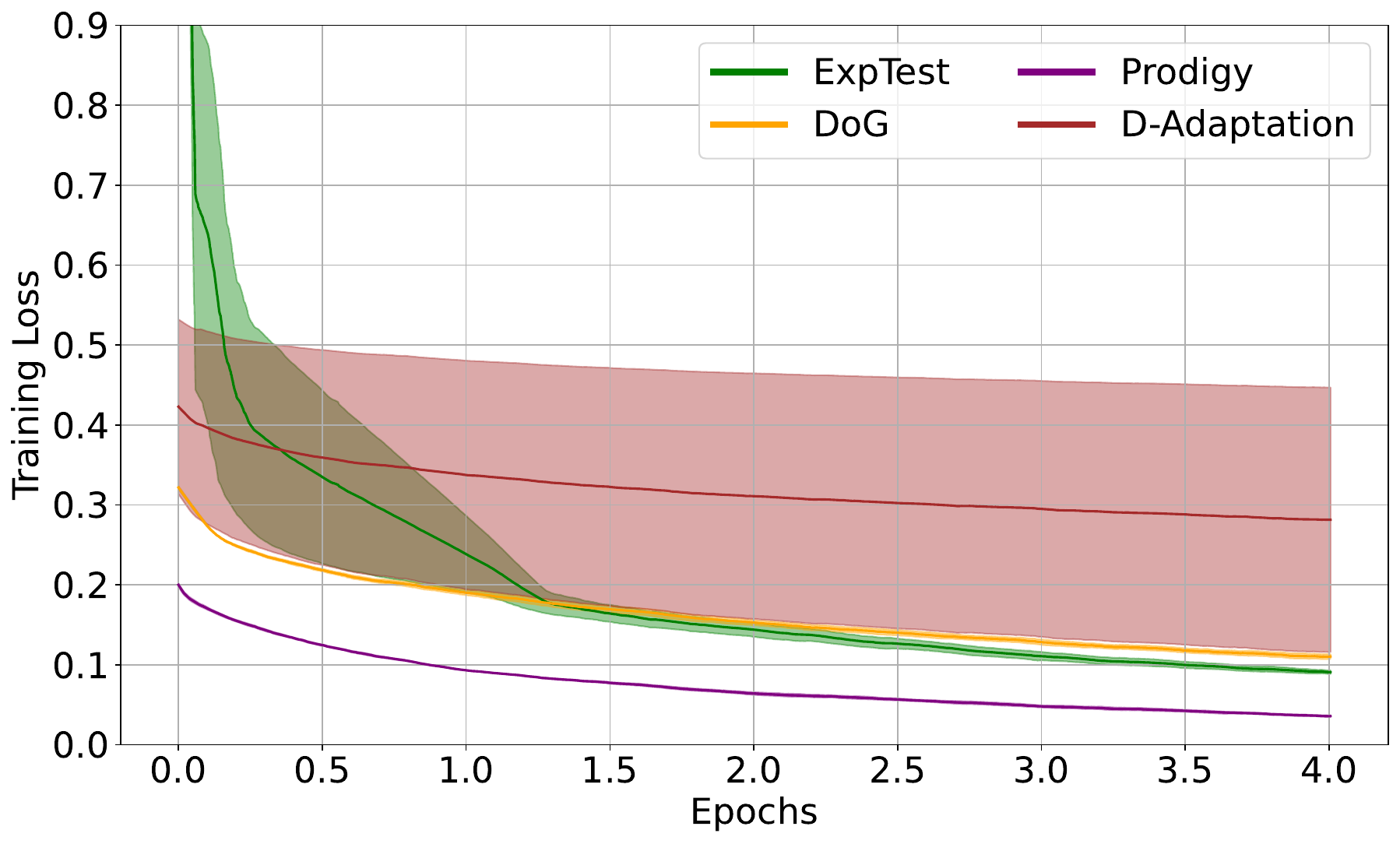}
    \caption{Training loss curves for BERT-base fine-tuning on SST-2 for Standard ExpTest and three learning-rate-free methods (D-Adaptation, Prodigy, DoG), moving averaged over window of 1 epoch. Center line shows mean of 3 trials with standard error shown by shading. The wide D-Adaptation band reflects one trial that collapsed to majority-class accuracy while the other two trials converged normally.}
    \label{fig:bert_lrfree}
\end{figure}

We additionally compare the standard ExpTest framework to the same three learning-rate-free methods considered for the CIFAR-10 experiments: D-Adaptation \cite{defazio2023dadaptation}, Prodigy \cite{mishchenko2024prodigy}, and DoG \cite{ivgi2023dog}. For D-Adaptation, we use the SGD variant (DAdaptSGD) with momentum 0.9. Prodigy is evaluated in its Adam-like form, with the authors' recommended \texttt{safeguard\_warmup=True} setting to limit early overshoot during fine-tuning \cite{mishchenko2024prodigy}. DoG is used with the polynomial-decay weight averager for evaluation, per the authors' recommended protocol. All three methods use an adapted-step multiplier of $1.0$ and their default remaining parameters, with all other experimental settings held identical to the preceding AdamW comparison. We note a protocol asymmetry that favors interpretability over uniformity: ExpTest's model re-initialization on learning rate reduction during the search phase (Section~\ref{sec:additional_variations}) has no direct analogue among the three baselines, so we report each method as its authors intend it to be used. Validation accuracies are reported in Table~\ref{tab:bert_lrfree} and training loss curves are shown in Figure~\ref{fig:bert_lrfree}.

Standard ExpTest and DoG achieve identical mean validation accuracies (92.39\%) with overlapping standard deviations, and both match the best-tuned AdamW baseline. Prodigy trails slightly (92.24\%) despite achieving the lowest training loss (Figure~\ref{fig:bert_lrfree}), indicating mild overfitting. D-Adaptation again exhibits unstable behavior: one of three trials collapsed to majority-class accuracy (50.92\%), while the remaining two trials matched the other methods (92.55\%), yielding a mean of 78.67\% with a standard deviation of 19.62\%. This replicates the failure mode observed on CIFAR-10 and is consistent with D-Adaptation's known sensitivity on non-convex problems, which partly motivated the development of Prodigy \cite{mishchenko2024prodigy}. On this fine-tuning task, ExpTest and DoG both match the best-tuned AdamW baseline, and were the only parameter-free methods to do so consistently across all trials.

\begin{table}[htbp]
    \centering
    \caption{Validation Accuracy on SST-2 with BERT-base for Learning-Rate-Free Methods ($\mu \pm \sigma$, $n=3$)}
    \label{tab:bert_lrfree}
    \begin{tabular}{@{}c|cc@{}}
        \toprule
        Method & $\mu \pm \sigma$ (\%) & Best (\%) \\ \midrule
        Standard ExpTest & $\mathbf{92.39 \pm 0.42}$ & $\mathbf{92.89}$ \\
        DoG & $\mathbf{92.39 \pm 0.35}$ & $\mathbf{92.89}$ \\
        Prodigy & $92.24 \pm 0.52$ & $92.66$ \\
        D-Adaptation & $78.67 \pm 19.62$ & $92.55$ \\
        \bottomrule
    \end{tabular}
\end{table}

\section{Discussion} \label{sec:discussion}
Across the six tasks considered, ExpTest is competitive with hand-tuned SGD-based baselines and recent learning-rate-free methods, while requiring no manual initial learning-rate selection or predefined schedule. On MNIST logistic regression and California Housing regression, ExpTest is competitive with the best-performing optimizer at its optimal learning rate across five learning rates spanning four orders of magnitude. On CIFAR-10 with modified VGG-16, standard ExpTest (83.44\%) exceeds the best hand-tuned SGD-with-momentum result (81.42\%) and outperforms DoG (81.41\%), Prodigy (77.89\%), and D-Adaptation (30.82\%, mean). On the Informer/ETTh1 forecasting task, ExpTest reduces MSE and MAE relative to the published Informer results at four of five prediction horizons. On ImageNet with ResNet-50 from scratch, ExpTest reaches 75.73\% top-1 accuracy without scheduling. On BERT fine-tuning for SST-2, ExpTest matches AdamW at its optimal learning rate (92.39\%) and matches the strongest learning-rate-free baseline among those tested (DoG), while AdamW itself spans 50.92\% to 92.39\% across its learning rate range. Taken together, these results suggest that ExpTest operates effectively across regression and classification, small and large scale, training-from-scratch and fine-tuning, and across fully-connected, convolutional, transformer-based, and pretrained architectures.

ExpTest occupies a distinct position among learning-rate-free methods. D-Adaptation, Prodigy, and DoG derive their step sizes from optimization-theoretic quantities (distance to the optimum, hypergradient feedback, or accumulated gradient norms) \cite{defazio2023dadaptation, mishchenko2024prodigy, ivgi2023dog}. ExpTest instead derives its schedule from the statistical structure of the loss curve, motivated by the exponential-decay behavior predicted under linearized network dynamics \cite{NEURIPS2018_5a4be1fa, misiakiewicz2023lectureslinearizedneuralnetworks}. Empirically, this distinction manifests in two ways. First, ExpTest does not exhibit D-Adaptation's known failure mode on non-convex problems \cite{mishchenko2024prodigy}, which we replicated on both CIFAR-10 and SST-2. Second, ExpTest matches DoG (the strongest of the parameter-free baselines we tested) on both of these tasks, while retaining the flexibility to operate with plain SGD or SGD with momentum as the base optimizer.

The framework performs consistently across the conditions we tested. Loss curves and final performance are stable across batch sizes spanning three orders of magnitude (Figure~\ref{fig:bs_multi}, SM Section~\ref{sec:sm_mnist_bs}), and performance varies little across the ExpTest hyperparameters $\alpha$ and $\beta$ (Table~\ref{tab:cifar_hp_accuracies}, SM Section~\ref{sec:sm_mnist_hp}). The observed variability under ExpTest across hyperparameter ranges is substantially smaller than the variability of conventional optimizers across learning rate ranges.

Several limitations warrant explicit discussion. We emphasize that ExpTest is not intended to displace extensively hand-tuned training recipes where maximum benchmark performance is the sole objective; rather, it targets the practical setting in which manual initial-learning-rate selection is burdensome or unavailable, and autonomous adaptation during training is valuable. First, the learning rate upper bound $\eta_{\text{max}}$ is derived from the eigenstructure of the input covariance matrix, which requires a pre-training pass over the training data and an eigendecomposition. For high-dimensional inputs, na\"ive covariance computation is prohibitive; for ImageNet we therefore computed the covariance over $16 \times 16$ patches of the input, a principled but task-specific adaptation tied to the local receptive fields of convolutional architectures. Second, because $\eta_{\text{max}}$ is computed once before training, ExpTest in its current form assumes an approximately stationary input distribution, and is not directly applicable to streaming, online, or continual-learning regimes in which the data distribution shifts during training. Third, the theoretical motivation is strongest in the linearized regime, exact for infinite-width networks at initialization, and increasingly approximate for narrower or deeper networks and for later phases of training. Lipschitz recalibration addresses departures from this regime with a principled but heuristic correction, and the empirical validation across diverse architectures complements rather than replaces the theoretical analysis. Fourth, ExpTest manages only the global learning rate; momentum coefficients, weight decay, and adaptive-optimizer second-moment parameters remain user-specified, and the framework does not currently compose with explicit schedulers such as warmup or cosine annealing.

These limitations also delineate clear directions for extending the framework. The hypothesis-testing approach (detecting theoretically-motivated convergence signatures in a real-time signal) is not intrinsic to the learning rate, and the same machinery can in principle be applied to the autonomous management of momentum, weight decay, or adaptive-optimizer coefficients, yielding a unified, signal-driven approach to hyperparameter-free training. Composition with known-effective schedulers is a second natural direction: warmup phases could precede the ExpTest search \cite{goyal2017accuratelargeminibatchsgd}, and outer-loop cosine or step-decay schedules could sit above ExpTest's tests \cite{loshchilov2017sgdr}, combining the reliability of established schedules at scale with ExpTest's autonomous reductions in response to loss-curve dynamics; the residual ImageNet gap to the hand-tuned ResNet-50 recipe (75.73\% vs.\ 79.26\%, albeit under differing batch size and regularization \cite{he2016deep}) suggests schedule-aware variants as a promising path for large-scale training from scratch. The strong performance observed for BERT fine-tuning on SST-2 further suggests the framework is well-matched to the fine-tuning of foundation models, a regime in which learning rate selection is particularly consequential \cite{mosbach2021stability} and where ExpTest's autonomous search avoids substantial departure from pretrained weights through its re-initialization mechanism. More broadly, data-adaptive strategies for $\eta_{\text{max}}$ computation (online covariance estimation, streaming eigendecomposition, or architecture-aware patching) would extend ExpTest to non-stationary and very-high-dimensional settings. Collectively, these directions position the statistical-signal approach introduced here as a foundation for broader autonomous training frameworks, continuing the progression toward deep neural network training that is effective, accessible, and free from manual hyperparameter tuning.

\section{Conclusion} \label{sec:conclusion}
We have presented ExpTest, a framework for autonomous learning rate management in deep neural network training. ExpTest combines a principled upper bound on the learning rate derived from linearized network theory, a window-corrected hypothesis test for exponential-decay convergence signatures, and a subsequent plateau-detection test that triggers learning rate reductions. The framework introduces two interpretable hyperparameters ($\alpha$ and $\beta$), to which training outcomes are consistent within the ranges tested, and eliminates the selection of the global initial learning rate entirely. Across six tasks spanning regression and classification, small and large scale, training-from-scratch and fine-tuning, and a diverse set of architectures, ExpTest matches or exceeds both hand-tuned SGD-based baselines and recent learning-rate-free methods. In removing the learning-rate-selection step, ExpTest addresses one of the most persistent barriers to the accessible and reproducible training of deep neural networks.

\section*{Acknowledgment}

The authors thank Sarah Hooper and Nicholas Asendorf for their comments, and Daniel Orozco, Pelayo-Alvarez-Brecht, Noam Rotenberg, and Dalia Fantini for discussions.

\ifCLASSOPTIONcaptionsoff
  \newpage
\fi

\bibliographystyle{IEEEtran}
\bibliography{bibtex/bib/lrpaper}

\clearpage

\setcounter{section}{0}
\renewcommand{\thesection}{\Alph{section}} 

\setcounter{figure}{0}
\renewcommand{\thefigure}{S\arabic{figure}}
\setcounter{table}{0}
\renewcommand{\thetable}{S\arabic{table}}
\setcounter{equation}{0}
\renewcommand{\theequation}{S\arabic{equation}}

\input{suppmat.tex}

\end{document}

%% file: suppmat.tex























\twocolumn[
\begin{center}
    \noindent{\LARGE Supplementary Material for ExpTest: Loss-Curve Hypothesis Testing for Autonomous Learning-Rate Selection in Deep Neural Networks}
    
    \vspace{1.5em}
    {\large Zan Chaudhry
        and Naoko Mizuno} \\
    \vspace{0.5em}
    {\small Z. Chaudhry and N. Mizuno were with the Laboratory of Structural Cell Biology, National Institutes of Health, Bethesda, MD, 20892 USA; Z. Chaudhry is now at the Ragon Institute, Cambridge, MA, 02139.}
\end{center}
\vspace{1em}]



\maketitle

\section{Mathematical Motivation} \label{sec:sm_math}
\subsection{The Linear Case} \label{sec:sm_linear}
We continue from Eq.~\ref{eq:gd_update} in the main text:
\begin{align} \label{eq:sm_gd_update}
\mathbf{T}[k+1]=\mathbf{T}[k] - \eta \frac{\partial\mathcal{L}}{\partial \mathbf{T}} \Biggr|_{\mathbf{T}=\mathbf{T}[k]}
\end{align}
Given:
\begin{align}
\mathcal{L}=\frac{1}{2ms}\text{tr}\Big(\Big(\mathbf{\hat{Y}}-\mathbf{Y}\Big)^{\text{T}}\Big(\mathbf{\hat{Y}}-\mathbf{Y}\Big)\Big)
\end{align}
and
\begin{align}
\mathbf{T} \mathbf{X} = \mathbf{\hat{Y}}
\end{align}

We set the condition that $\text{span}\{X\}=\mathbb{R}^{n}$. The condition ensures two facts: there exists a unique $\mathbf{T}$ that minimizes $\mathcal{L}$, and the sample covariance matrix for $X$ is invertible (which will soon become important).
Differentiating:
\begin{align} \label{eq:sm_gradient}
\frac{\partial \mathcal{L}}{\partial \mathbf{T}}=\frac{1}{ms}\Big(\mathbf{T}\mathbf{X}\mathbf{X}^\text{T}-\mathbf{Y}\mathbf{X}^\text{T}\Big)
\end{align}
Thus:
\begin{align} \label{eq:one}
\mathbf{T}[k+1] = \mathbf{T}[k]\mathbf{A} + \mathbf{B} 
\end{align}
Where we have made the substitution, $\mathbf{A} = \mathbf{I_n} - \frac{\eta}{ms} \mathbf{X}\mathbf{X}^\text{T}$, and $\mathbf{B}=\frac{\eta}{ms} \mathbf{Y}\mathbf{X}^\text{T}$.
Given the recursive form, we extrapolate the explicit:
\begin{align}
\mathbf{T}[k]=\mathbf{T}[0]\mathbf{A}^k
+ \sum_{i=0}^{k-1}\mathbf{B}\mathbf{A}^i
\end{align}
We then rewrite the series (having previously guaranteed the invertibility of $\mathbf{X}\mathbf{X}^\text{T}$):
\begin{align}
\sum_{i=0}^{k-1}\mathbf{B}\mathbf{A}^i &= \mathbf{B}\sum_{i=0}^{k-1}\mathbf{A}^i \nonumber \\
&= \mathbf{B}\Big(\mathbf{I_n} - \mathbf{A}\Big)^{-1}\Big(\mathbf{I_n} - \mathbf{A}^{k}\Big) \nonumber \\
&= \mathbf{B}\Big(\frac{\eta}{ms} \mathbf{X}\mathbf{X}^\text{T}\Big)^{-1}\Big(\mathbf{I_n} - \mathbf{A}^{k}\Big)  \nonumber \\
&= \mathbf{Y}\mathbf{X}^\text{T}\Big(\mathbf{X}\mathbf{X}^\text{T}\Big)^{-1}\Big(\mathbf{I_n} - \mathbf{A}^{k}\Big)
\end{align}
Returning:
\begin{align}
\mathbf{T}[k] &= \mathbf{T}[0]\mathbf{A}^k
+ \mathbf{Y}\mathbf{X}^\text{T}\Big(\mathbf{X}\mathbf{X}^\text{T}\Big)^{-1}\Big(\mathbf{I_n} - \mathbf{A}^{k}\Big) \nonumber \\
&= \Bigg(\mathbf{T}[0]
- \mathbf{Y}\mathbf{X}^\text{T}\Big(\mathbf{X}\mathbf{X}^\text{T}\Big)^{-1}\Bigg)\mathbf{A}^k + \mathbf{Y}\mathbf{X}^\text{T}\Big(\mathbf{X}\mathbf{X}^\text{T}\Big)^{-1}
\end{align}
Notice that $\mathbf{Y}\mathbf{X}^\text{T}(\mathbf{X}\mathbf{X}^\text{T})^{-1}$ is the exact solution for the $\mathbf{T}$ that minimizes $\mathcal{L}$, obtained by equating the derivative to the zero matrix. We thus substitute with $\mathbf{T_\infty} = \mathbf{Y}\mathbf{X}^\text{T}(\mathbf{X}\mathbf{X}^\text{T})^{-1}$:
\begin{align}
\mathbf{T}[k] = (\mathbf{T}[0]-\mathbf{T_\infty})\mathbf{A}^k + \mathbf{T_\infty}
\end{align}

We note that the function only converges in the case that $\mathbf{A}^k$ converges as $k \rightarrow \infty$. Thus, in convergence conditions, the eigenvalues of $\mathbf{A}$ must have magnitude less than one. We now introduce the condition that the input data is normalized, such that the mean of the vectors in $X$ is $\Vec{\mu}=\mathbf{0}$ and the variance along each dimension of the vectors in $X$ is $\sigma^2=1$. This simplifies the analysis and is common practice in machine learning. By definition, the sample covariance matrix of $X$ is:
\begin{align}
\mathbf{\hat{\Sigma}_{xx}}=\frac{1}{s-1}\sum_{i=1}^s \Big(\Vec{x}_i-\Vec{\mu}\Big)\Big(\Vec{x}_i-\Vec{\mu}\Big)^\text{T}
\end{align}
Given the normalization, this becomes:
\begin{align}
\mathbf{\hat{\Sigma}_{xx}}=\frac{1}{s-1}\sum_{i=1}^s \Vec{x}_i\Vec{x}_i^\text{T} = \frac{1}{s-1} \mathbf{X}\mathbf{X}^\text{T}
\end{align}
Thus, we rewrite $\mathbf{A}$ as $\mathbf{A} = \mathbf{I_n}-\frac{\eta (s-1)}{ms}\mathbf{\hat{\Sigma}_{xx}}$. By construction, the eigenvalues of $\mathbf{\hat{\Sigma}_{xx}}$ are all in the range: $(0, \text{ tr}(\mathbf{\hat{\Sigma}_{xx}}))=(0, n)$. We could define a more exact bound by calculating the eigenvalues and finding the maximum to give a revised range of: $(0, \lambda_{\text{max}})$. We have that the eigenvalues of $\mathbf{A}$ are given in terms of the corresponding eigenvalues of $\mathbf{\hat{\Sigma}_{xx}}$ by:
\begin{align}
\lambda_\mathbf{A} = 1 - \frac{\eta (s-1)}{m s}\lambda_{\mathbf{\Sigma}}
\end{align}

Now we arrive at a classic result \cite{10.5555/249049}. We can define two boundaries on the learning rate: one that requires no additional computation but is in general smaller than the optimal learning rate for the fastest convergence, and one that requires computing the eigenvalues of the sample covariance matrix but will guarantee the fastest possible convergence. In practice, if $n$ and $s$ are very large, it may be preferable to use the first bound to reduce computations. Starting with the first bound, we have the range of $\lambda_\mathbf{A}$ as: $(1-\frac{\eta n (s-1)}{m s}, 1)$. To guarantee convergence, the magnitude of the eigenvalues of $\mathbf{A}$ must be less than one; thus, the lower bound must be greater than negative one:

\begin{align}
1-\frac{\eta n (s-1)}{m s} &> -1 \nonumber \\
\therefore \text{ }\frac{2ms}{n(s-1)} &> \eta
\end{align}

Previous work studying the distribution of the maximum eigenvalue of Wishart random matrices has shown that it is unlikely for $\lambda_{\text{max}} \approx \text{tr}(\mathbf{\hat{\Sigma}_{xx}})$ \cite{CHIANI201469}. Thus, this bound is likely much lower than the true maximum learning rate, which is given by the second bound:
\begin{align}
\text{ }\frac{2ms}{\lambda_{\text{max}}(s-1)} > \eta
\end{align}

Now we proceed to diagonalize $\mathbf{A}$, with $\mathbf{A}=\mathbf{P}\mathbf{D}\mathbf{P}^{-1}$, where $\mathbf{D}$ is a diagonal matrix containing the eigenvalues of $\mathbf{A}$ along the diagonal. Substituting:
\begin{align}
\mathbf{T}[k] = (\mathbf{T}[0]-\mathbf{T_\infty})\mathbf{P}\mathbf{D}^k\mathbf{P}^{-1} + \mathbf{T_\infty}
\end{align}

Now we consider the elements of $\mathbf{T}[k]$. First, we rewrite the eigenvalues of $\mathbf{A}$ as: $\lambda_{\mathbf{A}, i} = \beta_i e^{-\alpha_i}$, where $\beta_i$ can only be $1$ or $-1$. Then we note that $\mathbf{D}^k$ has the form:
$$
\begin{bmatrix}
\beta_1 e^{-\alpha_1k} & 0 & \cdots & 0 \\
0 & \beta_2 e^{-\alpha_2k} & \cdots & 0 \\
\vdots & \vdots & \ddots & \vdots \\
0 & 0 & \cdots & \beta_n e^{-\alpha_nk}
\end{bmatrix}
$$
We compute $\mathbf{V}=(\mathbf{T}[0] - \mathbf{T_\infty})\mathbf{P}$. Then (absorbing the $\beta$'s into the $v$ constants):
$$
\mathbf{V}\mathbf{D}^k = 
\begin{bmatrix}
v_{11} e^{-\alpha_1k} & v_{12}e^{-\alpha_2k} & \cdots & v_{1n} e^{-\alpha_nk} \\
v_{21} e^{-\alpha_1k} & v_{22} e^{-\alpha_2k} & \cdots & v_{2n} e^{-\alpha_nk} \\
\vdots & \vdots & \ddots & \vdots \\
v_{m1} e^{-\alpha_1k} & v_{m2}e^{-\alpha_2k} & \cdots & v_{mn}e^{-\alpha_nk}
\end{bmatrix}
$$
Finally, we define $\mathbf{Q} = \mathbf{P}^{-1}$, giving:
$$\mathbf{V}\mathbf{D}^k \mathbf{Q}= 
\begin{bmatrix}
\sum_{i=1}^n v_{1i} q_{i1} e^{-\alpha_ik} &  \cdots & \sum_{i=1}^n v_{1i} q_{in} e^{-\alpha_ik} \\
\vdots & \ddots & \vdots \\
\sum_{i=1}^n v_{mi} q_{i1} e^{-\alpha_ik} &  \cdots & \sum_{i=1}^n v_{mi} q_{in} e^{-\alpha_ik} 
\end{bmatrix}
$$
Thus, the function $\mathbf{T}[k]$ is defined at each element of $\mathbf{T}$ by a linear combination of decaying exponentials plus the constant term, $\mathbf{T_\infty}$. Now we can characterize the behavior of the discretized loss function with a Taylor approximation:
\begin{align}
\Delta \mathcal{L} \approx \frac{\partial\mathcal{L}}{\partial \mathbf{T}} \cdot \Delta \mathbf{T}=  \frac{\partial\mathcal{L}}{\partial \mathbf{T}} \cdot - \eta \frac{\partial\mathcal{L}}{\partial \mathbf{T}} = - \eta \Bigg|\Bigg|\frac{\partial\mathcal{L}}{\partial \mathbf{T}}\Bigg|\Bigg| ^2 = - \eta ||\nabla\mathcal{L}|| ^2
\end{align}

The learning rate discretizes the loss function, so we can parameterize in terms of time: $\eta=\Delta t$. Returning:
\begin{align}
\Delta \mathcal{L} &\approx - \Delta t ||\nabla\mathcal{L}|| ^2 \nonumber \\
\frac{\Delta \mathcal{L}}{\Delta t} &\approx - ||\nabla\mathcal{L}|| ^2
\end{align}
For analysis purposes, we adopt the limit of infinitesimal step size, arriving at continuous time dynamics (``gradient flow''):
\begin{align}
\frac{d \mathcal{L}}{dt}= -||\nabla\mathcal{L}|| ^2
\end{align}

Now let us consider the form of $f(t)= ||\nabla \mathcal{L}(t)||^2$. We substitute the diagonalized expression of $\mathbf{T}(t)$ into the gradient, omitting the constant factors (they can be absorbed into $\mathbf{V}$):

\begin{align}
\nabla \mathcal{L}(t) &=  \mathbf{T}(t)\mathbf{X}\mathbf{X}^\text{T}-\mathbf{Y}\mathbf{X}^\text{T} \nonumber \\
&= (\mathbf{V} \mathbf{D}^t\mathbf{Q} + \mathbf{T_\infty})\mathbf{X}\mathbf{X}^\text{T}-\mathbf{Y}\mathbf{X}^\text{T} \nonumber \\
&= \mathbf{V} \mathbf{D}^t\mathbf{Q} \mathbf{X}\mathbf{X}^\text{T} + \mathbf{Y}\mathbf{X}^\text{T}(\mathbf{X}\mathbf{X}^\text{T})^{-1}\mathbf{X}\mathbf{X}^\text{T} -\mathbf{Y}\mathbf{X}^\text{T} \nonumber \\
&= \mathbf{V} \mathbf{D}^t\mathbf{Q} \mathbf{X}\mathbf{X}^\text{T}
\end{align}
Clearly, the elements of $\nabla \mathcal{L}(t)$ will be linear combinations of decaying exponentials of the form:
$$\sum_{i=1}^n C_i e^{-\alpha_i t}$$
Thus:
\begin{align}
||\nabla \mathcal{L}||^2 = \sum_{j=1}^{mn}\Bigg(\sum_{i=1}^n C_{ij} e^{-\alpha_it}\Bigg)^2
\end{align}
Returning to the differential equation:
\begin{align}
d\mathcal{L} &= -dt \sum_{j=1}^{mn}\Bigg(\sum_{i=1}^n C_{ij} e^{-\alpha_it}\Bigg)^2 \nonumber \\
\mathcal{L} &= C_{\text{int}} + \sum_{j=1}^{mn}\sum_{i=1}^n \sum_{h=1}^n \frac{C_{hj}C_{ij}}{\alpha_{hj} + \alpha_{ij}} e^{-(\alpha_i + \alpha_h) t}
\end{align}

Now we aim to approximate this sum of exponentials as a single exponential. In practice, we can solve numerically for a single exponential decay that fits the loss curve (a least-squares regression). However, we also provide two illustrative, analytical approximations. Consider the Taylor polynomial of a sum of exponentials, $A_1 e^{a_1x} + A_2 e^{a_2x} + ...$ For each $(A_i, a_i)$:
\begin{align}
A_i e^{a_ix} = \sum_{n=0}^\infty A_i a_i^n \frac{x^n}{n!}
\end{align}
Then for the sum:
\begin{align}
\sum_{i=1}^N A_i e^{a_ix} = \sum_{n=0}^\infty \Bigg(\sum_{i=1}^N A_i a_i^n\Bigg) \frac{x^n}{n!}
\end{align}
Approximated with the first two terms as:
\begin{align}
\sum_{i=1}^N A_i e^{a_ix} = \sum_{i=1}^N A_i + x \sum_{i=1}^N A_i a_i + R_1(x)
\end{align}
We can use the first two terms to fit a single exponential, $Ce^{cx}$:
\begin{align}
Ce^{cx} = C + xCc + R_1(x)
\end{align}
Equating the first two terms of each, we have:
\begin{align}
C &= \sum_{i=1}^N A_i \\
c &= \frac{1}{C}\sum_{i=1}^N A_i a_i \nonumber \\
&= \frac{\sum_{i=1}^N A_i a_i}{\sum_{i=1}^N A_i}
\end{align}
The error in the remainders is given by:
\begin{align}
\epsilon = \sum_{n=2}^\infty \Bigg(\sum_{i=1}^N A_i a_i^n - \frac{(\sum_{i=1}^N A_i a_i)^n}{(\sum_{i=1}^N A_i)^{n-1}} \Bigg)\frac{x^n}{n!}
\end{align}
In the case that $a_1=a_2=...=a_N$, we note that the error completely disappears. Thus, this is a good approximation in the case that the $a_i$ are similar. If they are dissimilar, we can assess the dominant term(s), noting that for $\sum_{i=1}^N A_i e^{-a_ix}$ if some $a_d \ll a_i \text{ }\forall \text{ }d \neq i$ then $e^{-a_dx} \ggg e^{-a_ix}$ as $x \rightarrow \infty$, and thus for large $x$:

\begin{align}
A_d e^{-a_dx} = C e^{-cx}\approx \sum_{i=1}^N A_i e^{-a_ix}
\end{align}

If some subset of the $a_i$'s, $\{ a_{d1}, a_{d2, ...}\}$ are similar in magnitude to each other but less than the remaining $a_i$'s, we can calculate the dominant term using our Taylor polynomial method for the $a_d$'s and then use this computed dominant term to approximate the entire sum. Therefore, $\mathcal{L}$ has the form:
$$\mathcal{L} = C_{\text{int}} + \sum_{j=1}^{mn}\sum_{i=1}^n \sum_{h=1}^n \frac{C_{hj}C_{ij}}{c_{hj} + c_{ij}} e^{-(\alpha_i + \alpha_h) t} \approx C_{\text{int}} + C e^{-ct}$$

We have demonstrated two approximation methods, depending on the values of the given $(A_i, a_i)$ pairs. In practice, a least-squares estimator of $(C_\text{int}, C, c)$ will balance these two extremes. Furthermore, any approximation of this sum as a single exponential is significantly improved by the guarantee of monotonicity of $\mathcal{L}$ (Equation~\ref{eq:gradient_flow} in the main text).

\subsection{Extension to Stochastic Gradient Descent} \label{sec:sm_sgd}
We have shown that in the linear case, where all of the input data is assembled as a matrix, $\mathbf{X}$, the discrete loss function is approximately exponential. However, in practice, $\mathbf{X}$ is a prohibitively large matrix, and thus the method of stochastic gradient descent (SGD) is used. We now present some classic results to extend the argument to SGD, where at each step, $k$, a random input vector $\Vec{x}$ is sampled from the set $X$ (and its associated output vector, $\Vec{y}$) and used for the gradient descent calculation. It is trivial to further extend the proof for mini-batch SGD by considering the expected value of the mini-batch mean gradient. We now demonstrate the same exponential decay-like behavior in the discrete loss function for SGD by considering the behavior of the expected value of $\mathbf{T}$: $\mathbb{E}(\mathbf{T}[k])$. We rewrite the gradient as:

\begin{align}
\frac{\partial \mathcal{L}}{\partial \mathbf{T}}=\frac{1}{m}\Big(\mathbf{T}\Vec{x}\Vec{x}^\text{T}-\Vec{y}\Vec{x}^\text{T}\Big)
\end{align}

Thus, following the same logic as before, but this time with $\mathbf{A} = \mathbf{I_n} - \frac{\eta}{m}\Vec{x}\Vec{x}^\text{T}$ and $\mathbf{B} = \frac{\eta}{m}\Vec{y}\Vec{x}^\text{T}$:
\begin{equation}
\mathbf{T}[k + 1] = \mathbf{T}[k]\mathbf{A} + \mathbf{B}
\end{equation}

Since at each time step, we sample a different $\Vec{x}$, $\mathbf{A}$ and $\mathbf{B}$ change as well. We denote the $k$-th $\mathbf{A}$ and $\mathbf{B}$ as $\mathbf{A_k}$ and $\mathbf{B_k}$. Then we can write $\mathbf{T}[k]$ explicitly as:
\begin{align}
\mathbf{T}[k] = \mathbf{T}[0] \Bigg(\prod_{i=1}^k \mathbf{A_i}\Bigg) +\mathbf{B_k}+ \sum_{i=1}^{k-1} \mathbf{B_i}\Bigg(\prod_{j=i+1}^k \mathbf{A_j}\Bigg)
\end{align}
However, if we assume the dataset is large ($s \rightarrow \infty$), then the sampling is independent, and if we apply expectation, we have:
\begin{align}
\mathbb{E}\Bigg(\prod_{i=1}^k \mathbf{A_i}\Bigg) =\prod_{i=1}^k\mathbb{E}( \mathbf{A_i}) =\mathbb{E}( \mathbf{A})^k
\end{align}
For $\mathbf{B}$, we have similarly: 
\begin{align}
\mathbb{E}\Bigg(\mathbf{B_i}\prod_{j=i+1}^k \mathbf{A_j}\Bigg)=\mathbb{E}(\mathbf{B_i})\prod_{j=i+1}^k \mathbb{E}(\mathbf{A_j})=\mathbb{E}(\mathbf{B})\mathbb{E}(\mathbf{A})^{k-i}
\end{align}
Thus, we begin by calculating $\mathbb{E}( \mathbf{A})$:
\begin{align}
\mathbb{E}( \mathbf{A}) &= \mathbb{E}\Big(\mathbf{I_n} - \frac{\eta}{m}\Vec{x}\Vec{x}^\text{T}\Big) \nonumber \\
&= \mathbf{I_n} - \frac{\eta}{m} \mathbb{E}\Big(\Vec{x}\Vec{x}^\text{T}\Big) \nonumber \\
&= \mathbf{I_n} - \frac{\eta}{m} \mathbf{\Sigma_{xx}}
\end{align}
Where we have used the definition of the covariance matrix, $\mathbf{\Sigma_{xx}} = \mathbb{E}(\Vec{x}\Vec{x}^\text{T})$. Then for $\mathbb{E}(\mathbf{B})$:
\begin{align}
\mathbb{E}(\mathbf{B}) &= \mathbb{E}\Big(\frac{\eta}{m}\Vec{y}\Vec{x}^\text{T}\Big) \nonumber \\
&= \frac{\eta}{m}\mathbf{\Sigma_{yx}}
\end{align}
Where $\mathbf{\Sigma_{yx}}$ is the cross-covariance matrix of $\Vec{y}$ with respect to $\Vec{x}$. Thus (omitting some of the steps used in subsection A):
\begin{align}
\mathbb{E}(\mathbf{T}[k]) = \Big(\mathbf{T}[0]-\mathbf{T_\infty}\Big)\Big(\mathbf{I_n} - \frac{\eta}{m}\mathbf{\Sigma_{xx}}\Big)^k + \mathbf{T_\infty}
\end{align}
From here, it clearly follows that the exponential decay behavior will be preserved in expectation, though with revised bounds on the learning rate.

\subsection{Nonlinearities and Classification Loss Functions} \label{sec:sm_nonlinear}
We extend the same principles to nonlinear networks (in linearized form). We return to the arbitrary depth neural network (Eq.~\ref{eq:layer} in the main text), $f(\theta, \Vec{x})$, where $\theta$ refers to the vectorized parameters of the network (we omit the vector arrow for $f$ and $\theta$  for clarity). We restate some of the equations here for clarity. Under gradient descent:
\begin{align}
\Delta \theta &= -\eta \frac{\partial \mathcal{L}}{\partial \theta} \nonumber \\
&= -\Delta t \frac{\partial \mathcal{L}}{\partial \theta} \nonumber \\
\frac{\Delta \theta}{\Delta t} &= -\frac{\partial \mathcal{L}}{\partial \theta}
\end{align}

In the continuous limit:
\begin{align}
\frac{d \theta}{d t} &= -\frac{\partial \mathcal{L}}{\partial \theta} \\
&= -\nabla_{\theta}f ^\text{ T } \nabla_{f}\mathcal{L} \\
\frac{df}{dt} &= \nabla_{\theta}f\frac{d \theta}{dt} \\
&= -\nabla_{\theta}f \text{ } \nabla_{\theta}f ^\text{ T }\nabla_{f}\mathcal{L}
\end{align}

If we adopt a linear approximation of $f$ around $\theta = \theta(0)$, then we arrive at the neural tangent kernel (NTK) description, which is an exact solution for infinite-width networks \cite{NEURIPS2018_5a4be1fa}. We substitute, $\mathbf{K} =\nabla_{\theta_0}f \text{ } \nabla_{\theta_0}f ^\text{ T}$ (notice that this is a positive semi-definite matrix):

\begin{align}
\frac{df}{dt} = -\mathbf{K}\nabla_{f}\mathcal{L}
\end{align}

Now we expand from the main text and show the solution for this equation under MSE loss and CE-loss (up to second order). For MSE-loss:
$$\mathcal{L}=\frac{1}{2m}\sum_{i=1}^{m} \Big(f_i-{y}_i\Big)^2=\frac{1}{2m}\Big(f-\Vec{y}\Big)^{\text{T}}\Big(f-\Vec{y}\Big)$$
Clearly:
\begin{align}
\nabla_f \mathcal{L} &= \frac{1}{m} \Big( f - \Vec{y}\Big) \\
&=  \frac{1}{m} \mathbf{I}\Big( f - \Vec{y}\Big) \\
&= \mathbf{C} \Big( f - \Vec{c}\Big)
\end{align}
Here, since $\mathbf{C}$ is a positive semi-definite matrix, we have the solution:
\begin{align}
f(t) = e^{-t\mathbf{KC}}\Big(f_0 - \Vec{c}\Big) + \Vec{c}
\end{align}

Now for CE-loss, let us consider a second-order approximation. For any arbitrary loss function, $\mathcal{L}$, a second order approximation around the initial output is given by:

\begin{equation}
\mathcal{L}_2 = \mathcal{L}_0 + \Big(\nabla_{f_0}\mathcal{L}\Big)^\text{T}(f-f_0) + \frac{1}{2} (f - f_0)^\text{ T }\nabla^2_{f_0}\mathcal{L_{\text{ }}}(f - f_0)
\end{equation}

Thus (for loss functions with a symmetric Hessian):

\begin{equation}
    \nabla_f \mathcal{L}_2 = \nabla_{f_0}\mathcal{L} + \nabla^2_{f_0}\mathcal{L_{\text{ }}}(f - f_0)
\end{equation}

For CE-loss:
\begin{equation}
    \mathcal{L} = -\sum_{i=1}^m y_i \log{f_i}    
\end{equation}

We have the gradient:

\begin{equation}
    \nabla\mathcal{L}=\Bigg[\frac{-y_1}{f_1} \cdots \frac{-y_m}{f_m} \Bigg]^\text{T}
\end{equation}
And the Hessian:
\begin{equation}
\nabla^2\mathcal{L}=
\begin{bmatrix}
y_1 / f_{1}^{\text{ }2} & 0 & ... & 0 \\
0 & y_2 / f_{2}^{\text{ }2} & ... & 0 \\
\vdots & \vdots & \ddots & \vdots \\
0 & 0 & ... & y_m / f_{m}^{\text{ }2}
\end{bmatrix}
\end{equation}

Now consider the product:
\begin{align}
-\nabla^2\mathcal{L_{\text{ }}}f &= -\begin{bmatrix}
y_1 / f_{1}^{\text{ }2} & 0 & ... & 0 \\
0 & y_2 / f_{2}^{\text{ }2} & ... & 0 \\
\vdots & \vdots & \ddots & \vdots \\
0 & 0 & ... & y_m / f_{m}^{\text{ }2}
\end{bmatrix} 
\begin{bmatrix}
f_{1}\\
f_{2}\\
\vdots\\
f_{m}
\end{bmatrix} \nonumber \\
&=
\begin{bmatrix}
-y_1/f_{1}\\
-y_2/f_{2}\\
\vdots\\
-y_m/f_{m}
\end{bmatrix} = \nabla \mathcal{L}
\end{align}

Putting this result into $\nabla_{f} \mathcal{L}_2$:
\begin{align}
\nabla_f \mathcal{L}_2 &= \nabla_{f_0}\mathcal{L} + \nabla^2_{f_0}\mathcal{L_{\text{ }}}(f - f_0) \nonumber \\
&= \nabla_{f_0}\mathcal{L} + \nabla^2_{f_0}\mathcal{L_{\text{ }}}f -\nabla^2_{f_0}\mathcal{L_{\text{ }}}f_0 \nonumber \\
&= -\nabla^2_{f_0}\mathcal{L_{\text{ }}}f_0 + \nabla^2_{f_0}\mathcal{L_{\text{ }}}f -\nabla^2_{f_0}\mathcal{L_{\text{ }}}f_0 \nonumber \\
&= \nabla^2_{f_0}\mathcal{L_{\text{ }}}f -2\nabla^2_{f_0}\mathcal{L_{\text{ }}}f_0 \nonumber \\
&= \nabla^2_{f_0}\mathcal{L_{\text{ }}} (f -2f_0) \nonumber \\
&= \mathbf{C}  \Big( f - \Vec{c}\Big)
\end{align}
And thus, we arrive at the same solution:
\begin{align}
f(t) = e^{-t\mathbf{KC}}\Big(f_0 - \Vec{c}\Big) + \Vec{c}
\end{align}

\section{Algorithm Development} \label{sec:sm_alg}
\subsection{Learning Rate Bounds for CE-Loss} \label{sec:sm_ce_bound}
We consider the training of a linear network, $f = \mathbf{T}\Vec{x}$, by gradient descent with a second order approximation of CE-loss. For tractability we treat $f$ directly as the probability-space output of the classifier; a full logit-space softmax+CE analysis would introduce softmax-Jacobian terms but yields a qualitatively similar scaling. The resulting estimate serves as an initial value and is subsequently refined during training by Lipschitz recalibration and the autonomous ExpTest schedule.

\begin{equation}
\mathcal{L}_2 = \mathcal{L}_0 + \Big(\nabla_{f_0}\mathcal{L}\Big)^\text{T}(f-f_0) + \frac{1}{2} (f - f_0)^\text{ T }\nabla^2_{f_0}\mathcal{L_{\text{ }}}(f - f_0)
\end{equation}

Differentiating with respect to $\mathbf{T}$:
\begin{equation}
\frac{\partial \mathcal{L}_2}{\partial \mathbf{T}} = \nabla_{f_0}\mathcal{L}\Vec{x}^\text{ T} + \nabla^2_{f_0}\mathcal{L}\mathbf{T} \Vec{x} \Vec{x}^\text{ T}
\end{equation}

This is of the same form as encountered previously in Eq.~\ref{eq:sm_gradient}, though this time with the Hessian included. We can upper bound the action of the Hessian due to the strong convexity of CE-loss. Thus: $ \nabla^2\mathcal{L} \preceq \lambda_{\text{max,}H} \mathbf{I_m}$, where $\lambda_{\text{max,}H}$ is the maximum eigenvalue of the Hessian. Therefore, to describe the convergence behavior, we can substitute as an upper bound:

\begin{equation}
\frac{\partial \mathcal{L}_2}{\partial \mathbf{T}} \approx \nabla_{f_0}\mathcal{L}\Vec{x}^\text{ T} + \lambda_{\text{max,}H} \mathbf{T} \Vec{x} \Vec{x}^\text{ T}
\end{equation}

We can solve for the convergence behavior by putting this expression into the gradient descent update rule (Eq.~\ref{eq:sm_gd_update}), giving a revised learning rate bound of:

\begin{equation}
\frac{2}{\lambda_{\text{max,}H} \lambda_{\text{max,} \Sigma}} > \eta
\end{equation}

Now we showed in Section~\ref{sec:sm_nonlinear} that the Hessian is a diagonal matrix, and thus its eigenvalues are the values along the diagonal, given by: $y_i/f_i^{\text{ }2}$. In the context of classification, $y_i$ can only be 1 or 0, and $f_i$ represents a probability in the range $(0, 1)$. Therefore, $\lambda_{\text{max,}H} > 1$ necessarily, and thus:

\begin{equation}
\frac{2}{\lambda_{\text{max,}H} \lambda_{\text{max,} \Sigma}} < \frac{2}{\lambda_{\text{max,} \Sigma}}
\end{equation}

The expression on the right is the MSE-loss upper bound for a linear network with output dimensionality 1, and it \textit{exceeds} the CE-loss upper bound on the left by a factor of $\lambda_{\text{max,}H}$. We therefore use $2/\lambda_{\text{max,} \Sigma}$ not as a strict stability bound for CE-loss, but as a convenient, architecture-agnostic initial estimate that over-approximates the CE-loss upper bound from above. As described in Section~\ref{sec:lr_upper} of the main text, this optimistic initial value is subsequently refined by Lipschitz recalibration during training and, in all cases, by the autonomous ExpTest schedule.

\section{Window Size Justification} \label{sec:sm_window}
Curvature is defined as:
\begin{equation}
    \kappa(t) = \frac{|f''(t)|}{(1+f'(t)^2)^\frac{3}{2}}
\end{equation}
For $f(t)=C_{\text{exp}}e^{-\eta \lambda t}$, we have:
\begin{equation}
\kappa(t) = \frac{|C_{\text{exp}} \eta^2 \lambda^2 e^{-\eta \lambda t}|}{(1+C_{\text{exp}}^{2} \eta^2 \lambda^2 e^{-2 \eta \lambda t})^\frac{3}{2}}
\end{equation}
Necessarily, $C_{\text{exp}} > 0$:
\begin{equation}
\kappa(t) = \frac{C_{\text{exp}} \eta^2 \lambda^2 e^{-\eta \lambda t}}{(1+C_{\text{exp}}^{2} \eta^2 \lambda^2 e^{-2 \eta \lambda t})^\frac{3}{2}}
\end{equation}

We can differentiate with respect to $t$ and set the derivative to zero, arriving at an expression for the time-point of maximum curvature:
\begin{equation}
t_{\text{max}} = \frac{\ln\left(\sqrt{2} \, C_{\text{exp}} {\eta}{\lambda}\right)}{{\eta}{\lambda}}
\end{equation}

We can differentiate this expression with respect to $\lambda$ and set the derivative to zero, arriving at an expression for the eigenvalue that maximizes the time-point of maximum curvature:
\begin{equation}
\lambda_{\text{max}} = \frac{\mathrm{e}}{\sqrt{2} \, C_{\text{exp}} {\eta}}
\end{equation}

Finally, we can input this value of $\lambda$ into the expression for $t_{\text{max}}$, arriving at an expression for the maximum possible time-point that maximizes the curvature:
\begin{equation}
t_{\text{max}} = \frac{\sqrt{2} C_{\text{exp}}}{e}
\end{equation}

We can set the initial loss, $\mathcal{L}_0$, as an upper bound for $C_{\text{exp}}$. Since the learning rate discretizes the time step, we convert to number of steps by dividing by $\eta$. Furthermore, we take a window of double $t_{\text{max}}$ to capture points both before and after the maximum curvature. Finally, we round to the nearest integer to avoid a fractional window size:
\begin{equation}
w = \Biggr\lfloor \frac{2 \sqrt{2}\mathcal{L}_0}{\eta e} + \frac{1}{2} \Biggr\rfloor
\end{equation}

\subsection{Review of SGD Convergence} \label{sec:sm_sgd_convergence}
We reestablish convergence conditions for SGD \cite{bottou2018}. Let us start with the following definitions:
\begin{itemize}
\item We have some scheme of sampling inputs (SGD or mini-batch SGD), such that there is a randomness, $\xi$, introduced
\item We aim to minimize a function: $f(\theta)=\mathbb{E}_{\xi}[F(\theta; \xi)]$
\item We have access to the noisy gradient estimates: $g(\theta_{k}, \xi_{k}) = \nabla F(\theta_{k}, \xi_{k})$
\end{itemize}
Then we have the following assumptions:
\begin{itemize}
\item Our gradient estimates are unbiased estimators: $\mathbb{E}_{\xi_{k}}[g(\theta_k, \xi_k)] = \nabla f(\theta_k)$
\item The gradient is Lipschitz continuous (Eq.~\ref{eq:lipschitz} in the main text) with Lipschitz constant $L$
\item The variance of our gradient estimates is bounded: $\mathbb{E}_{\xi_{k}}[||g(\theta_k, \xi_k) - \nabla f(\theta_k)||^2\text{ } | \text{ }\theta_k] \leq\sigma^2$ for $0<\sigma<\infty$
\item The function, $f$, has some global minimum value $f^*\geq0$
\end{itemize}

Now we start with a straightforward property given our assumptions \cite{bottou2018}:
\begin{equation}
f(\theta_{k+1}) \leq f(\theta_{k}) + \nabla f(\theta_{k})^\text{T}(\theta_{k+1}-\theta_{k}) + \frac{L}{2}||\theta_{k+1} -\theta_{k}||^2
\end{equation}
The gradient descent update is:
\begin{equation}
\theta_{k+1} = \theta_{k} - \eta_{k}g(\theta_k, \xi_k)
\end{equation}
Plugging in:
\begin{equation}
f(\theta_{k+1}) \leq f(\theta_{k}) -\eta_{k} \nabla f(\theta_{k})^\text{T}g(\theta_k, \xi_k) + \eta_{k}^2\frac{L}{2}||g(\theta_k, \xi_k)||^2
\end{equation}
Now we take conditional expectations:
\begin{equation}
\mathbb{E} [\nabla f(\theta_{k})^\text{T}g(\theta_k, \xi_k) \text{ }|\text{ }\theta_{k}]=||\nabla f(\theta_{k})||^2
\end{equation}
because $g$ is an unbiased estimator. For the next piece, we recall that the variance of our gradient estimates is bounded and expand:
\begin{align}
||g(\theta_k, \xi_k) - \nabla f(\theta_k)||^2 &= (g(\theta_k, \xi_k)- \nabla f(\theta_k))^{\text{T}}(g(\theta_k, \xi_k) \nonumber \\ &\,\,\,\,\,\,\,\,- \nabla f(\theta_k))\nonumber \\
&= ||g(\theta_k, \xi_k)||^2 -2\nabla f(\theta_k)^{\text{T}}g(\theta_k, \xi_k) \nonumber \\
&\,\,\,\,\,\,\,\,+ ||\nabla f(\theta_k))||^2
\end{align}
Now with conditional expectation:
\begin{align}
\mathbb{E}[||g(\theta_k, \xi_k) - \nabla f(\theta_k)||^2 \text{ | }\theta_{k}] &= \mathbb{E}[||g(\theta_k, \xi_k)||^2 \text{ | }\theta_{k}] \nonumber\\
&\,\,\,\,\,\,\,\, -2\nabla f(\theta_k)^{\text{T}} \nonumber \\
&\,\,\,\,\,\,\,\, \mathbb{E}[g(\theta_k, \xi_k) \text{ | }\theta_{k}] \nonumber \\
&\,\,\,\,\,\,\,\, + ||\nabla f(\theta_k))||^2 \nonumber \\
&= \mathbb{E}[||g(\theta_k, \xi_k)||^2 \text{ | }\theta_{k}] \nonumber \\
&\,\,\,\,\,\,\,\, -2|| \nabla f(\theta_k)||^2 \nonumber \\
&\,\,\,\,\,\,\,\, + ||\nabla f(\theta_k))||^2 \nonumber \\
&= \mathbb{E}[||g(\theta_k, \xi_k)||^2 \text{ | }\theta_{k}] \nonumber \\ 
&\,\,\,\,\,\,\,\, -|| \nabla f(\theta_k)||^2
\end{align}
Rearranging:
\begin{align}
\mathbb{E}[||g(\theta_k, \xi_k)||^2 \text{ | }\theta_{k}] &= || \nabla f(\theta_k)||^2 \nonumber \\ 
&\,\,\,\,\,\,\,\, + \mathbb{E}[||g(\theta_k, \xi_k) - \nabla f(\theta_k)||^2 \text{ | }\theta_{k}] \nonumber \\
&\leq || \nabla f(\theta_k)||^2 + \sigma^2
\end{align}
Now we can plug these results back in:
\begin{align}
\mathbb{E}[f(\theta_{k+1}) \text{ | }\theta_{k} ] &\leq f(\theta_{k}) -\eta_{k} ||\nabla f(\theta_{k})||^2 \nonumber \\ 
&\,\,\,\,\,\,\,\, + \eta_{k}^2\frac{L}{2}\Big(||\nabla f(\theta_{k})||^2+\sigma^2\Big) \nonumber \\
&\leq f(\theta_{k}) -\Bigg(\eta_{k}- \eta_{k}^2\frac{L}{2}\Bigg)||\nabla f(\theta_{k})||^2 \nonumber \\
&\,\,\,\,\,\,\,\, + \eta_{k}^2\frac{L}{2}\sigma^2
\end{align}
Now if we take full expectation:
\begin{align}
\mathbb{E}[f(\theta_{k+1})] &\leq \mathbb{E}[f(\theta_{k})] -\Bigg(\eta_{k}- \eta_{k}^2\frac{L}{2}\Bigg)\mathbb{E}[||\nabla f(\theta_{k})||^2] \nonumber \\ 
&\,\,\,\,\,\,\,\, + \eta_{k}^2\frac{L}{2}\sigma^2
\end{align}
Now if we sum this for all previous steps, $k$, up until $K$:
\begin{align}
\mathbb{E}[f(\theta_{K+1})] &\leq f(\theta_{1}) -\sum^{K}_{k=1} \Bigg(\eta_{k}- \eta_{k}^2\frac{L}{2}\Bigg)\mathbb{E}[||\nabla f(\theta_{k})||^2] \nonumber \\ 
&\,\,\,\,\,\,\,\, + \frac{L}{2}\sigma^2\sum^{K}_{k=1}\eta_{k}^2
\end{align}
Rearranging:
\begin{align} \label{eq:sm_sgd_bound}
\sum^{K}_{k=1} \Bigg(\eta_{k}- \eta_{k}^2\frac{L}{2}\Bigg)\mathbb{E}[||\nabla f(\theta_{k})||^2] &\leq f(\theta_{1}) - \mathbb{E}[f(\theta_{K+1})] \nonumber \\
&\,\,\,\,\,\,\,\,+ \frac{L}{2}\sigma^2\sum^{K}_{k=1}\eta_{k}^2 \nonumber \\
&\leq f(\theta_{1}) - f^*+ \frac{L}{2}\sigma^2\sum^{K}_{k=1}\eta_{k}^2
\end{align}
Analyzing this, we see that for the right-hand side to converge as we push $K\rightarrow \infty$, we must have: 
\begin{equation}
\sum^{\infty}_{k=1}\eta_{k}^2 < \infty
\end{equation}
Now on the left-hand side, we factor:
\begin{align}
\Bigg(\eta_{k}- \eta_{k}^2\frac{L}{2}\Bigg)\mathbb{E}[||\nabla f(\theta_{k})||^2] &= \eta_{k}\Bigg(1- \eta_{k}\frac{L}{2}\Bigg)\mathbb{E}[||\nabla f(\theta_{k})||^2] \nonumber \\
&\leq \eta_{k}\mathbb{E}[||\nabla f(\theta_{k})||^2] 
\end{align}
for $\eta \in (0, 2/L)$ (which is a condition from Section~\ref{sec:revising_bound} in the main text and from \cite{bottou2018}). Then:
\begin{align}
\sum^{\infty}_{k=1}\eta_{k}\mathbb{E}[||\nabla f(\theta_{k})||^2] \leq C
\end{align}
Now let's assume that there is some $M_G>0$ such that $\mathbb{E}[||\nabla f(\theta_{k})||^2 \geq M_G$ for all $k$ above some arbitrary $K_C$. Then:
\begin{equation}
\sum^{\infty}_{k=1}\eta_{k}\mathbb{E}[||\nabla f(\theta_{k})||^2] \geq M_G \sum^{\infty}_{k=1}\eta_{k}
\end{equation}
If:
\begin{equation}
\sum^{\infty}_{k=1}\eta_{k} < \infty
\end{equation}
then our convergence conditions are satisfied. However, then $\mathbb{E}[||\nabla f(\theta_{k})||^2 \geq M_G$, so we have an arbitrarily large minimum expected squared magnitude of the gradient (no form of convergence to a stationary point). Therefore, we set the condition that:
\begin{equation}
\sum^{\infty}_{k=1}\eta_{k} = \infty
\end{equation}
Now there can be no such $M_G$ if the sequence is to converge (which we know is required by the right-hand side). Thus, there is no $K_C$ above which $\mathbb{E}[||\nabla f(\theta_{k})||^2$ must stay positive, so there are infinitely many $k$ such that $\mathbb{E}[||\nabla f(\theta_{k})||^2=0$, and thus:
\begin{equation}
\lim_{k\rightarrow\infty} \inf{\mathbb{E}[||\nabla f(\theta_{k})||^2}=0
\end{equation}

\subsection{Proof of ExpTest Convergence (Theorem 1)} \label{sec:sm_convergence_proof}
We explicitly show that ExpTest almost surely achieves the same convergence as SGD with decreasing learning rate (lim inf convergence) under some light assumptions. We reestablish the convergence conditions for SGD in an illustrative manner in Section~\ref{sec:sm_sgd_convergence} as a helpful guide for this proof. We know from classic results (and as described in Section~\ref{sec:sm_sgd_convergence}) that there are three things we require of the learning rate for convergence that ExpTest must satisfy \cite{bottou2018}:
\begin{enumerate}
    \item $\eta \in (0,2/L)$
    \item $\sum^{\infty}_{k=1}\eta_{k} = \infty$
    \item $\sum^{\infty}_{k=1}\eta_{k}^2 < \infty$
\end{enumerate}
The first condition is not satisfied immediately; however, as long as there is sufficient learning rate decay, it will eventually be satisfied. Even if the initial learning rate is optimistic, ExpTest decays $\eta$ multiplicatively, so under the same mild assumption used below (that decays continue to occur with non-vanishing probability), $\eta_\kappa$ almost surely falls below $2/L$ after finitely many windows. Now for the second condition, consider the window size calculation:
\begin{align}
w &= \Bigg \lfloor \frac{2 \sqrt{2} \mathcal{L}_0}{\eta e} +\frac{1}{2}\Bigg \rfloor \\
&= \Bigg \lfloor \frac{C}{\eta} +\frac{1}{2}\Bigg \rfloor
\end{align}
We can set bounds:
\begin{equation}
 \frac{C}{\eta} -\frac{1}{2} < \Bigg \lfloor \frac{C}{\eta} +\frac{1}{2}\Bigg \rfloor \leq  \frac{C}{\eta} +\frac{1}{2}
\end{equation}
Now we must show that using the lower bound window size, the sum of learning rates diverges. Consider the sum over a single window, $w_{\kappa}$ from $k=\kappa$ to $k=\kappa +w_{\kappa}$. ExpTest holds learning rates constant over a window, thus we have:
\begin{equation}
S_{w_{\kappa} = }\sum^{\kappa + w_{\kappa}}_{k=\kappa} \eta_{k}=w_\kappa \eta_{\kappa}
\end{equation}
Now using the window size lower bound:
\begin{align}
S_{w_{\kappa}} &\ge \Bigg( \frac{C}{\eta_{\kappa}}-\frac{1}{2}\Bigg) \eta_{\kappa} \\
&\ge C-\eta_{\kappa}/2
\end{align}
Now, if we sum over infinitely many such windows ($\kappa$ being incremented by window sizes):
\begin{align}
\sum_{k=1}^{\infty} \eta_{k} \ge \sum^{\infty}_{\kappa=1} C-\eta_{\kappa}/2
\end{align}
The sum over the constant term alone certainly diverges. Now let us consider the window-wise $\eta$ term. We note that by design of ExpTest, $\eta_k$ can only decrease (excluding one initial Lipschitz recalibration window, which will only add a constant term). By the monotone convergence theorem (MCT):
\begin{equation}
\lim_{\kappa \rightarrow \infty}\eta_{\kappa}=\inf \eta_{\kappa}
\end{equation}
Now let us consider the structure of $\eta_{\kappa}$ as we go from window to window. At the start of each window, we have:
\begin{equation}
    \eta_{\kappa} = \begin{cases}
        \eta_{\kappa-1}, & \text{if } H=0\\
        \beta \eta_{\kappa-1}, & \text{if } H=1
        \end{cases}
\end{equation}
Here $H$ is the result of a hypothesis test, a binary indicator variable related to the accumulated loss data (and the signficance parameter, $\alpha$), and $0<\beta<1$ is the multiplicative decay constant. Let us assume that at the end of a given window, $w_{\kappa}$, we have the probability for the next window: $P(H=1 \text{ | } L_{w_{\kappa}})=p_{{\kappa}}$, where $L_{w_{\kappa}}$ is the discrete loss data observed over $w_{\kappa}$. Then we have:
\begin{align}
    \mathbb{E}\Big[\eta_{\kappa} \text{ }\Big|\text{ } L_{w_{\kappa}} \Big] &= \beta \eta_{\kappa-1}p_{\kappa} + \eta_{\kappa-1}(1-p_{\kappa}) \\
    &= \Big(1+ (\beta-1) p_{\kappa} \Big) \eta_{\kappa-1}
\end{align}
Taking full expectation:
\begin{align}
    \mathbb{E}[\eta_{\kappa}] &= \Big(1+ (\beta-1) p_{\kappa} \Big) \mathbb{E}[\eta_{\kappa-1}]\\
    &= \eta_{1}\prod^{\gamma = \kappa}_{\gamma=w_1} \Big(1+ (\beta-1) p_{\gamma} \Big)
\end{align}
Here, we have telescoped with $\gamma$ incremented by windows. Now let us assume that there is some $p_{\min}>0$, such that $\forall \gamma > \Gamma$, $p_{\gamma} > p_{\min}$. Then:
\begin{align}
\lim_{\kappa \rightarrow \infty} \mathbb{E}[\eta_{\kappa}] &\leq C_{\Gamma} \eta_{1} \prod^{\infty}_{\gamma = \Gamma} \Big(1+ (\beta-1) p_{\min} \Big) \\
&\leq \lim_{\tau \rightarrow \infty}C_{\Gamma} \eta_{1}  \Big(1+ (\beta-1) p_{\min} \Big)^{\tau}
\end{align}
Since $1\geq p_{\min}>0$ and $0<\beta<1$:
\begin{equation}
0 < \Big(1+ (\beta-1) p_{\min} \Big) < 1
\end{equation}
Thus:
\begin{equation}
\lim_{\tau \rightarrow \infty}C_{\Gamma} \eta_{1}  \Big(1+ (\beta-1) p_{\min} \Big)^{\tau} = 0
\end{equation}
Now since $\eta_{\kappa}$ is bounded below by zero, we must have the equality:
\begin{equation}
\lim_{\kappa \rightarrow \infty} \mathbb{E}[\eta_{\kappa}] = 0
\end{equation}
Now returning to the MCT:
\begin{equation}
\lim_{\kappa \rightarrow \infty} \mathbb{E}[\eta_{\kappa}] = \mathbb{E}\Big[\lim_{\kappa \rightarrow \infty} \eta_{\kappa}\Big] = \mathbb{E} [\inf \eta_{\kappa}]=0
\end{equation}
Moving the expectation out of the limit follows from applying Fatou's lemma and the MCT (consider creating the non-decreasing, non-negative sequence $H_{\kappa}=\eta_0 - \eta_{\kappa}$, taking the limit, and applying expectation). Now, since $\inf\eta_{\kappa}$ is non-negative and discrete, we have:
\begin{equation}
    \mathbb{E}[\inf \eta_{\kappa}] = 0=\sum^{\mathcal{K}} (\inf \eta_{\kappa} )P(\inf \eta_{\kappa} = \mathcal{K})
\end{equation}
Now, we aim to show that $\inf \eta_{\kappa}=0$ almost surely (with probability 1). Consider the case where there is some probability mass at $\inf \eta_{\kappa} = \epsilon >0$. Then:
\begin{equation}
     \epsilon P(\inf \eta_{\kappa} = \epsilon) > 0
\end{equation}
And thus:
\begin{equation}
    \sum^{\mathcal{K}} (\inf \eta_{\kappa} )P(\inf \eta_{\kappa} = \mathcal{K}) > 0
\end{equation}
We know this cannot be the case, thus, $P(\inf \eta_{\kappa} = 0)=1$.
Then:
\begin{align}
\sum_{k=1}^{\infty} \eta_{k} = \sum^{\infty}_{\kappa=1} C-\eta_{\kappa}/2 = \infty \text{ }\text{ a.s.}
\end{align}
This last result follows because from some $\kappa$ onwards, $C>\eta_{\kappa}$ necessarily.
The assumption of $p_{\min}$ can be justified by noting that even though we traditionally have only the guarantee of lim inf convergence, convergence to zero magnitude gradients is observed empirically, suggesting $p_{\kappa}$ roughly increases with $\kappa$ (i.e. the loss curve becomes flatter over time, or in other words, there exist locally quadratic minima). This is also corroborated by investigating Eq.~\ref{eq:sm_sgd_bound} in the case of fixed learning rates over a window (see \cite{bottou2018}, Eq.~4.28b), which demonstrates a decreasing expected average gradient magnitude with increasing SGD steps, suggesting an expected flattening of the loss curve in later windows, and thus a roughly increasing $p_{\kappa}$.

Now we set out to show the third condition on the learning rate, $\sum^{\infty}_{k=1}\eta_{k}^2 < \infty$. As before, ExpTest holds $\eta_k$ constant within each window. The squared-sum over a single window is:
\begin{equation}
S^{(2)}_{w_\kappa} = \sum_{k=\kappa}^{\kappa+w_\kappa} \eta_k^2 = w_\kappa \eta_\kappa^2.
\end{equation}
Using the upper bound from rounding:
\begin{equation}
w_\kappa = \left\lfloor \frac{C}{\eta_\kappa}+\frac12 \right\rfloor \le \frac{C}{\eta_\kappa}+\frac{1}{2}
\end{equation}
we obtain the per-window bound:
\begin{align}
S^{(2)}_{w_\kappa} = w_\kappa \eta_\kappa^2
&\le \left(\frac{C}{\eta_\kappa}+\frac12\right)\eta_\kappa^2 \\
&= C\eta_\kappa + \frac{1}{2} \eta_\kappa^2
\end{align}
Summing over windows ($\kappa$ incremented by window sizes) yields
\begin{equation}
\sum_{k=1}^{\infty}\eta_k^2
=
\sum_{\kappa=1}^{\infty} w_\kappa \eta_\kappa^2
\le
C\sum_{\kappa=1}^{\infty}\eta_\kappa + \frac12\sum_{\kappa=1}^{\infty}\eta_\kappa^2
\end{equation}
Rearranging (and noting that the sum of squared learning rates on the right side is necessarily less than the sum on the left side) gives
\begin{equation} \label{eq:sm_eta2_bound}
\sum_{k=1}^{\infty}\eta_k^2
\le
2C\sum_{\kappa=1}^{\infty}\eta_\kappa
\end{equation}
So we must simply show that the window-wise sum converges: $\sum_{\kappa=1}^{\infty}\eta_\kappa < \infty$ (almost surely).
We now use the same window-wise stochastic recursion introduced above:
\begin{equation}
\eta_{\kappa} =
\begin{cases}
\eta_{\kappa-1}, & \text{if } H=0\\
\beta \eta_{\kappa-1}, & \text{if } H=1,
\end{cases}
\qquad 0<\beta<1
\end{equation}
Conditioning on the observed loss data $L_{w_\kappa}$ and letting $p_\kappa = P(H=1 \, | \,L_{w_\kappa})$, we previously obtained
\begin{equation}
\mathbb{E}\left[\eta_\kappa \, \big| \, L_{w_\kappa}\right] =
\left(1+(\beta-1)p_\kappa\right)\eta_{\kappa-1}
\end{equation}
Assume (as in the previous argument) that there exists $\Gamma$ and $p_{\min}>0$ such that for all $\kappa\ge \Gamma$, $p_\kappa \ge p_{\min}$. Then for all $\kappa\ge \Gamma$,
\begin{equation}
\mathbb{E}[\eta_\kappa]
\le
q \, \mathbb{E}[\eta_{\kappa-1}],
\qquad
q = 1+(\beta-1)p_{\min} \in (0,1)
\end{equation}
Iterating yields the geometric bound
\begin{equation}
\mathbb{E}[\eta_\kappa]
\le
\mathbb{E}[\eta_\Gamma] q^{\kappa-\Gamma},
\qquad \kappa\ge \Gamma
\end{equation}
and therefore
\begin{align}
\sum_{\kappa=1}^{\infty}\mathbb{E}[\eta_\kappa]
&=
\sum_{\kappa=1}^{\Gamma-1}\mathbb{E}[\eta_\kappa]
+
\sum_{\kappa=\Gamma}^{\infty}\mathbb{E}[\eta_\kappa] \\
&\le
\sum_{\kappa=1}^{\Gamma-1}\mathbb{E}[\eta_\kappa]
+
\mathbb{E}[\eta_\Gamma]\sum_{j=0}^{\infty}q^j \\
&< \infty
\end{align}
We know this sum converges because the first sum on the right-hand-side is finite, and the second is a geometric series with $|q|<1$, which necessarily converges. Since $\eta_\kappa \ge 0$, Tonelli's theorem implies
\begin{equation}
\mathbb{E}\left[\sum_{\kappa=1}^{\infty}\eta_\kappa\right] =
\sum_{\kappa=1}^{\infty}\mathbb{E}[\eta_\kappa]
<\infty,
\end{equation}
which in turn implies $\sum_{\kappa=1}^{\infty}\eta_\kappa < \infty$ almost surely (otherwise the expectation would be infinite). Combining this with Eq.~\ref{eq:sm_eta2_bound} gives
\begin{equation}
\sum_{k=1}^{\infty}\eta_k^2 < \infty
\qquad \text{a.s.}
\end{equation}
This establishes the third learning-rate condition, completing proof of ExpTest's convergence.

\section{Experimental Results} \label{sec:sm_experiments}
\subsection{Handwritten Digit Classification} \label{sec:sm_mnist}
We begin with the MNIST handwritten digit classification task \cite{6296535}. We use this as a toy task for validating ExpTest. We train logistic regression models with a default mini-batch size of 32 and a fixed epoch limit of 5. We also explore hyperparameter choices and batch size effects. In this experiment, we exclude Lipschitz recalibration, since this is a single-layer task.

\subsubsection{MNIST Learning Rate Range} \label{sec:sm_mnist_lr}
\begin{figure}[htbp]
    \centering
    \begin{subfigure}[t]{0.324\textwidth}
        \centering
        \includegraphics[width=\linewidth]{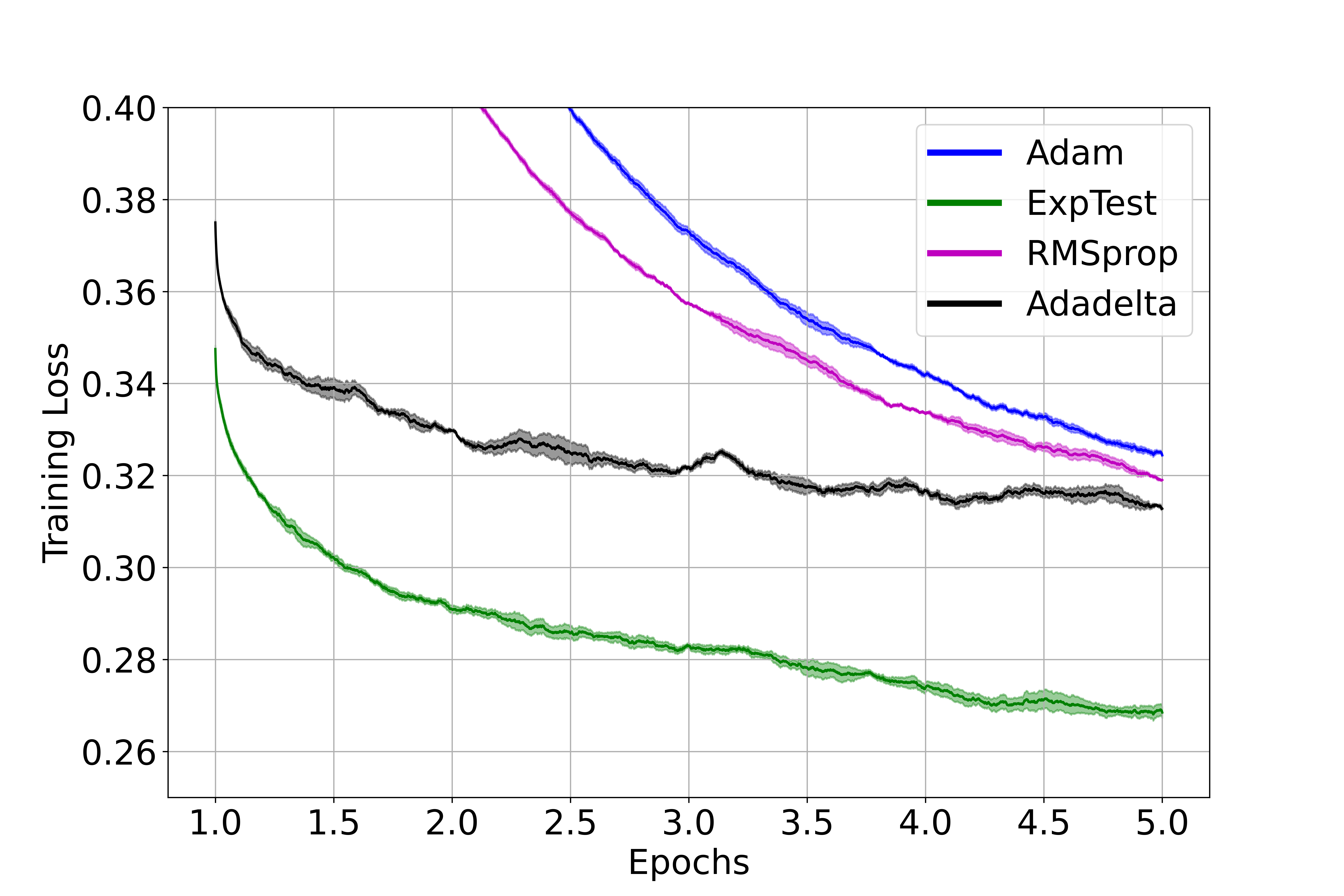}
        \caption{$\eta = 0.001 \eta_{\text{max}}$}
        \label{fig:mnist_lr_sub1}
    \end{subfigure}
    \begin{subfigure}[t]{0.324\textwidth}
        \centering
        \includegraphics[width=\linewidth]{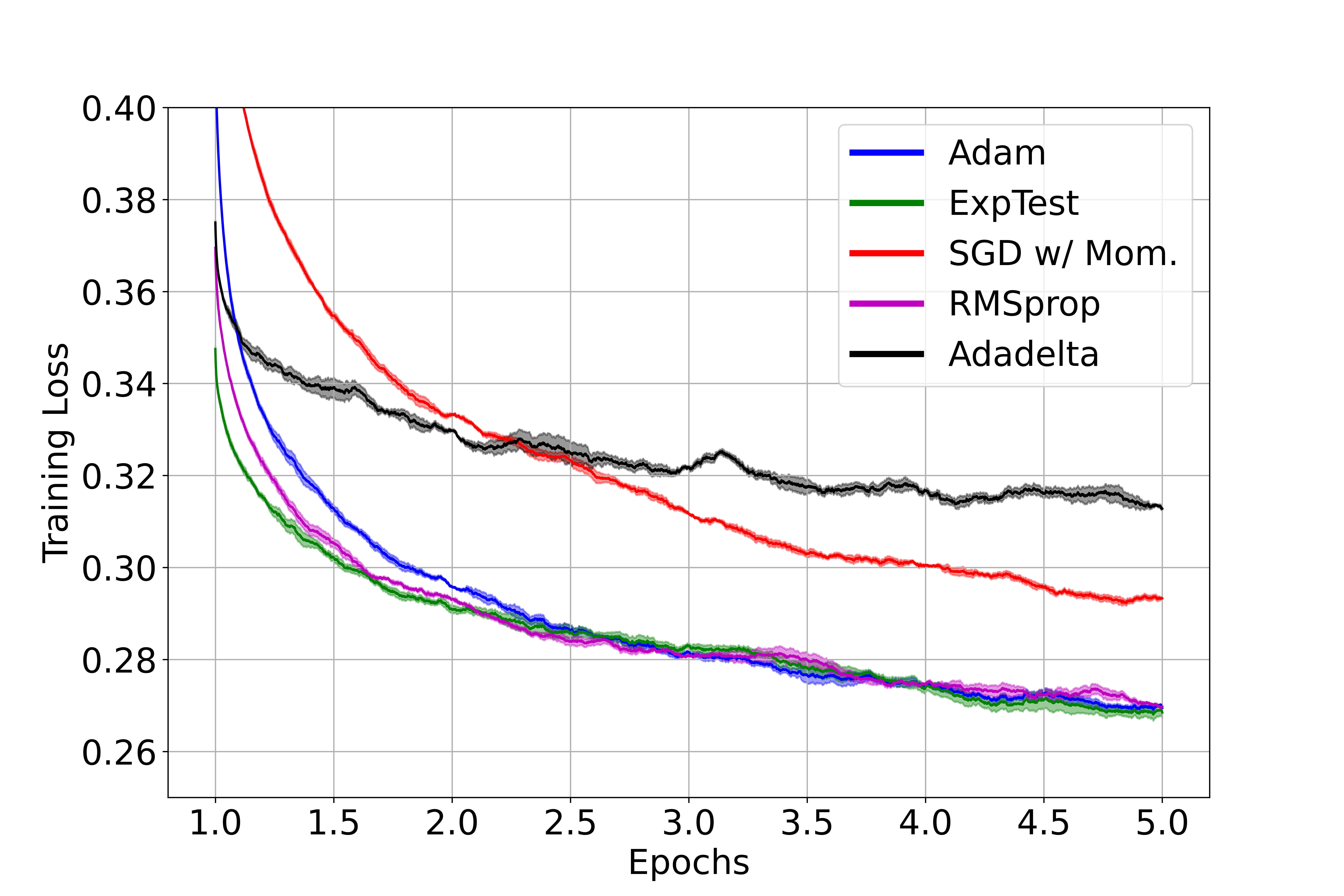}
        \caption{$\eta = 0.01 \eta_{\text{max}}$}
        \label{fig:mnist_lr_sub2}
    \end{subfigure}
    \begin{subfigure}[t]{0.324\textwidth}
        \centering
        \includegraphics[width=\linewidth]{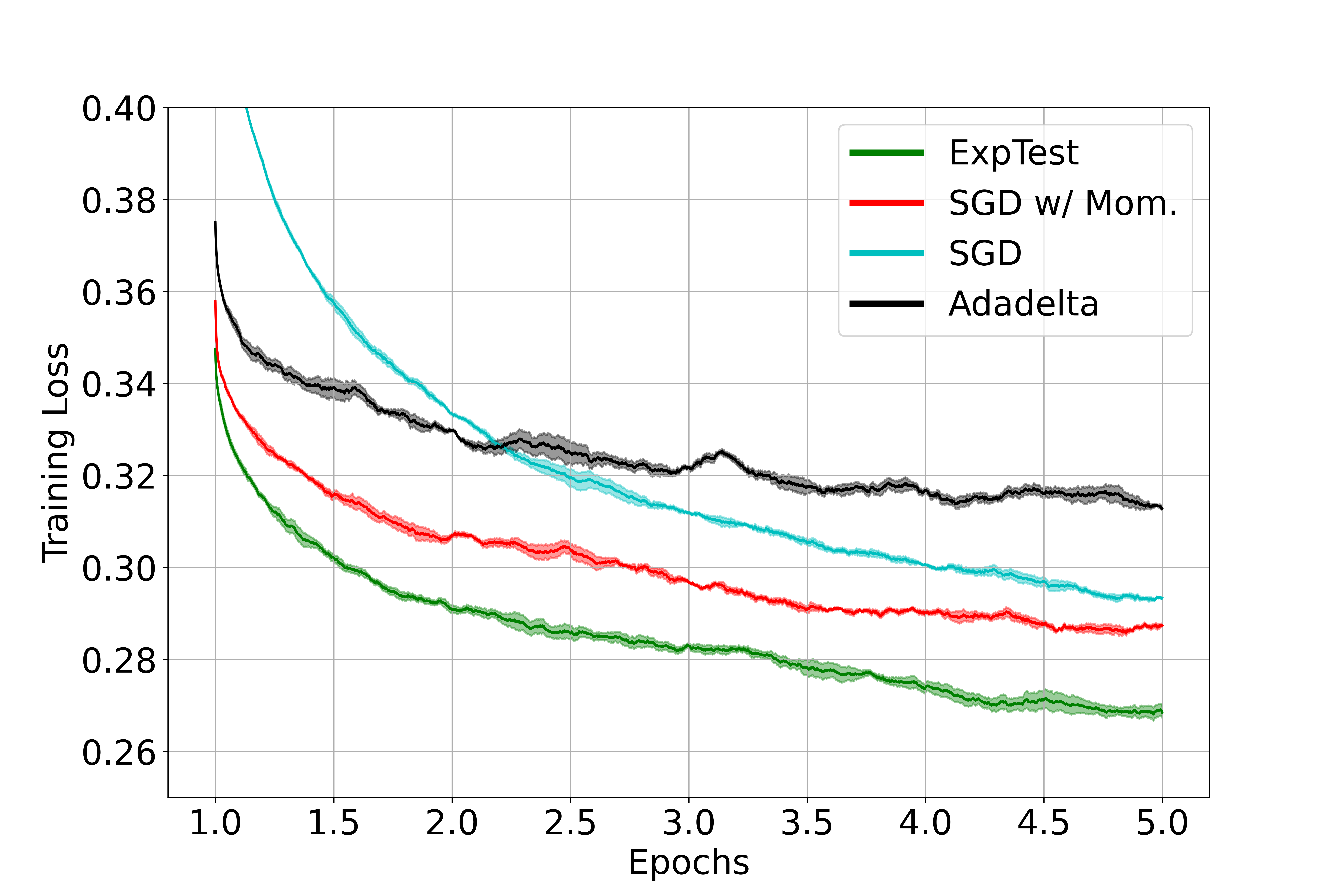}
        \caption{$\eta = 0.1 \eta_{\text{max}}$}
        \label{fig:mnist_lr_sub3}
    \end{subfigure}
    \begin{subfigure}[t]{0.324\textwidth}
        \centering
        \includegraphics[width=\linewidth]{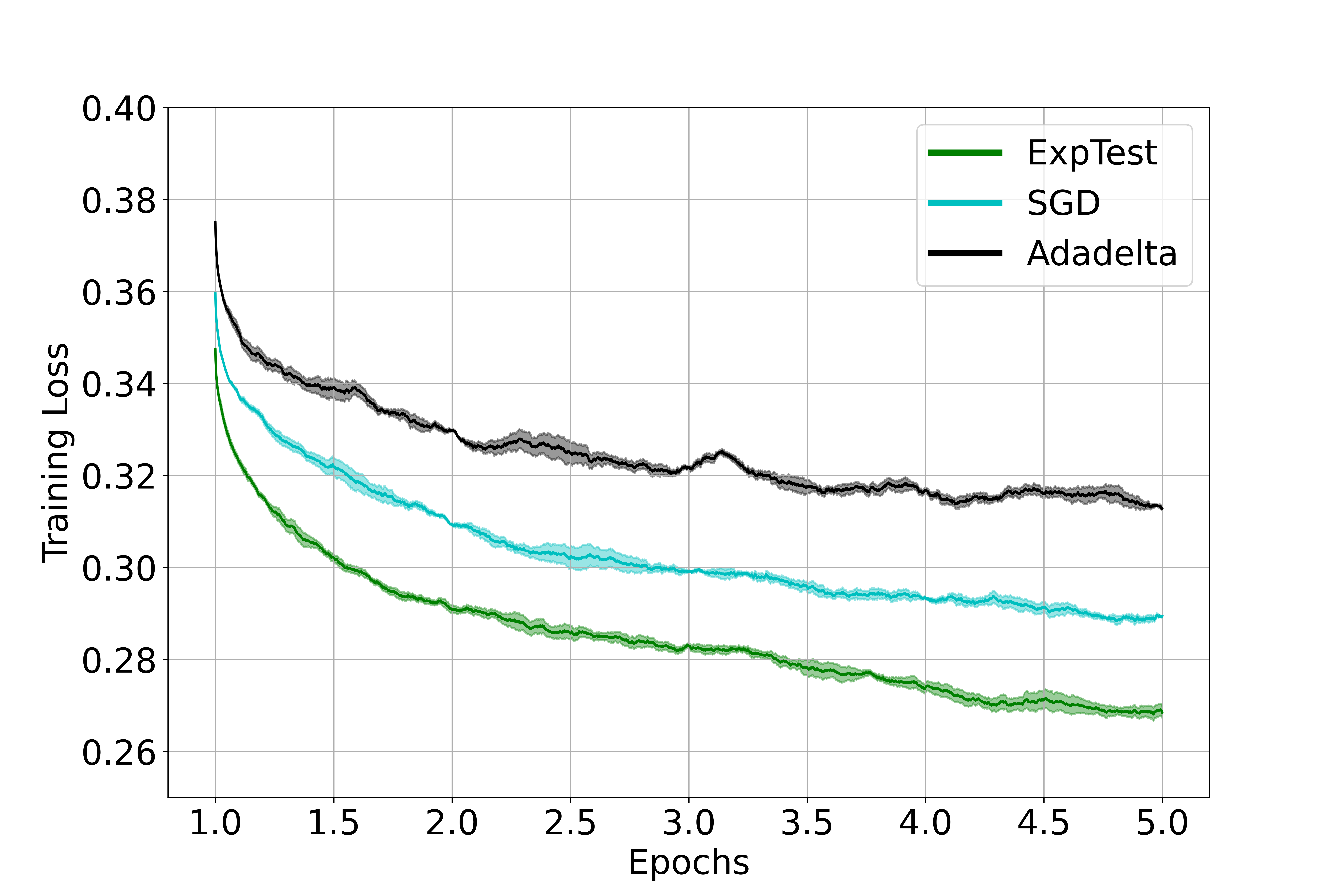}
        \caption{$\eta = \eta_{\text{max}}$}
        \label{fig:mnist_lr_sub4}
    \end{subfigure}
    \begin{subfigure}[t]{0.324\textwidth}
        \centering
        \includegraphics[width=\linewidth]{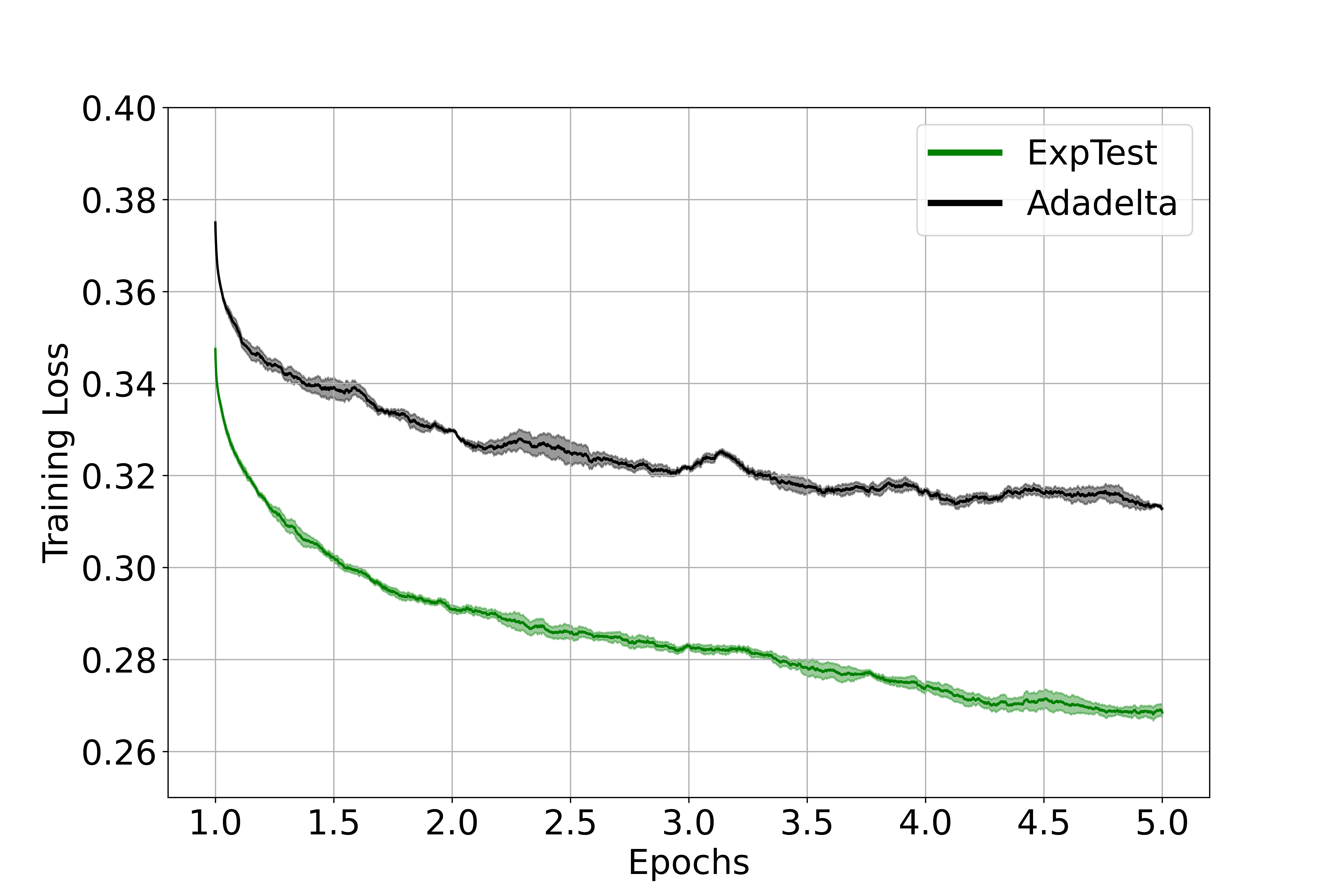}
        \caption{$\eta = 10 \eta_{\text{max}}$}
        \label{fig:mnist_lr_sub5}
    \end{subfigure}
    \caption{Training loss curves for logistic regression on MNIST, moving averaged over window of 1 epoch. Center line shows mean of 5 trials with standard error shown by shading. Optimizers with high loss excluded from legend for clarity.}
    \label{fig:mnist_lr_multi}
\end{figure}

We calculate $\eta_{\text{max}}$ on the MNIST dataset, as described in Algorithm~\ref{alg:exptest} of the main text for classification tasks, calculating the covariance matrix over the entire training dataset. We consider five learning rates over a logarithmic range: [$0.001\eta_{\text{max}}$, $0.01\eta_{\text{max}}$, $0.1\eta_{\text{max}}$, $\eta_{\text{max}}$, and $10\eta_{\text{max}}$]. We train with SGD, SGD with momentum, Adam, and RMSprop at each of these initial learning rates and compare to ExpTest ($\alpha = 0.05$, $\beta = 0.33$) and Adadelta without choosing a global initial learning rate. We perform five trials with different random initializations for each experiment. The training loss curves are visible in the five panels of Fig.~\ref{fig:mnist_lr_multi}, and the test accuracies in Table~\ref{tab:mnist_lr_accuracies}. For clarity, loss curves above the cut-off loss (0.4) are excluded, and the data is moving averaged by 1875 (number of mini-batches per epoch).

ExpTest displays the fastest initial convergence compared to all optimizers at all learning rates. Additionally, ExpTest achieves the minimum training loss compared to Adadelta and all optimizers with learning rate selection, except for Adam and RMSprop at $\eta = 0.01 \eta_{\text{max}}$. ExpTest achieves the maximum test accuracy compared to all optimizers, except the two previously mentioned, though it still produces comparable performance. All optimizers with learning rate selection (Adam, SGD, SGD w/ Mom., and RMSprop) display considerable variability in loss curve behavior and test accuracy over the learning rate range.

\begin{table*}[htbp]
    \centering
    \caption{Test Accuracies on MNIST Logistic Regression ($\mu \pm \sigma$, $n=5$)}
    \label{tab:mnist_lr_accuracies}
    \begin{tabular}{@{}c|ccccc@{}} 
        \toprule
        Optimizer & $0.001\eta_{\text{max}}$ & $0.01\eta_{\text{max}}$ & $0.1\eta_{\text{max}}$ & $\eta_{\text{max}}$ & $10\eta_{\text{max}}$ \\ \midrule
        ExpTest & $\mathbf{92.30 \pm 0.07}$ & $92.30 \pm 0.07$ & $\mathbf{92.30 \pm 0.07}$ & $\mathbf{92.30 \pm 0.07}$ & $\mathbf{92.30 \pm 0.07}$ \\
        Adam & $91.45 \pm 0.07$ & $\mathbf{92.35 \pm 0.06}$ & $90.20 \pm 0.50$ & $88.92 \pm 0.76$ & $88.30 \pm 1.55$ \\
        SGD & $82.74 \pm 0.60$ & $89.86 \pm 0.16$ & $91.96 \pm 0.05$ & $91.65 \pm 0.29$ & $87.96 \pm 2.65$ \\
        SGD w/ Mom. & $89.86 \pm 0.09$ & $91.93 \pm 0.08$ & $91.94 \pm 0.28$ & $88.48 \pm 1.06$ & $87.62 \pm 1.25$ \\
        RMSprop & $91.53 \pm 0.04$ & $92.34 \pm 0.18$ & $89.77 \pm 1.69$ & $88.90 \pm 1.87$ & $88.71 \pm 1.01$ \\
        Adadelta & $91.46 \pm 0.35$ & $91.46 \pm 0.35$ & $91.46 \pm 0.35$ & $91.46 \pm 0.35$ & $91.46 \pm 0.35$ \\ \bottomrule
    \end{tabular}
\end{table*}

\subsubsection{Robustness to Mini-Batch Size} \label{sec:sm_mnist_bs}
\begin{figure}[htbp]
    \centering
    \begin{subfigure}[t]{0.324\textwidth}
        \centering
        \includegraphics[width=\linewidth]{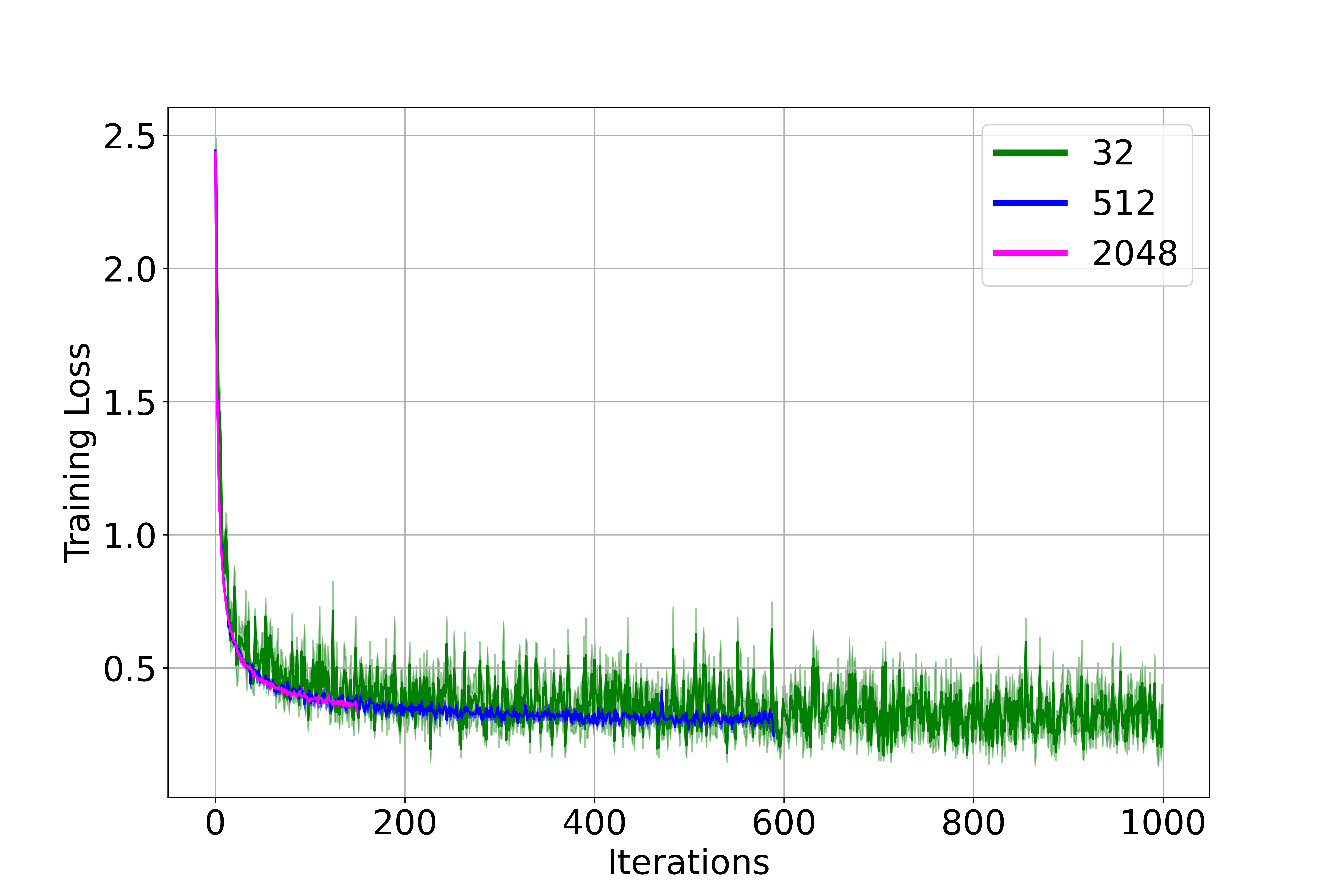}
        \caption{First 1000 training iterations}
        \label{fig:mnist_bs_sub1}
    \end{subfigure}
    \begin{subfigure}[t]{0.324\textwidth}
        \centering
        \includegraphics[width=\linewidth]{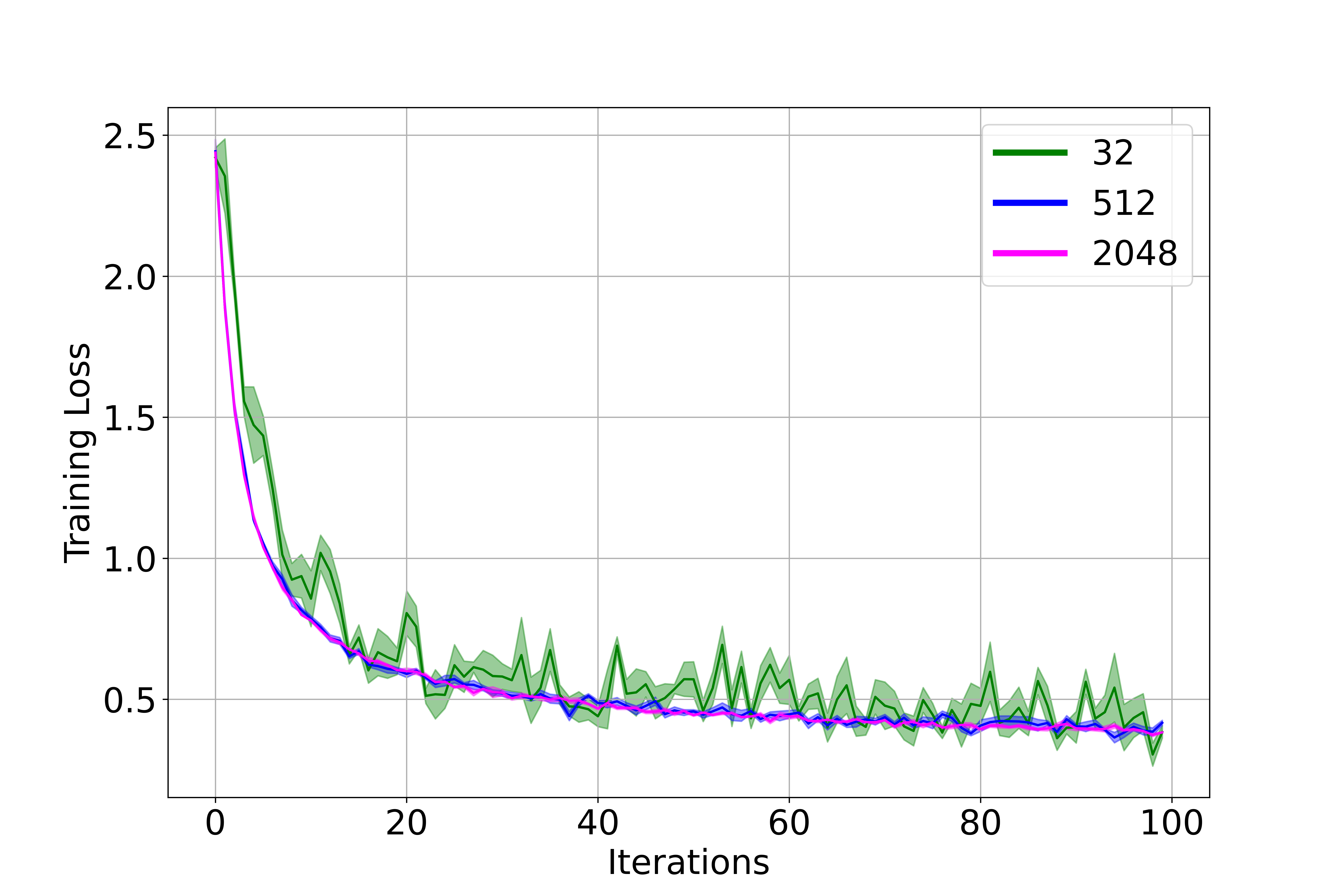}
        \caption{First 100 training iterations}
        \label{fig:mnist_bs_sub2}
    \end{subfigure}
    \caption{Training loss curves on MNIST logistic regression for ExpTest ($\alpha = 0.05$, $\beta = 0.33$) at three different mini-batch sizes. Lines represent mean of 5 trials with standard error shown by shading.}
    \label{fig:mnist_bs_multi}
\end{figure}

We train with ExpTest at three different mini-batch sizes: [32, 512, 2048] over 5 epochs. The training loss curves are plotted over iteration number for the first 1000 iterations in Fig.~\ref{fig:mnist_bs_sub1} and first 100 iterations in Fig.~\ref{fig:mnist_bs_sub2}. ExpTest displays very similar convergence properties at all three mini-batch-sizes, though the loss decay rate and smoothness of the curve increase with increasing batch size (as expected with noisy gradient estimates).

\subsubsection{Robustness to Hyper-Parameter Selection} \label{sec:sm_mnist_hp}
We train ExpTest at three different values of $\alpha$: [0.1, 0.05, 0.01] and three different values of $\beta$: [0.05, 0.1, 0.33] over 5 epochs. The test accuracy results are displayed in Table~\ref{tab:mnist_hp_accuracies}. The variability in performance across hyper-parameter choices for ExpTest is significantly less than for optimizers with learning rate selection across learning rate choices (see Table~\ref{tab:mnist_lr_accuracies}). The optimal hyper-parameters ($\alpha = 0.05$, $\beta = 0.33$) give very similar test accuracy results to Adam and RMSprop with optimal learning rate selection: 92.30\% for ExpTest (Table~\ref{tab:mnist_hp_accuracies}) vs. 92.35\% and 92.34\% for Adam and RMSProp, respectively (Table~\ref{tab:mnist_lr_accuracies}).

\begin{table}[htbp]
    \centering
    \caption{Test Accuracies on MNIST Logistic Regression Over Hyper-Parameters $\alpha$ and $\beta$ ($\mu \pm \sigma$, $n=5$)}
    \label{tab:mnist_hp_accuracies}
    \begin{tabular}{@{}c|ccc@{}} 
        \toprule
         & $\beta = 0.05$ & $\beta = 0.1$ & $\beta = 0.33$ \\ \midrule
         $\alpha = 0.1$ &$92.11 \pm 0.06$ &$92.11 \pm 0.10$ &$92.26 \pm 0.14$ \\
         $\alpha = 0.05$ &$92.06 \pm 0.07$ &$92.08 \pm 0.11$ &$\mathbf{92.30 \pm 0.07}$ \\
         $\alpha = 0.01$ &$92.05 \pm 0.06$ &$92.06 \pm 0.11$ &$92.25 \pm 0.12$ 
         \\ \bottomrule
    \end{tabular}
\end{table}

\subsection{Housing Price Regression Model Selection} \label{sec:sm_housing_model}
\begin{figure}[htbp] 
    \centering
    \includegraphics[width=\linewidth]{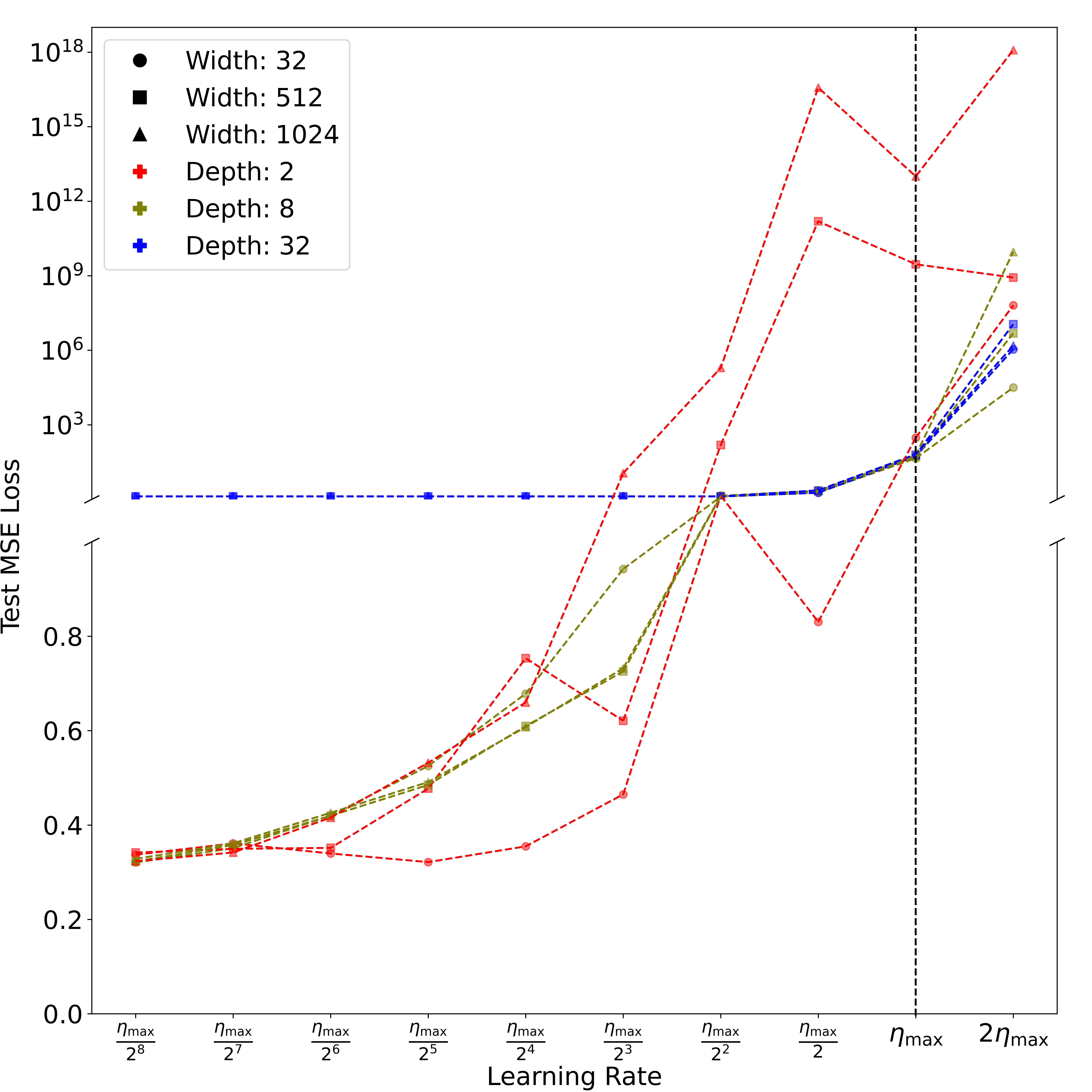}
    \caption{Test MSE-Losses for model width-depth combinations in the lattice: $w=[32, 512, 1024]$, $d=[2,8,32]$ over the displayed learning rate range. Shapes correspond to width and colors to depth. Values are the mean of 5 trials.}
    \label{fig:appE}
\end{figure}

We trained several fully-connected models of several widths and depths over a range of learning rates with plain SGD and a form of early-stopping. We split the dataset as described in Section~\ref{sec:housing} of the main text, training each model until validation loss increased, and evaluating the model weights with the best validation loss on the test set. The test MSE-losses of the models over the learning rate range are displayed in Fig.~\ref{fig:appE}. The model with $(\text{width},\text{depth})=(2,32)$ was selected, as it consistently displays the lowest test loss.

\subsection{Housing Price Regression Training Loss Curves} \label{sec:sm_housing_train}
\begin{figure}[htbp]
    \centering
    \begin{subfigure}[t]{0.324\textwidth}
        \centering
        \includegraphics[width=\linewidth]{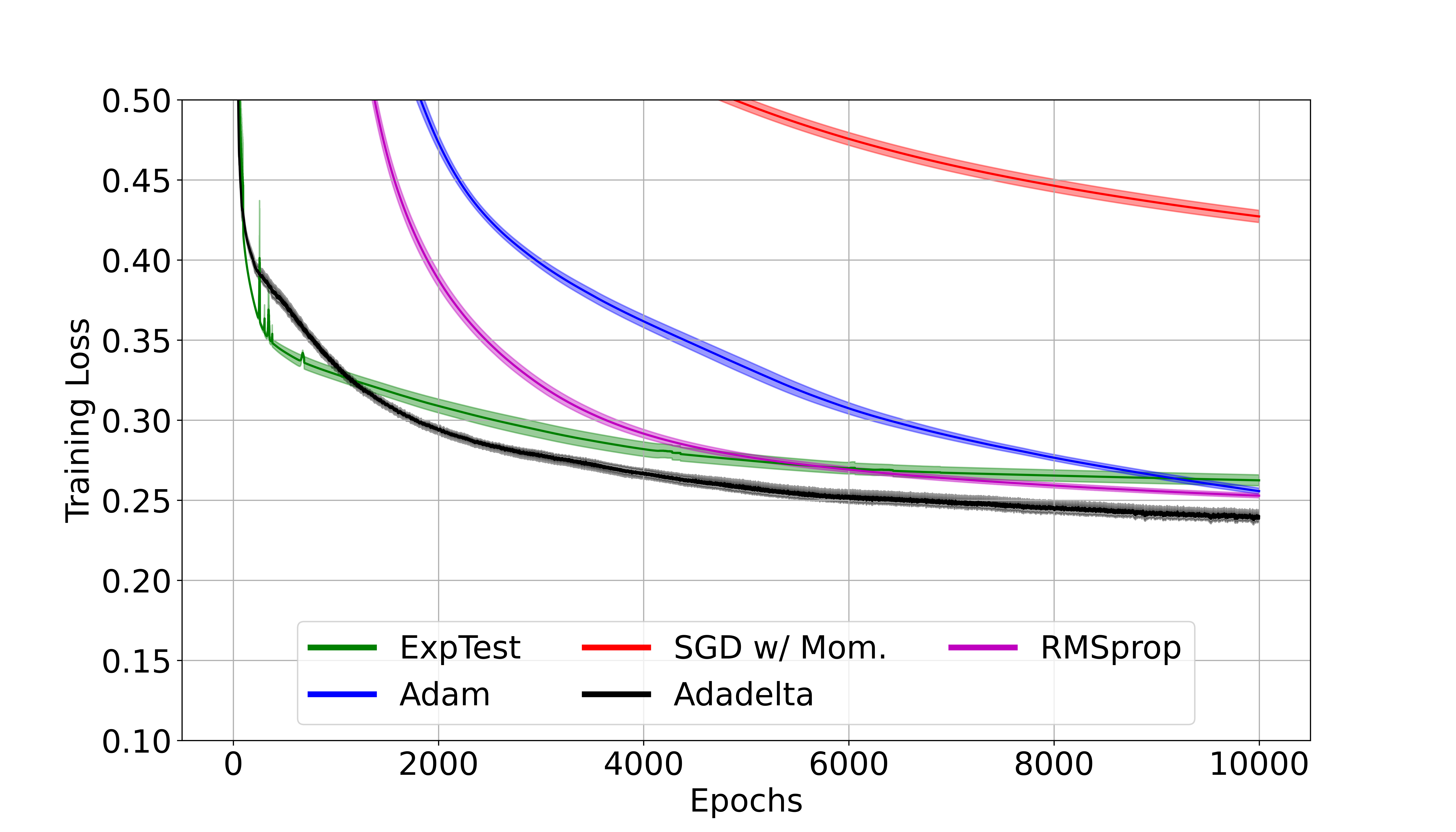}
        \caption{$\eta = 0.0001 \eta_{\text{max}}$}
        \label{fig:cali_train_sub1}
    \end{subfigure}
    \begin{subfigure}[t]{0.324\textwidth}
        \centering
        \includegraphics[width=\linewidth]{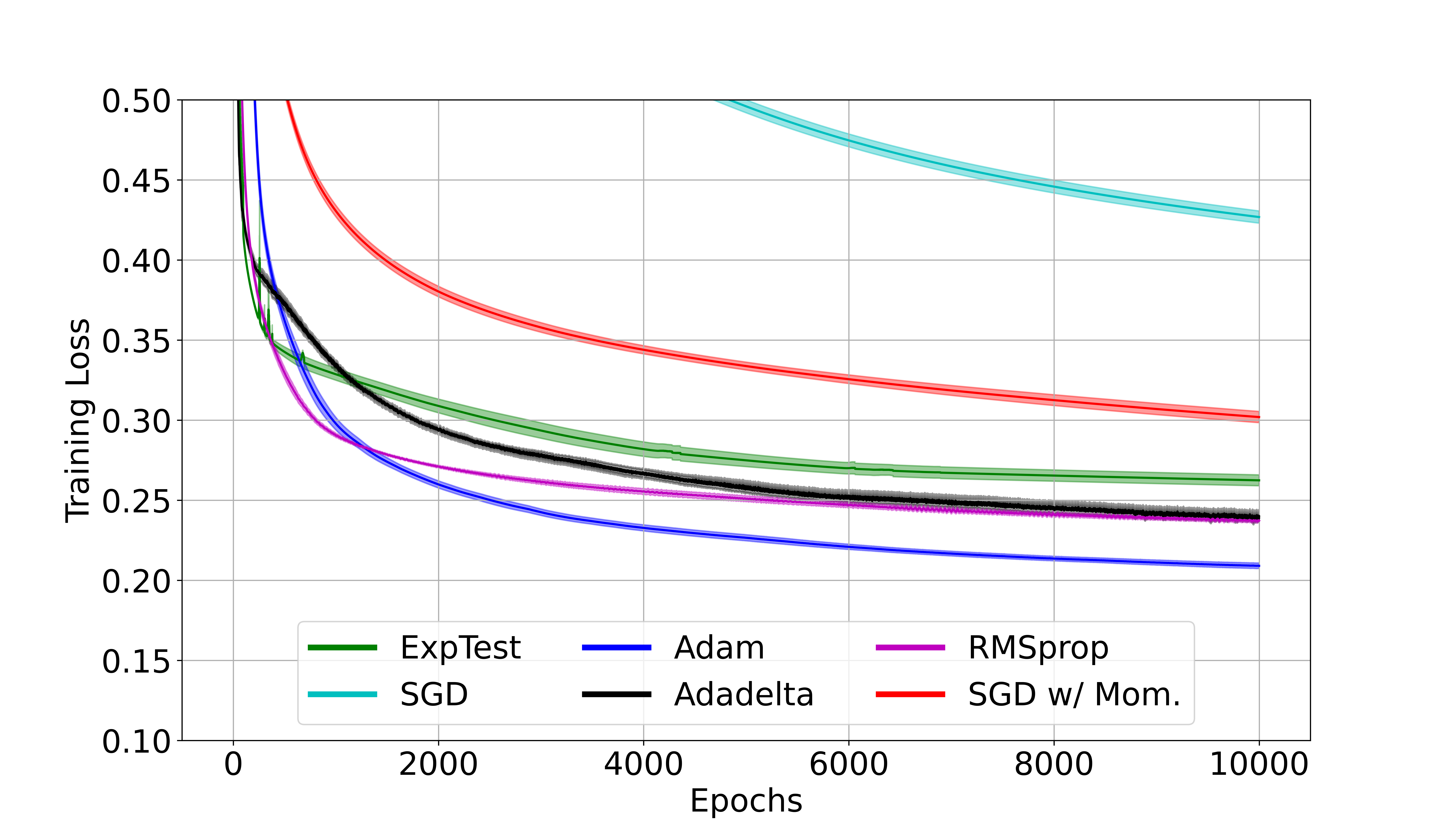}
        \caption{$\eta = 0.001 \eta_{\text{max}}$}
        \label{fig:cali_train_sub2}
    \end{subfigure}
    \begin{subfigure}[t]{0.324\textwidth}
        \centering
        \includegraphics[width=\linewidth]{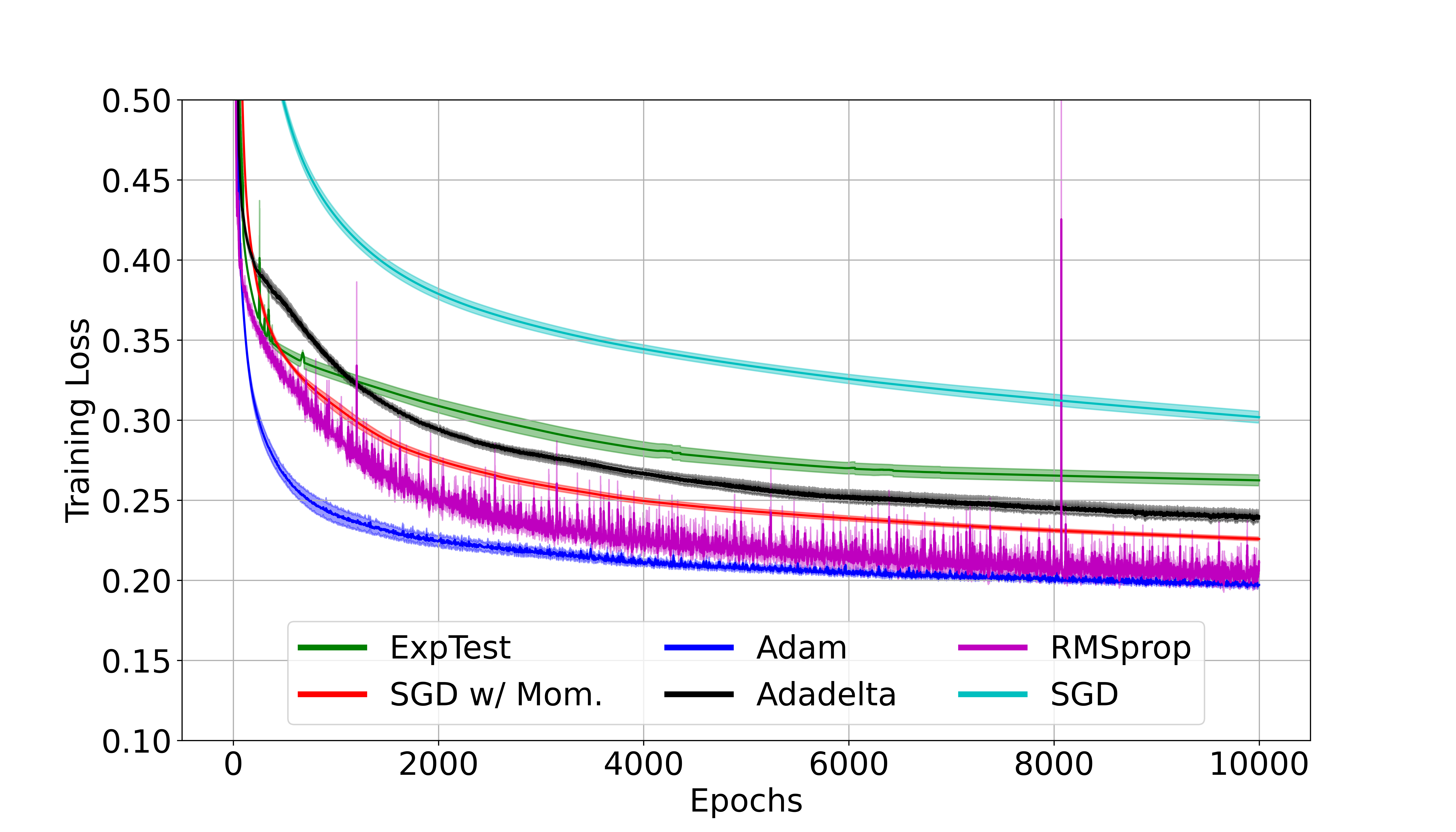}
        \caption{$\eta = 0.01 \eta_{\text{max}}$}
        \label{fig:cali_train_sub3}
    \end{subfigure}
    \begin{subfigure}[t]{0.324\textwidth}
        \centering
        \includegraphics[width=\linewidth]{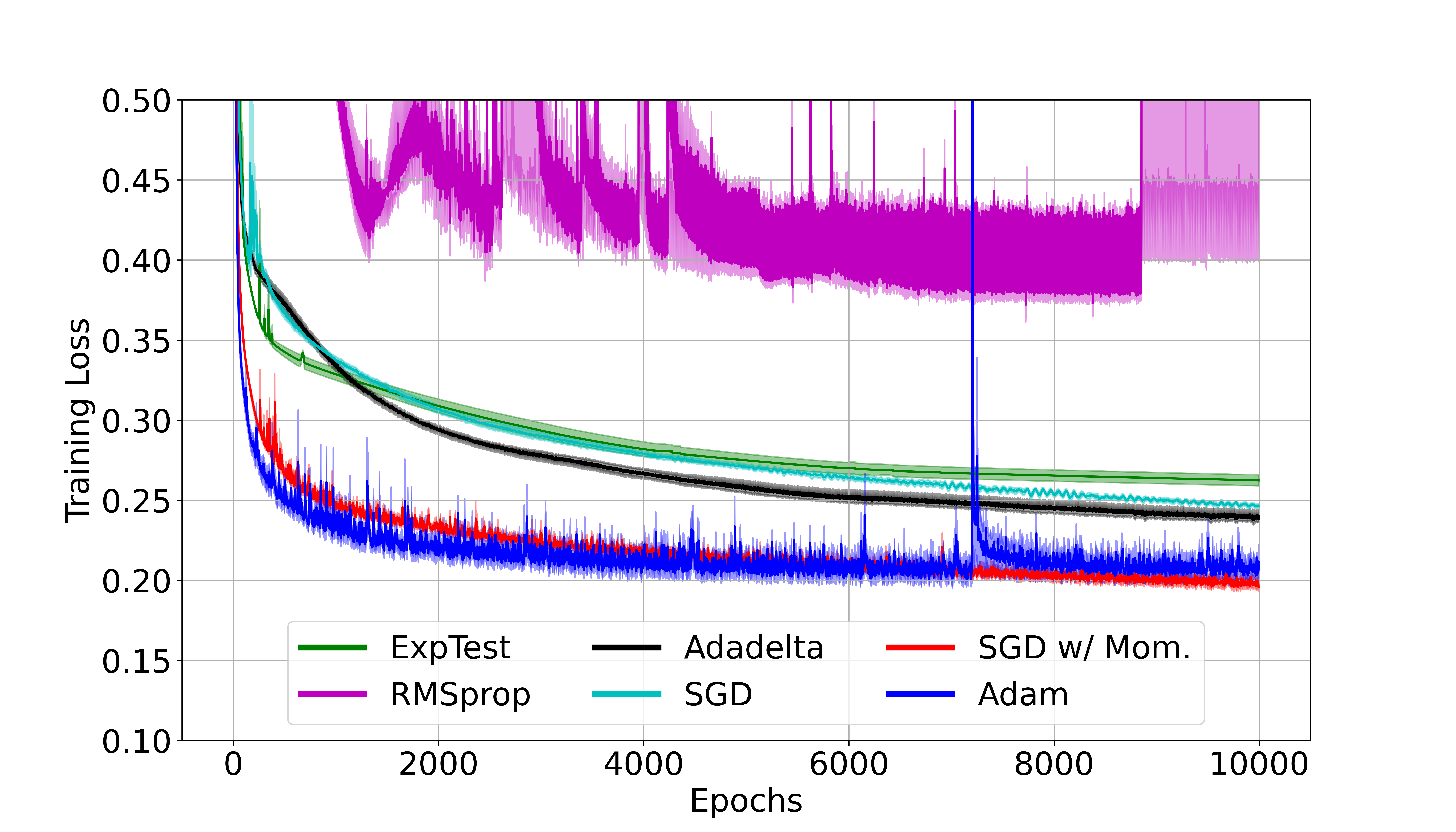}
        \caption{$\eta = 0.1 \eta_{\text{max}}$}
        \label{fig:cali_train_sub4}
    \end{subfigure}
    \begin{subfigure}[t]{0.324\textwidth}
        \centering
        \includegraphics[width=\linewidth]{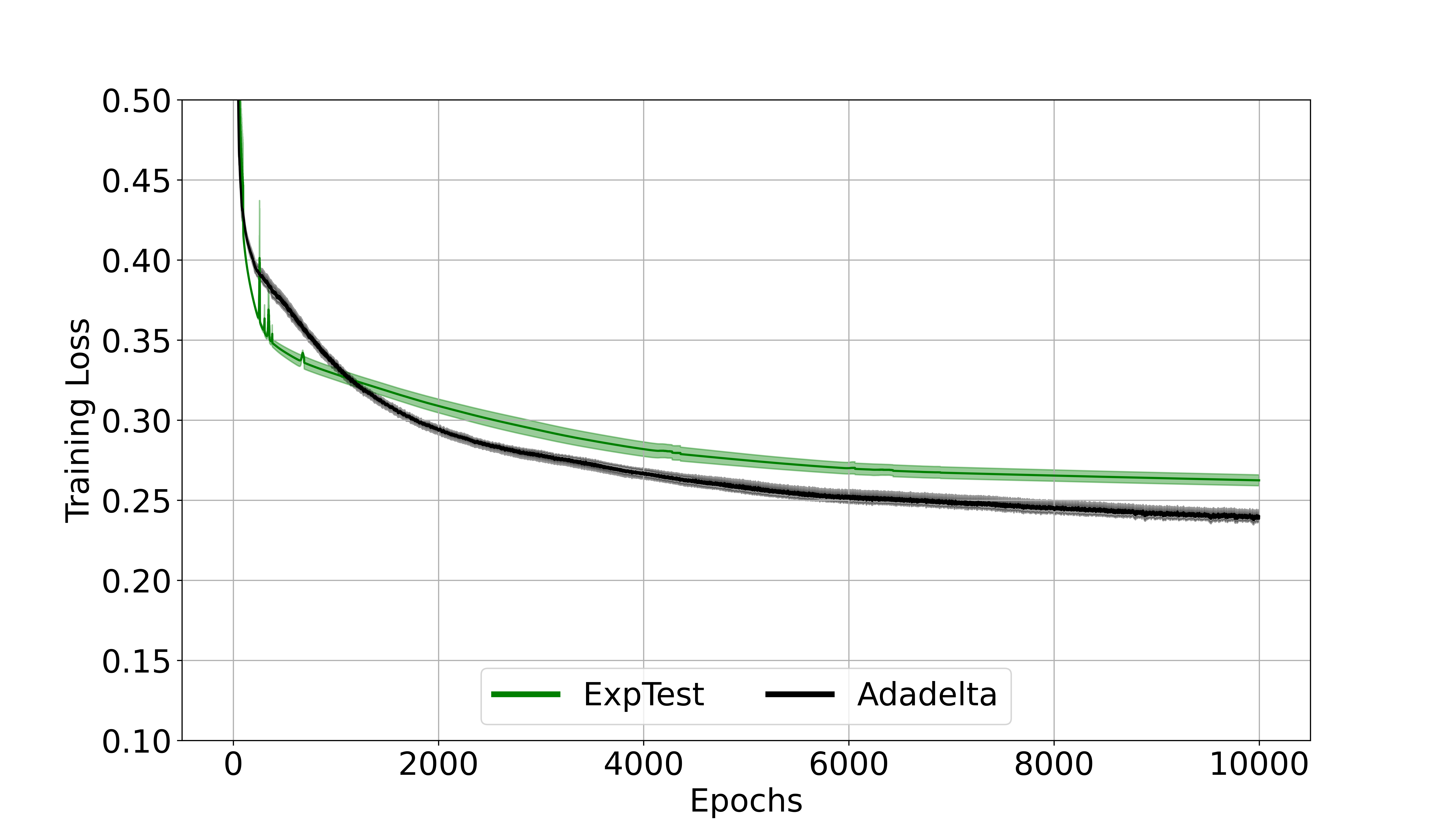}
        \caption{$\eta = \eta_{\text{max}}$}
        \label{fig:cali_train_sub5}
    \end{subfigure}
    \caption{Training loss curves for regression with fully connected network on California Housing Dataset. Center line shows mean of 5 trials with standard error shown by shading. $y$-axis is limited at 0.5 for clarity.}
    \label{fig:cali_train}
\end{figure}

We present the training loss curves for the experiments conducted in Section~\ref{sec:housing} of the main text in Fig.~\ref{fig:cali_train}.

\subsection{Housing Price Regression Validation Loss Curves} \label{sec:sm_housing_val}
\begin{figure}[htbp]
    \centering
    \begin{subfigure}[t]{0.324\textwidth}
        \centering
        \includegraphics[width=\linewidth]{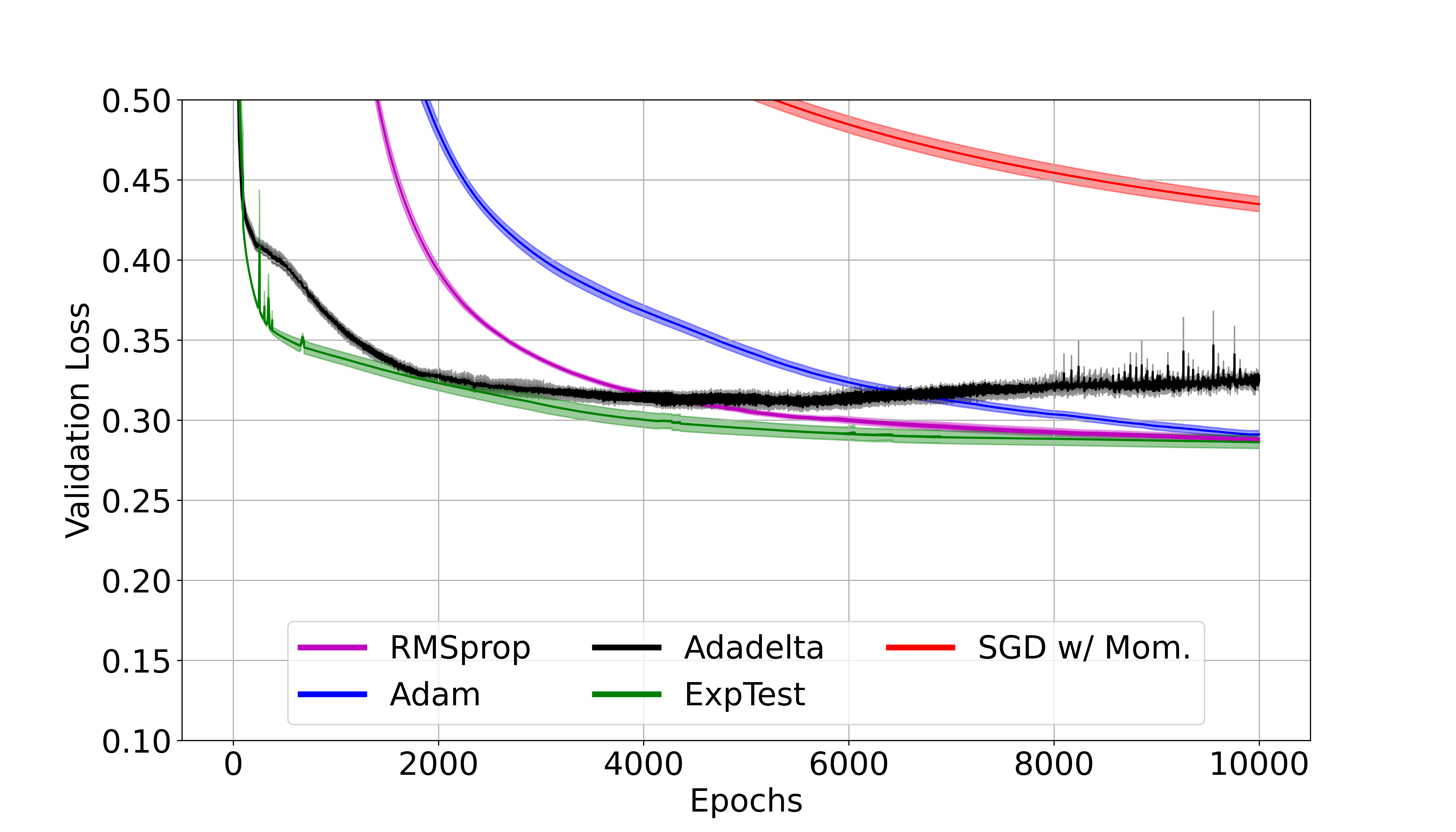}
        \caption{$\eta = 0.0001 \eta_{\text{max}}$}
        \label{fig:cali_val_sub1}
    \end{subfigure}
    \begin{subfigure}[t]{0.324\textwidth}
        \centering
        \includegraphics[width=\linewidth]{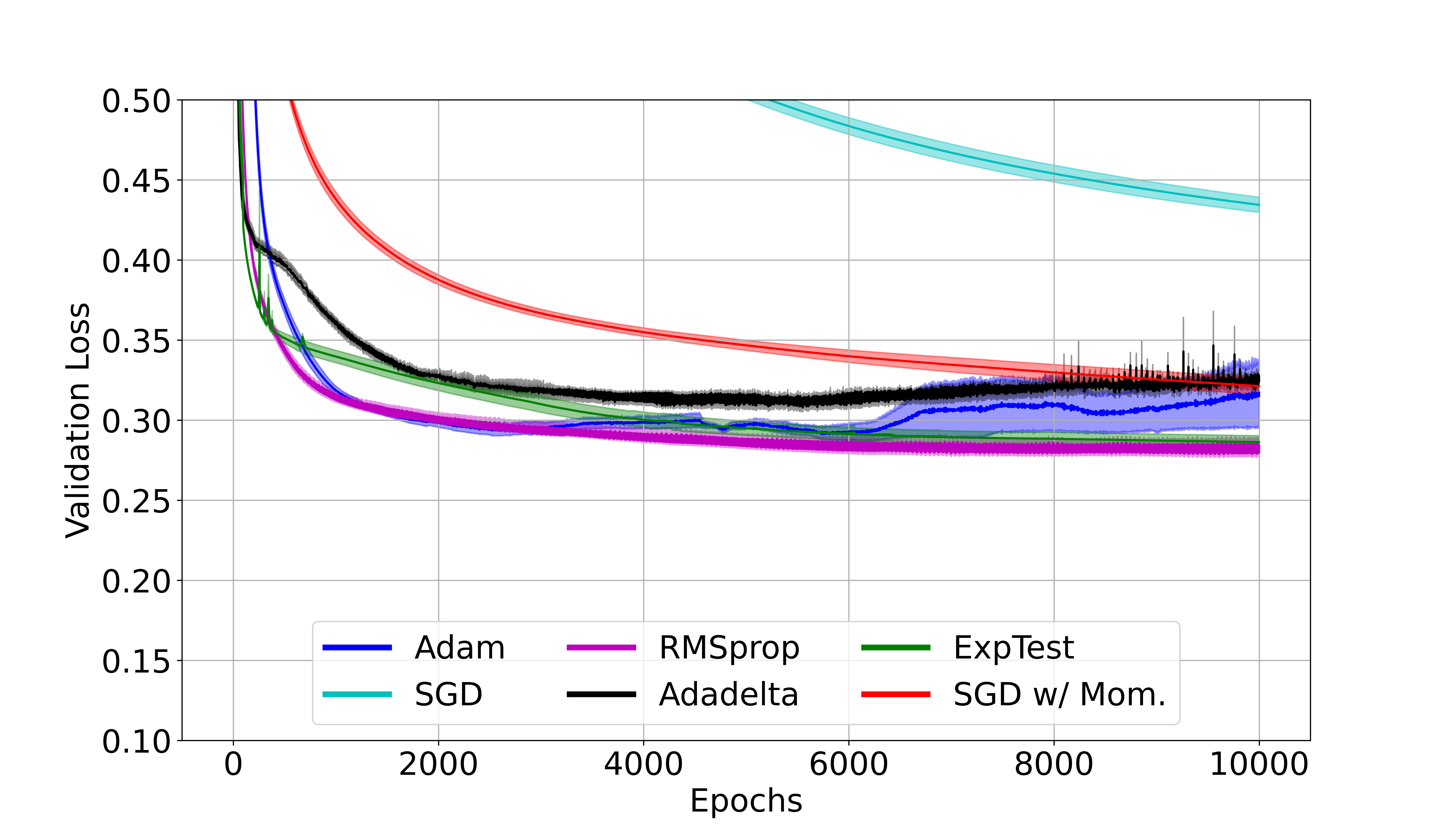}
        \caption{$\eta = 0.001 \eta_{\text{max}}$}
        \label{fig:cali_val_sub2}
    \end{subfigure}
    \begin{subfigure}[t]{0.324\textwidth}
        \centering
        \includegraphics[width=\linewidth]{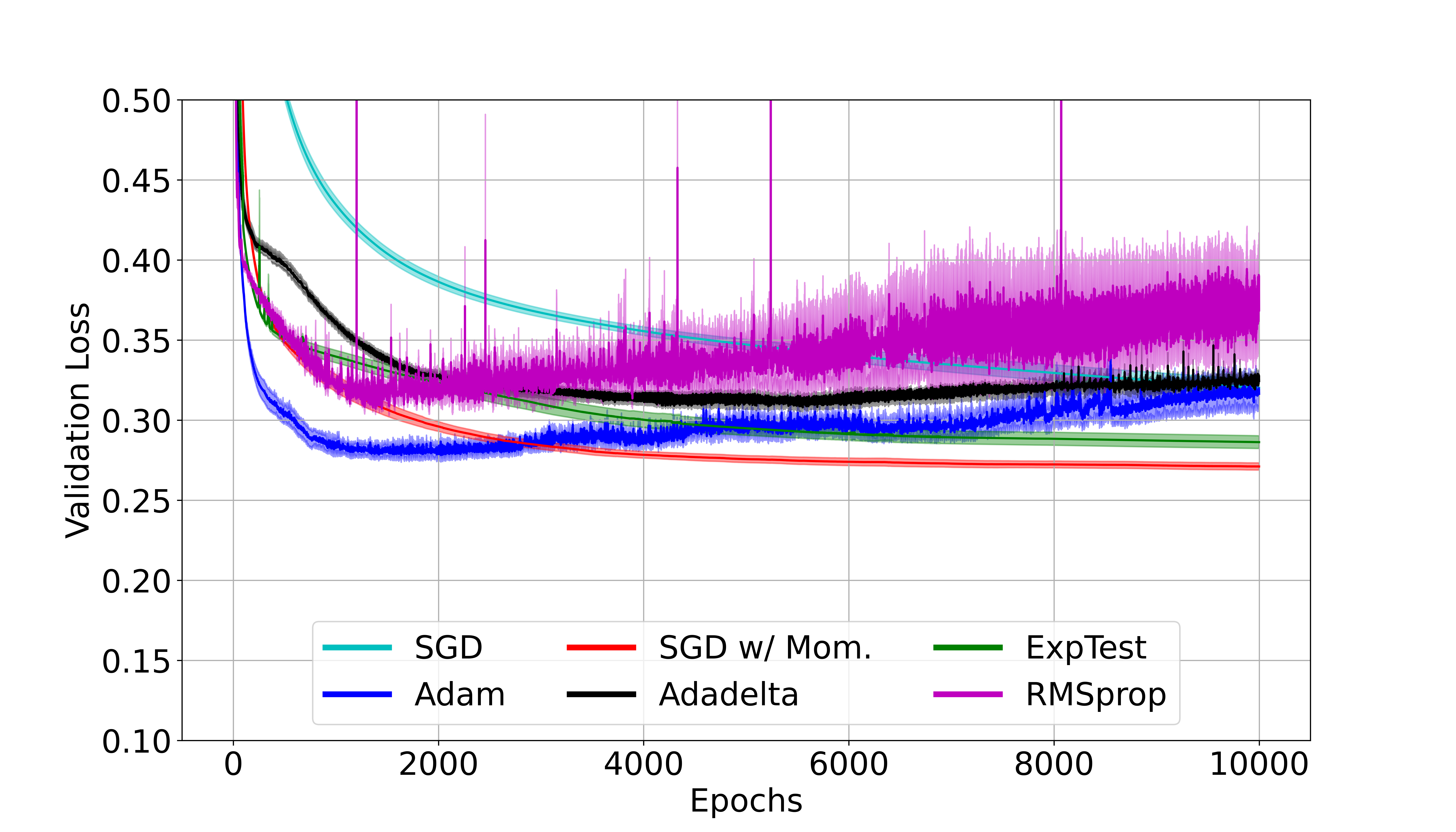}
        \caption{$\eta = 0.01 \eta_{\text{max}}$}
        \label{fig:cali_val_sub3}
    \end{subfigure}
    \begin{subfigure}[t]{0.324\textwidth}
        \centering
        \includegraphics[width=\linewidth]{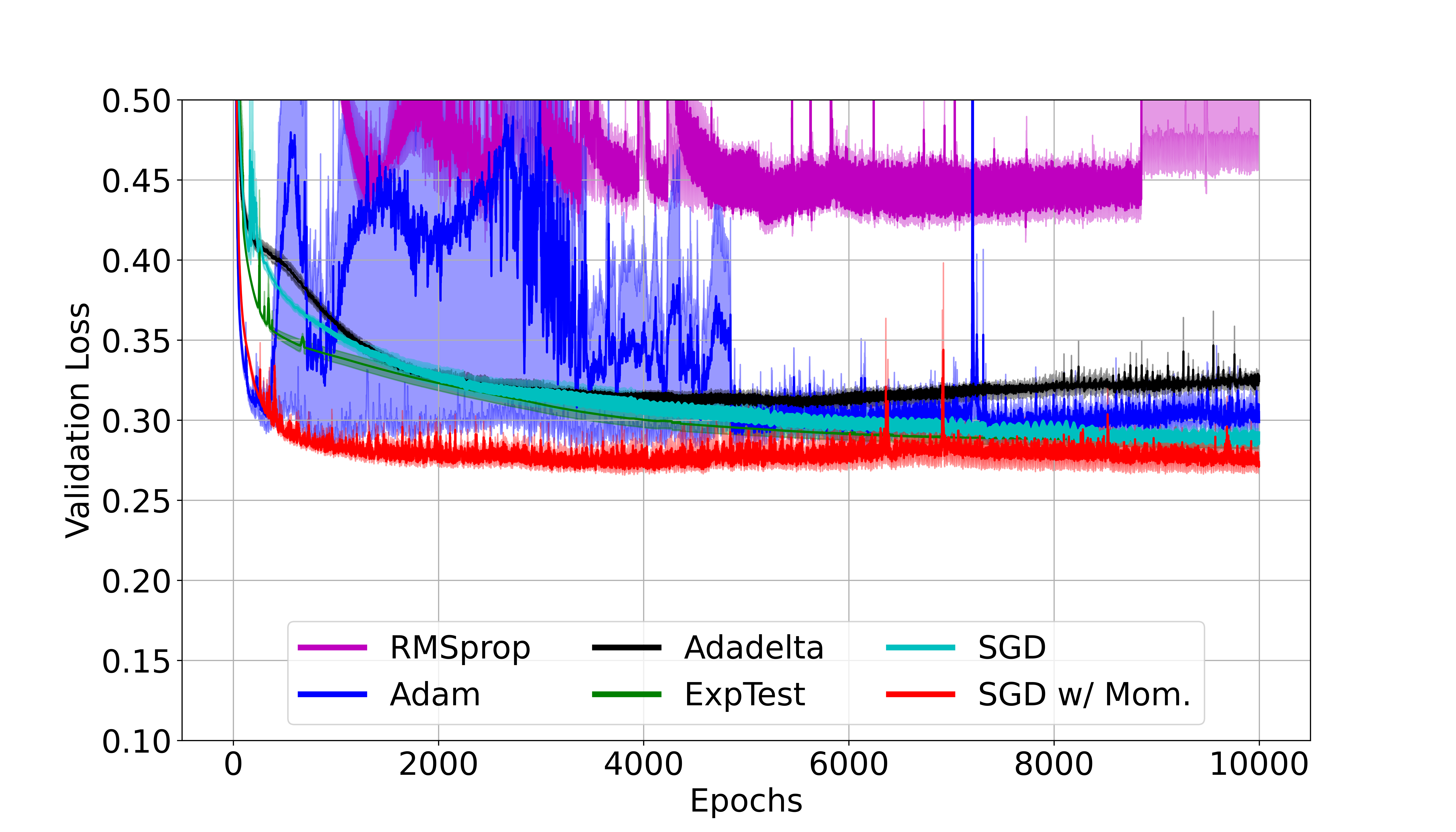}
        \caption{$\eta = 0.1\eta_{\text{max}}$}
        \label{fig:cali_val_sub4}
    \end{subfigure}
    \begin{subfigure}[t]{0.324\textwidth}
        \centering
        \includegraphics[width=\linewidth]{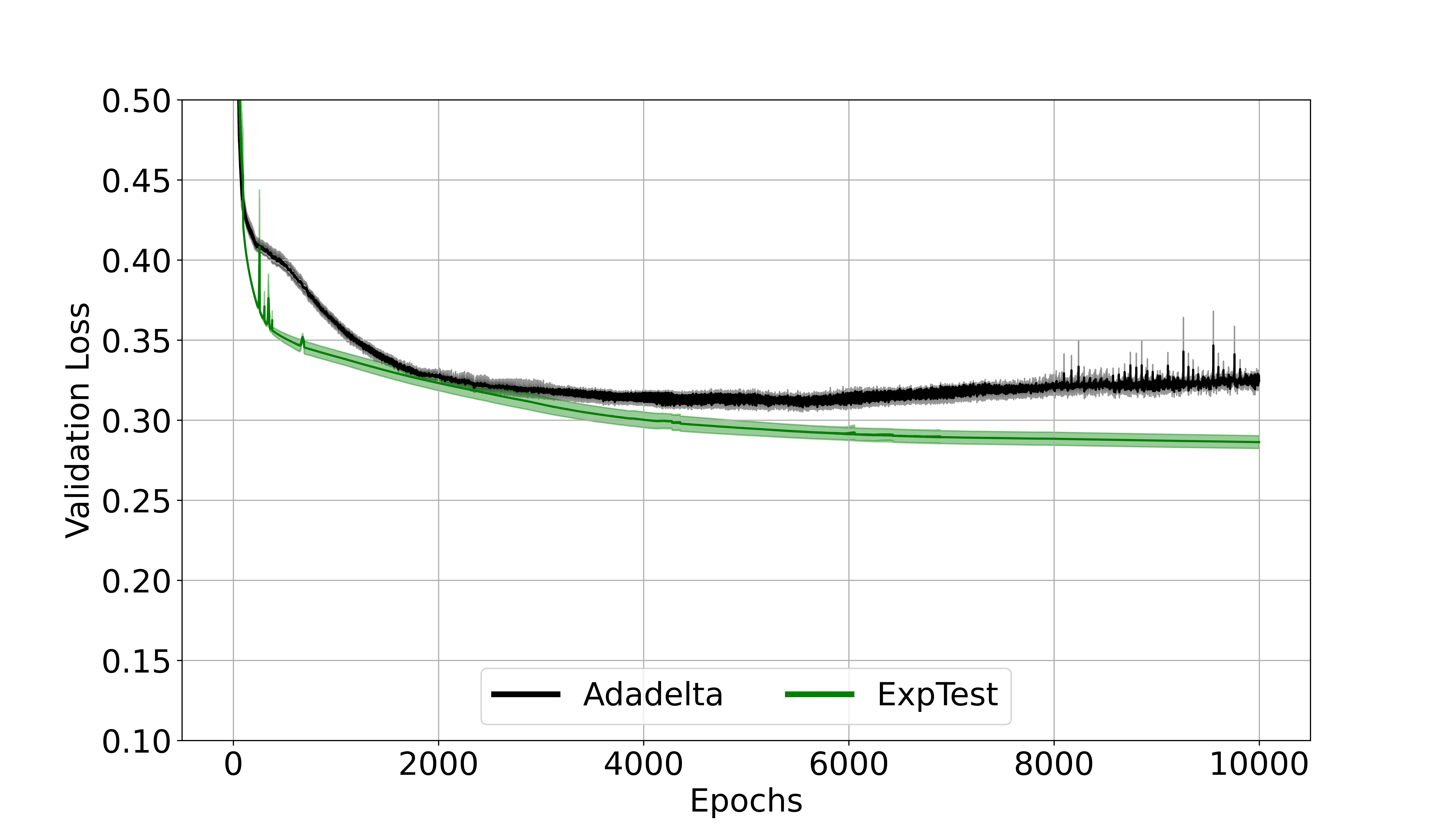}
        \caption{$\eta = \eta_{\text{max}}$}
        \label{fig:cali_val_sub5}
    \end{subfigure}
    \caption{Validation loss curves for regression with fully connected network on California Housing Dataset. Center line shows mean of 5 trials with standard error shown by shading. $y$-axis is limited at 0.5 for clarity.}
    \label{fig:cali_val_multi}
\end{figure}

We present the validation loss curves for the experiments conducted in Section~\ref{sec:housing} of the main text in Fig.~\ref{fig:cali_val_multi}. Notably, ExpTest displays a continually decreasing validation loss, whereas adaptive methods begin to rise, often before reaching the same minimum value as ExpTest. SGD with momentum displays similar behavior to ExpTest for well-chosen learning rates, but with considerably faster convergence. Interestingly, despite reaching much lower validation loss in the $0.01 \eta_{\text{max}}$ condition than any other optimizer, Adam does not display the best test set performance. Overall, these results corroborate work suggesting poor generalization capabilities of per-parameter adaptive methods \cite{10.5555/3294996.3295170}.

\ifCLASSOPTIONcaptionsoff
  \newpage
\fi

